\documentclass{article}
\pdfoutput=1

    \PassOptionsToPackage{numbers, compress}{natbib}
\usepackage[final]{neurips_data_2024}

\usepackage[utf8]{inputenc} % allow utf-8 input
\usepackage[T1]{fontenc}    % use 8-bit T1 fonts
\usepackage{hyperref}       % hyperlinks
\usepackage{url}            % simple URL typesetting
\usepackage{booktabs}       % professional-quality tables
\usepackage{amsfonts}       % blackboard math symbols
\usepackage{nicefrac}       % compact symbols for 1/2, etc.
\usepackage{microtype}      % microtypography
\usepackage[dvipsnames,table]{xcolor}         % colors

\usepackage{graphicx}
\usepackage{array}
\usepackage{amsmath}
\usepackage{arydshln}
\usepackage{xspace}
\usepackage{bm}
\usepackage{pifont}
\usepackage{subfigure}
\usepackage{todonotes}
\usepackage{enumitem}
\usepackage{colortbl}
\usepackage{listings}
\usepackage{multirow}
\definecolor{Gray}{gray}{0.6}
\definecolor{LightCyan}{rgb}{0.88,0.95,1}
\definecolor{blond}{rgb}{0.98, 0.94, 0.75}
\definecolor{light_gray}{gray}{0.92}
\definecolor{lighter_gray}{gray}{0.96}

\def \ie {\emph{i.e.}}
\def \eg {\emph{e.g.}}
\def \etal {\emph{et al.}}

\newcommand{\tit}[1]{\noindent\textbf{#1.}}
\newcommand{\titsheet}[1]{\noindent\textbf{#1}}

\newcommand{\tinytit}[1]{\noindent\textbf{#1.}}

\newcommand{\fullours}{Personalized Instance-based Navigation Embodied Dataset\xspace}
\newcommand{\ours}{PInNED\xspace}
\newcommand{\fulltask}{Personalized Instance-based Navigation\xspace}
\newcommand{\shorttask}{PIN\xspace}
\newcommand{\pin}{\includegraphics[scale=0.0033]{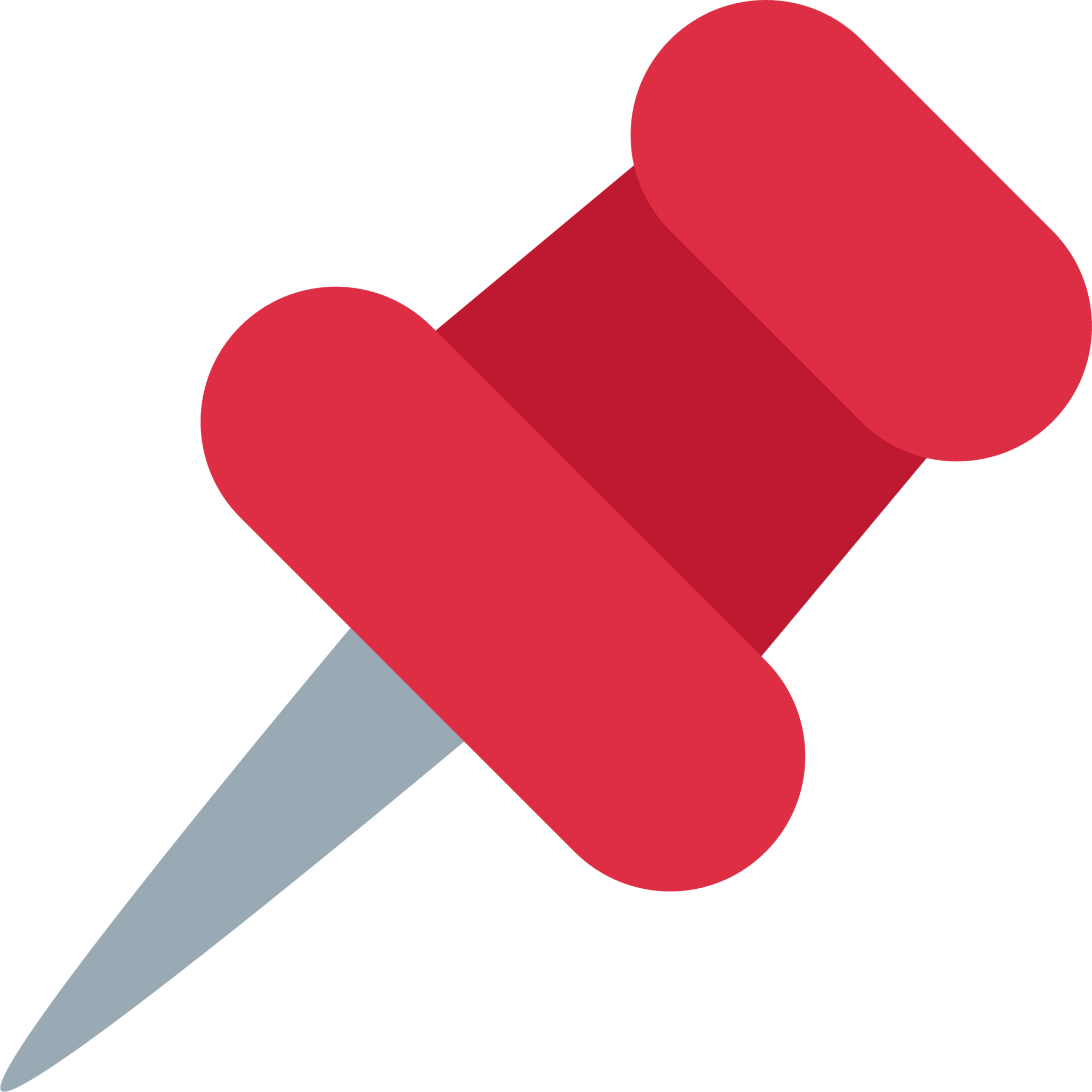}\xspace}
\newcommand{\bigpin}{\includegraphics[scale=0.0064]{figures/pin.png}\xspace}

\newcommand{\PreserveBackslash}[1]{\let\temp=\\#1\let\\=\temp}
\newcolumntype{C}[1]{>{\PreserveBackslash\centering}p{#1}}
\newcolumntype{R}[1]{>{\PreserveBackslash\raggedleft}p{#1}}
\newcolumntype{L}[1]{>{\PreserveBackslash\raggedright}p{#1}}
\makeatletter
  \renewcommand*\env@matrix[1][*\c@MaxMatrixCols c]{%
    \hskip -\arraycolsep
    \let\@ifnextchar\new@ifnextchar
  \array{#1}}
\makeatother

\newcommand{\cmark}{\ding{51}}%
\newcommand{\xmark}{\ding{55}}%

\newcommand{\ra}[1]{{R\#94wz}\xspace}
\newcommand{\rb}[1]{{R\#UcJy}\xspace}
\newcommand{\rc}[1]{{R\#kDUs}\xspace}
\newcommand{\rd}[1]{{R\#6hP2}\xspace}

\title{ 
\bigpin Personalized Instance-based Navigation Toward User-Specific Objects in Realistic Environments
}

\author{%
  Luca~Barsellotti\thanks{Equal contribution.} \quad Roberto~Bigazzi$^{*}$ \\ \textbf{Marcella~Cornia} \quad \textbf{Lorenzo~Baraldi} \quad \textbf{Rita~Cucchiara}\\
  University of Modena and Reggio Emilia, Italy \\
  \texttt{\{firstname.lastname\}@unimore.it} \\
  \vspace{-2mm}
  \\Project page: \href{https://aimagelab.github.io/pin}{aimagelab.github.io/pin}
}

\begin{document}

\maketitle
\vspace{-0.7cm}
\begin{abstract}
  In the last years, the research interest in visual navigation towards objects in indoor environments has grown significantly. This growth can be attributed to the recent availability of large navigation datasets in photo-realistic simulated environments, like Gibson and Matterport3D. However, the navigation tasks supported by these datasets are often restricted to the objects present in the environment at acquisition time. Also, they fail to account for the realistic scenario in which the target object is a user-specific instance that can be easily confused with similar objects and may be found in multiple locations within the environment. To address these limitations, we propose a new task denominated \textit{\fulltask} (\shorttask), in which an embodied agent is tasked with locating and reaching a specific personal object by distinguishing it among multiple instances of the same category. The task is accompanied by \pin \textit{\ours}, a dedicated new dataset composed of photo-realistic scenes augmented with additional 3D objects. In each episode, the target object is presented to the agent using two modalities: a set of visual reference images on a neutral background and manually annotated textual descriptions. Through comprehensive evaluations and analyses, we showcase the challenges of the \shorttask task as well as the performance and shortcomings of currently available methods designed for object-driven navigation, considering modular and end-to-end agents.
\end{abstract}

\vspace{-.4cm}
\begin{figure}[!ht]
    \begin{center}
    \resizebox{.96\linewidth}{!}{
    \includegraphics{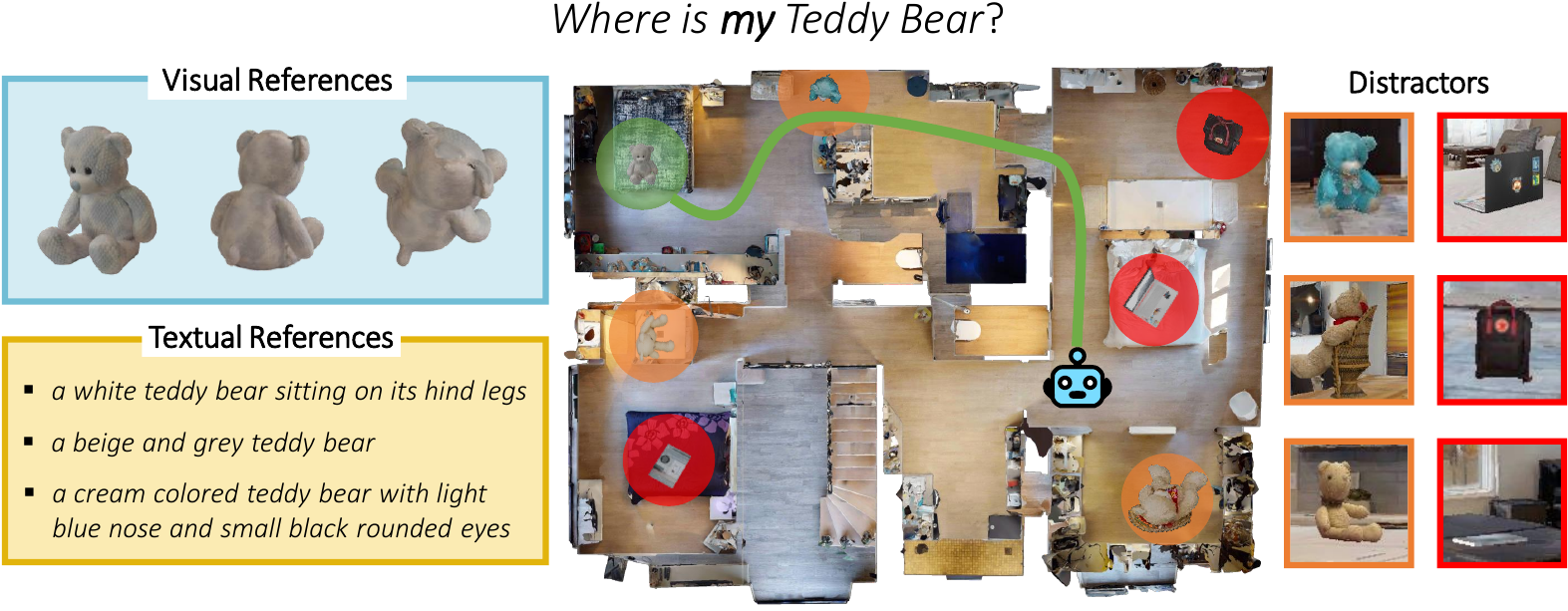}
    }
    \end{center}
    \vspace{-.3cm}
    \caption{We introduce the \shorttask task, where the agent is asked to navigate toward a personalized object instance using multimodal references and distinguish it from distractors (\ie, other objects of the same category as the target or of other categories). The target object, same category distractors, and other distractors are circled, respectively, in \textcolor{ForestGreen}{green}, \textcolor{orange}{orange}, and \textcolor{red}{red}.
    The total number of available objects in the dataset is 338, corresponding to different instances of 18 object categories.
    }
    \label{fig:persnav_example}
\end{figure}

\newpage

\section{Introduction}
\label{sec:intro}

Imagine a scenario where your child wants his favorite teddy bear, and he lost it somewhere in your house. In the foreseeable future, a ``smart'' domestic robot could be asked to find it. In that case, the robot will start roaming through the environment searching for the teddy bear. However, a-priori knowledge of the object category and visual cues related to the surroundings are not enough to solve the task, as the teddy bear has no predetermined location in the scene, could be potentially situated in several different places, and can be confused with other stuffed toys. While the recent advances in Embodied AI have significantly fostered the development of autonomous agents that can locate predefined target object categories, a benchmark that evaluates how agents tackle the challenges of reaching personal object instances in a photo-realistic environment is absent.

\tit{Motivation} The majority of current object-driven navigation tasks in Embodied AI define their goals as a general semantic category represented through text~\cite{anderson2018evaluation,batra2020objectnav,wani2020multion} (\eg, ``chair'', ``sofa'') or as a specific target instance defined by an image or description including the surrounding context in which the object can be found~\cite{bigazzi2023embodied,choi2021image,khanna2024goat,krantz2022instance,zhu2017target}. Moreover, these datasets rely on objects which were present at the time of acquisition of the environment~\cite{bigazzi2024mapping,chang2017matterport3d,dai2023think,khanna2024goat,krantz2022instance,majumdar2022zson,savva2019habitat,xia2018gibson,yadav2023habitat}. On the contrary, procedurally generated environments can freely contain additional objects and annotations~\cite{deitke2020robothor,deitke2022procthor,kolve2017ai2,li2021ion}. However, the appearance discrepancy between these environments and the real world or photo-realistic environments could affect the performance of the agents when deployed on robotic platforms~\cite{kadian2020sim2real}. Previous work has proposed loading additional 3D objects inside photo-realistic environments~\cite{maksymets2021thda} to improve agent navigation performance, to allow object interaction in static environments~\cite{shen2021igibson}, or to enable navigation towards multiple goals~\cite{wani2020multion}. However, no previous work has targeted loading objects that can be moved frequently and can appear in multiple contexts since loaded 3D models are kept in their initial spawn position.

\tit{Overview of the dataset} To overcome these issues, we propose the novel task of \textit{\fulltask} (\shorttask), where the agent needs to locate and reach a specific personalized target instance in the environment provided as reference images and textual descriptions, without information about the surrounding context.
An overview of \shorttask is shown in Fig.~\ref{fig:persnav_example}. In parallel with the definition of the task, we release \pin \ours (\fullours), a dedicated dataset of episodes for this setting that leverages the main advantages of both photo-realistic and procedurally generated embodied environments. In each episode, along with a unique target instance, distractors objects are placed in the scene to confound the navigation of the agent. 
Specifically, we built the dataset on top of the semantic annotations~\cite{yadav2023habitat} and scenes of Habitat-Matterport3D Dataset (HM3D)~\cite{ramakrishnan2021hm3d} with the injection of additional photo-realistic 3D objects accurately selected from Objaverse-XL~\cite{deitke2023objaverse}. The objects are positioned in each environment through a procedural spawning method on predefined suitable surfaces. \ours comprises 865.5k training episodes and 1.2k validation episodes built on top of 338 additional objects.

Finally, we adapt and test currently available navigation agents on the proposed dataset, showcasing the shortcomings of relevant approaches. In particular, we compare the performance of the two main categories of navigation agents for object-driven navigation, modular and end-to-end approaches, where we demonstrate that the versatility of modular methods leads to superior performance compared to the end-to-end counterparts; still, the task is far from being resolved. These experiments assess the difficulties posed by \shorttask task, highlighting the need for further research on the topic.
More details and release information on the codebase for the task, accompanying dataset, and evaluation benchmark are included in the Appendix.

\tit{Contributions} To sum up, our key contributions are threefold:\vspace{-0.1cm}
\begin{itemize}[left=3mm,noitemsep,topsep=0pt]
\item[\pin] We introduce the task of \fulltask (\shorttask). In this task, an agent must find and navigate towards a specific object instance without using the surrounding context. To increase the difficulty and compel the agent to learn to identify the correct instance, object distractors belonging to the same or different categories of the target are also added.
\item[\pin] We build and release \fullours (\ours), a task-specific dataset for embodied navigation based on photo-realistic personalized objects from Objaverse-XL dataset injected in the environments of HM3D dataset. Overall, it comprises 338 object instances belonging to 18 different categories positioned within 145 training and 35 validation environments, for a total of approximately 866.7k navigation episodes.
\item[\pin] We evaluate currently available object-driven methods on the newly proposed dataset demonstrating their limitations in tackling the proposed \shorttask task.
\end{itemize}
\section{Related Work}
\label{sec:related}
\tit{Object-based Embodied Datasets}
In recent years, research aimed at the development of intelligent autonomous agents has acquired increasing interest with the release of simulation platforms like Habitat~\cite{puig2023habitat,savva2019habitat,szot2021habitat}, AI2-THOR~\cite{kolve2017ai2}, RoboTHOR~\cite{deitke2020robothor}, and ProcTHOR~\cite{deitke2022procthor}, as well as datasets of scenes for robotic navigation like Gibson~\cite{shen2021igibson,xia2018gibson}, Matterport3D~\cite{chang2017matterport3d}, and Habitat-Matterport3D (HM3D)~\cite{ramakrishnan2021hm3d}.
The evaluation of the capabilities of such agents can be performed on multiple embodied tasks~\cite{anderson2018vision,rawal2024aigen,shridar2020alfred} mimicking different real-world requirements. 
PointGoal Navigation (PointNav)~\cite{anderson2018evaluation} requires the agent to reach specific relative coordinates to its starting position. 
In object-oriented navigation, the agent is tasked to find any instance of an object category (ObjectNav)~\cite{anderson2018evaluation,batra2020objectnav}, multiple objects in sequence (MultiON)~\cite{wani2020multion}, or a specific instance of a category (ION)~\cite{li2021ion}. 
Other embodied navigation tasks are ImageGoal navigation (ImageNav)~\cite{choi2021image,zhu2017target} that requires the agent to reach the position where the goal image has been taken, and a more object-oriented formulation of ImageNav called Instance-Specific Image Goal Navigation (InstanceImageNav)~\cite{krantz2022instance} that requires to reach a precise object instance given a photo of it. Recently, the GOAT-Bench benchmark has been introduced, which requires finding sequences of target objects using multimodal references~\cite{khanna2024goat}. However, GOAT-Bench targets are constrained to the objects captured in the environment at acquisition time.
To the best of our knowledge, \ours is the only dataset focused on navigation toward personalized targets that uses multimodal references, injects additional objects into photorealistic environments, and requires the agent to distinguish the correct instance from distractors without relying on context.

\tit{Object-based Navigation Agents}
Object-based methods for navigation agents can be divided into two categories depending on their design: modular approaches and end-to-end approaches.
Modular approaches are composed of multiple components, usually a mapping module, an exploration procedure, and an object detection method. Some approaches adapted the architecture proposed by ANS~\cite{chaplot2020neural} for object goal navigation by building semantic maps to locate the target~\cite{chaplot2020object,landi2022spot,ramakrishnan2022poni,zhu2022navigating}. Following, Stubborn~\cite{luo2022stubborn} proposed a strong baseline using a heuristic exploration method.
Among end-to-end methods, Mousavian~\etal~\cite{mousavian2019visual} and Yang~\etal~\cite{yang2018visual} worked on improving visual representations, Mayo~\etal~\cite{mayo2021visual} used spatial attention maps, and Ye~\etal~\cite{ye2021auxiliary} used auxiliary tasks.
Other related work leveraged object relation graphs~\cite{druon2020visual,du2020learning,pal2021learning}. THDA~\cite{maksymets2021thda}, instead, used 3D scans of objects from YCB dataset~\cite{calli2015ycb} to augment the training dataset.
Recently, PIRLNav~\cite{ramrakhya2023pirlnav} used a two-stage learning strategy, Chen~\etal~\cite{chen2023object} used a method based on recursive implicit maps, and OVRL~\cite{yadav2023ovrl,yadav2023offline} exploited self-supervised visual pretraining to boost agent capabilities.
Additionally, zero-shot object goal navigation has been recently explored by ZER~\cite{al2022zero}, ZSON~\cite{majumdar2022zson}, and ORION~\cite{dai2023think}.

\tit{Personalized Instance Recognition}
In recent years, foundation models have revolutionized the Computer Vision field. CLIP~\cite{radford2021learning} learned a multimodal embedding space by performing large-scale contrastive training, demonstrating impressive capabilities in zero-shot classification. DINO~\cite{caron2021emerging,oquab2023dinov2} is trained with a self-supervised paradigm achieving strong semantic correspondence properties among features~\cite{barsellotti2024fossil,barsellotti2024training,zhang2024tale}. Segment Anything (SAM)~\cite{kirillov2023segany} has been trained to predict precise class-agnostic masks given a prompt. The feature spaces learned by these models are semantically rich and can be exploited in tasks that involve the recognition of general object categories.
However, adapting a model for recognizing personalized objects in images remains an open challenge.
For example, SuperGlue~\cite{sarlin2020superglue} leveraged an attention-based graph neural network on the local descriptors extracted with the SuperPoint model~\cite{detone2018superpoint} to perform image matching and has been used in Mod-IIN~\cite{krantz2023navigating} and GOAT~\cite{chang2023goat} to tackle the InstanceImageNav task. IEVE~\cite{lei2024instance}, instead, proposes an Exploration-Verification-Exploitation framework that combines a segmentation model and a keypoint matcher to recognize distant objects and confirm them when the agent is closer;
while PerSAM~\cite{zhang2023personalize}, performed personalized segmentation allowing SAM to localize a user-provided target.
In the same setting, SegIc~\cite{meng2023segic} introduced a mask decoder with in-context instructions on top of the dense correspondences from DINOv2~\cite{oquab2023dinov2}, while Matcher~\cite{liu2023matcher} leveraged DINOv2 to extract prompts for SAM in a training-free paradigm.

\section{\fulltask}
\label{sec:task}

In this section, we outline the \fulltask task, highlighting its key characteristics and comparing it to existing embodied tasks. Following, we detail the composition and generation process of the \ours dataset.

\begin{table}[!t]
    \centering
    \caption{Comparison of the different object-driven datasets for embodied navigation, considering the photo-realism of scenes and targets, the availability of additional objects with variable spawn locations, the modalities of the provided references, and whether the dataset is instance-oriented.}
    \resizebox{\linewidth}{!}{
    \begin{tabular}{l c c c c c c c}
        \toprule
        & Photo-Realistic & Photo-Realistic & Additional & Visual & Descriptive & Variable & Instance\\
        Dataset & Scenes & Targets & Objects & Reference & Reference & Placement & Goal\\
        \midrule
        MP3D~\cite{chang2017matterport3d} & \cmark & \cmark & \xmark & \xmark & \xmark & \xmark & \xmark\\
        AI2-THOR~\cite{kolve2017ai2} & \xmark & \xmark & \cmark & \xmark & \xmark & \cmark & \xmark\\
        Gibson~\cite{xia2018gibson} & \cmark & \cmark & \xmark & \xmark & \xmark & \xmark & \xmark\\
        Robo-THOR~\cite{deitke2020robothor} & \xmark & \xmark & \cmark & \xmark & \xmark & \cmark & \xmark\\
        MultiON*~\cite{wani2020multion} & \cmark & \xmark & \cmark & \xmark & \cmark & \cmark & \xmark\\
        HM3D~\cite{ramakrishnan2021hm3d} & \cmark & \cmark & \xmark & \xmark & \xmark & \xmark & \xmark\\
        ProcTHOR & \xmark & \xmark & \cmark & \xmark & \xmark & \cmark & \xmark \\
        ION~\cite{li2021ion} & \xmark & \xmark & \cmark & \xmark & \cmark & \cmark & \cmark\\
        THDA~\cite{maksymets2021thda} & \cmark & \cmark & \cmark & \xmark & \xmark & \cmark & \xmark\\
        ZSON~\cite{majumdar2022zson} & \cmark & \cmark & \xmark & \xmark & \cmark & \xmark & \xmark \\
        InstanceImageNav~\cite{krantz2023navigating} & \cmark & \cmark & \xmark & \cmark & \xmark & \xmark & \cmark \\
        ZIPON~\cite{dai2023think} & \cmark & \cmark & \xmark & \xmark & \cmark & \xmark & \cmark \\
        GOAT-Bench~\cite{khanna2024goat} & \cmark & \cmark & \xmark & \cmark & \cmark & \xmark & \cmark \\
        \midrule
        \textbf{\ours (Ours)} & \cmark & \cmark & \cmark & \cmark & \cmark & \cmark & \cmark\\
        \bottomrule
    \end{tabular}
    }
    \vspace{-0.2cm}
\label{tab:dataset_comparison}
\end{table}

\subsection{Task Definition}
\label{subsec:task_definition}
The \shorttask task requires the agent to navigate toward a predetermined specific object instance (\eg, ``\textit{a yellow backpack with red straps}'') in an unexplored environment.
Each target object needs to be found in the environment, distinguishing it from multiple distractors of the same category and other objects of different categories. In this setting, the target object can be provided to the agent in two different modalities: (i) as a set of RGB images depicting the target object rendered in an isolated context on a neutral background, and (ii) as a set of textual descriptions of the object instance appearance.

At the beginning of each episode of \shorttask, the agent is initialized at a random pose $\bm{x}_0$ in an unseen environment. A single target instance $o^i$ is selected as the goal $g$, such that $g \in C^a \subset O$, where $C^a$ is a set of instances belonging to the same object category and $O$ is the set of all available objects. 
The goal $g$ is placed in the environment at a position $\bm{z}$. Additionally, $n$ distinct instances $o^j$ ($o^j \in C^a \wedge i \neq j$) are positioned in the environment, along with $m$ distinct instances $o^k$ ($o^k \in (O \setminus C^a)$). At the end of the episode, the navigation is considered successful if the agent selects the `\textit{stop}' action before the maximum allowed number of timesteps $T$, with an Euclidean distance between the position of the agent at the current timestep $\bm{x}_t$ and the target position $\bm{z}$ lower than 1 meter.
The action space of the agent for the task is defined by six possible actions, where at each timestep $t$, the action $a_t \in$ \{`\textit{stop}', `\textit{move ahead}', `\textit{turn left}', `\textit{turn right}', `\textit{tilt up}', `\textit{tilt down}'\}. 

\subsection{Comparison with Other Tasks}
\label{subsec:comparison_with_other_tasks}
The proposed task locates itself among PointNav~\cite{anderson2018evaluation}, ObjectNav~\cite{anderson2018evaluation,batra2020objectnav}, ImageNav~\cite{choi2021image}, and the recently defined task of InstanceImageNav~\cite{krantz2022instance}.
\shorttask exhibits similarities to ObjectNav, InstanceImageNav, and the recently introduced GOAT-Bench~\cite{khanna2024goat} (see Sec.~\ref{sec:related}).

However, it diverges from the traditional ObjectNav task because, differently from the standard objective of finding any instance of a general object category, \shorttask requires locating a specific instance, such as ``\textit{black and white striped trekking backpack}'' instead of any ``\textit{backpack}''.
\shorttask leverages zero-shot properties at the instance level, as the object instances used for the training split differ from those included in the validation episodes.
This requires agents to focus on the specific characteristics of the target object defined by the input references and avoid being misled by instances of the same category that are not the actual target.

Furthermore, \shorttask differs from InstanceImageNav and GOAT-Bench in various aspects. First, the target object is represented by a collection of images with neutral backgrounds, rather than being shown in its current spatial context. 
InstanceImageNav and GOAT-Bench are based on a set of general object categories that are included in the dataset of scenes and, therefore, these objects are static and frozen in the 3D model of the environment. Instead, the peculiarity of \shorttask is that it is created using a set of additional photo-realistic personal objects from a collection of 3D objects that can be placed and moved in different locations of the environment between different episodes. Using additional objects allows to avoid reconstruction errors and artifacts that can distort the appearance of the target.
This unique characteristic compels the agent to discern and extract the defining features of the target object while maintaining invariance to the surrounding context in which it is situated since personal objects can be moved frequently and could be placed in multiple suitable locations.

\begin{figure}[!t]
    \centering
    \resizebox{\linewidth}{!}{
        \setlength{\tabcolsep}{.06em}
        \begin{tabular}{c p{0mm} c p{0mm} c p{2mm}p{2mm} c p{2mm}p{2mm} c p{2mm}p{2mm} c}
            \addlinespace[2mm]
            \includegraphics[height=0.3\textwidth]{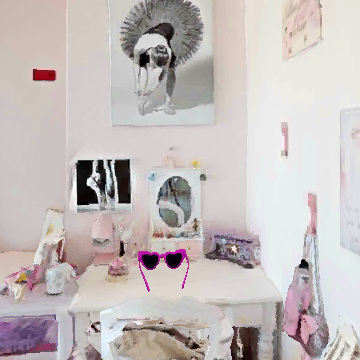} &&
            \includegraphics[height=0.3\textwidth]{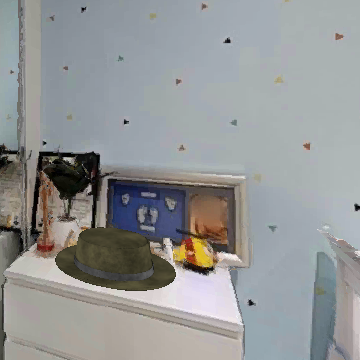} &&
            \includegraphics[height=0.3\textwidth]{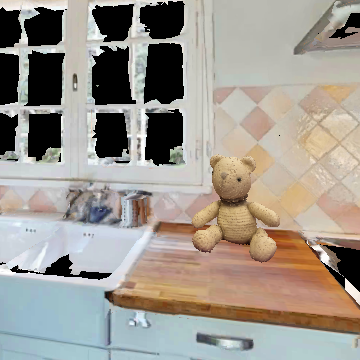} & & &
            \includegraphics[height=0.3\textwidth]{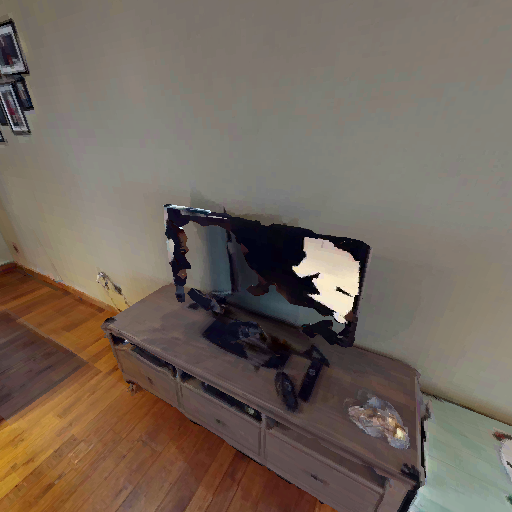} &&&
            \includegraphics[height=0.3\textwidth]{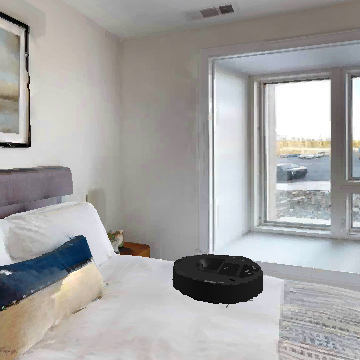} &&&
            \includegraphics[height=0.3\textwidth]{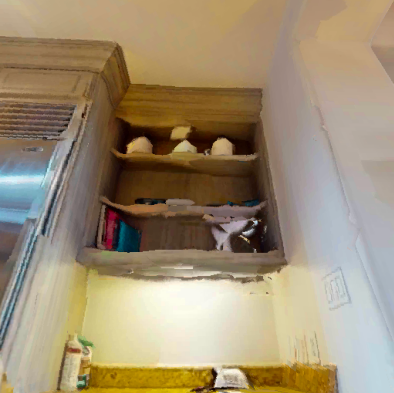} \\
            \addlinespace[2mm]
            \multicolumn{5}{c}{\Large\textbf{\ours (Ours)}} &&& \textbf{\Large InstanceImageNav} &&& \textbf{\Large MultiON} &&& \textbf{\Large GOAT-Bench} \\
        \end{tabular}
    }
    \caption{Comparison of observations depicting different targets in the embodied setting of our \ours dataset with the target objects of MultiON, InstanceImageNav, and GOAT-Bench datasets.}
    \vspace{-.2cm}
    \label{fig:vs_instance}
\end{figure}

Similarly to GOAT-Bench, \shorttask provides a multimodal input to the agent, including textual descriptions of the target instances alongside the images. However, GOAT-Bench ignores the presence of instances of the same category of the target in the scene, whereas this is the core challenge of \shorttask. Additionally, it is worth noting that while text alone can sometimes provide precise identification of the specific instance, it can also be ambiguous. Visual references, although generally clearer, are not always available in real-world scenarios.  Therefore, the two modalities cover different real-world requirements and both deserve to be studied.
An extensive comparison of current object-driven dataset properties is reported in Table~\ref{tab:dataset_comparison}, which presents the following columns:
\begin{itemize}[left=0.4cm]
    \item[-] \textit{Photo-Realistic Scenes}: the presence of photo-realistic scans taken from real-world environments (e.g. the scenes of HM3D are built from scans of real environments, while scenes in AI2-THOR are hand-built by professional 3D artists);
    \item[-] \textit{Photo-Realistic Targets}: the availability of photo-realistic objects that can be used as navigation targets. In \ours we carefully selected objects with realistic appearances. Procedurally-generated datasets, instead, tend to favor customizability over realism;
    \item[-] \textit{Additional Objects}: the inclusion of target objects that were not present at the time of capture. Datasets like GOAT-Bench target objects which were already present in the acquired scene, while \ours targets objects injected in the scene afterward;
    \item[-] \textit{Visual Reference}: providing visual target references for each navigation episode;
    \item[-] \textit{Descriptive Reference}: providing natural language descriptions as targets for each episode;
    \item[-] \textit{Variable Placement}: the possibility of having variable spawning positions for the targets within the dataset;
    \item[-] \textit{Instance Goal}: the inclusion of navigation episodes in which the goal is to reach the exact instance indicated to the agent.
\end{itemize}
Moreover, a qualitative comparison of goal objects observed in their position in the environment from different datasets is depicted in Fig.~\ref{fig:vs_instance}.

\subsection{Dataset}
\label{subsec:dataset}
\tit{Categories and Instances}
We selected a pool of 18 object categories from the assets contained in Objaverse-XL dataset~\cite{deitke2023objaverse}: `\textit{backpack}', `\textit{bag}', `\textit{ball}', `\textit{book}', `\textit{camera}', `\textit{cellphone}', `\textit{eyeglasses}', `\textit{hat}', `\textit{headphones}', `\textit{keys}', `\textit{laptop}',  `\textit{mug}', `\textit{shoes}', `\textit{teddy bear}', `\textit{toy}', `\textit{visor}', `\textit{wallet}', `\textit{watch}', for a total of 338 additional objects. Each category contains an average of 18.8 objects, with a standard deviation of 5.5.
The 3D objects are selected with human supervision to ensure photo-realism and uniqueness, which are critical requirements for tackling the \shorttask task. Finally, the 3D models of the objects are manually rescaled to have comparable dimensions to their real-world counterparts. In this procedure, we rendered each given object in a scene from HM3D and varied the scale of the object until the result was realistic according to our judgment. Hence, each of the 338 additional objects has a manually fixed scale that is adopted when the object is injected into the navigation episodes.

\begin{figure}[t]
    \centering
    \normalsize
    \resizebox{.98\linewidth}{!}{
        \setlength{\tabcolsep}{.05em}
        \begin{tabular}{c p{1mm} c p{1mm} c p{10mm} c p{1mm} c p{1mm} c p{10mm} c p{1mm} c p{1mm} c}
            \multicolumn{5}{c}{\Huge\textbf{Backpack}} && \multicolumn{5}{c}{\Huge\textbf{Ball}} &&
            \multicolumn{5}{c}{\Huge\textbf{Camera}} \\
            \addlinespace[2mm]
            \addlinespace[4mm]
            \includegraphics[height=0.3\textwidth]{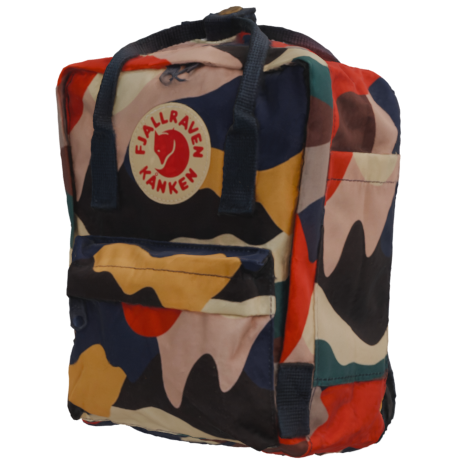} &&
            \includegraphics[height=0.3\textwidth]{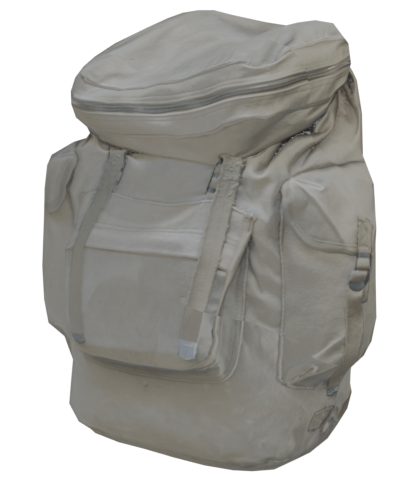} &&
            \includegraphics[height=0.3\textwidth]{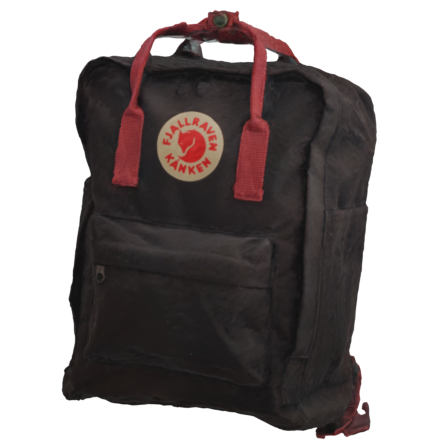} &&
            \includegraphics[height=0.3\textwidth]{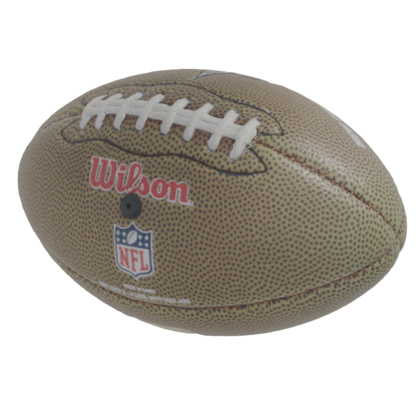} &&
            \includegraphics[height=0.3\textwidth]{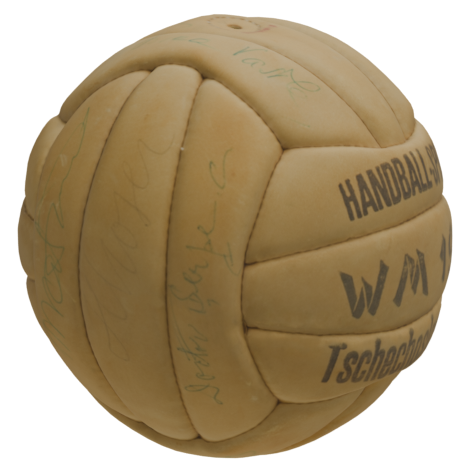} &&
            \includegraphics[height=0.3\textwidth]{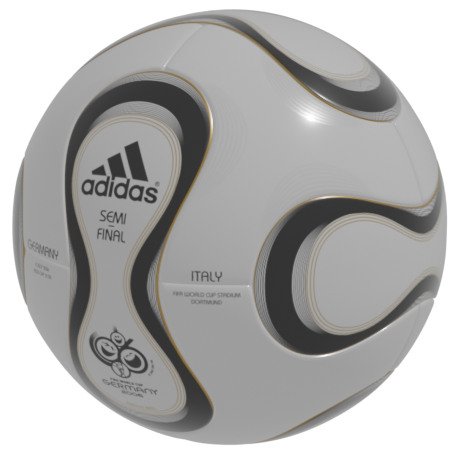} &&
            \includegraphics[height=0.3\textwidth]{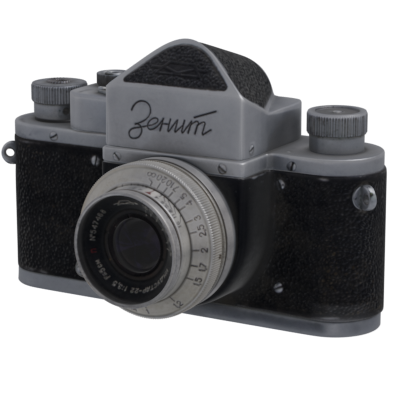} &&
            \includegraphics[height=0.3\textwidth]{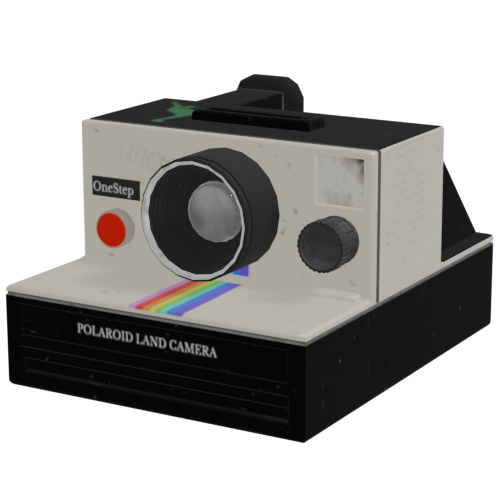} &&
            \includegraphics[height=0.3\textwidth]{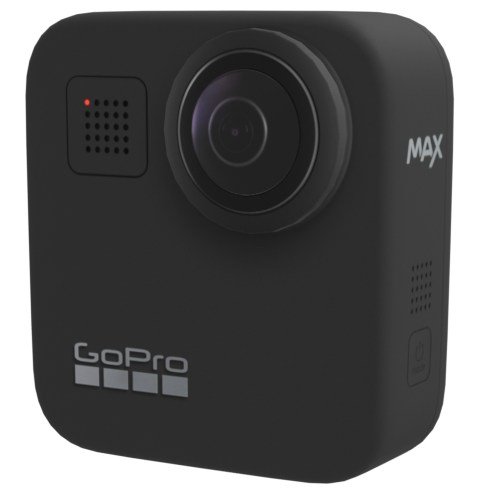}
            \\

            \addlinespace[4mm]
            \cmidrule{1-5} \cmidrule{7-11} \cmidrule{13-17}
            \addlinespace[4mm]
            \includegraphics[height=0.3\textwidth]{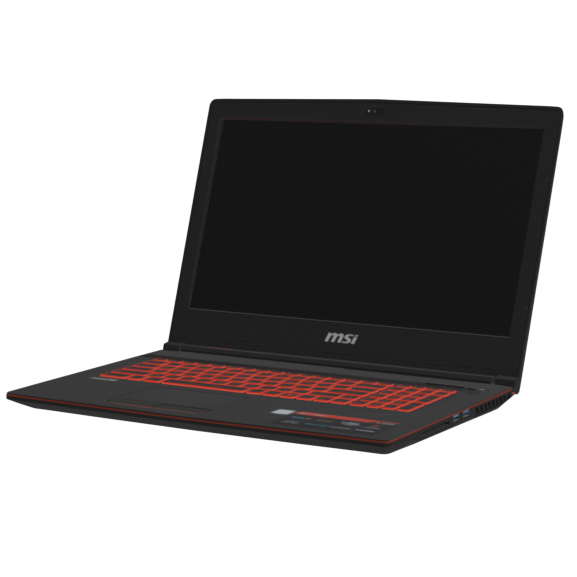} &&
            \includegraphics[height=0.3\textwidth]{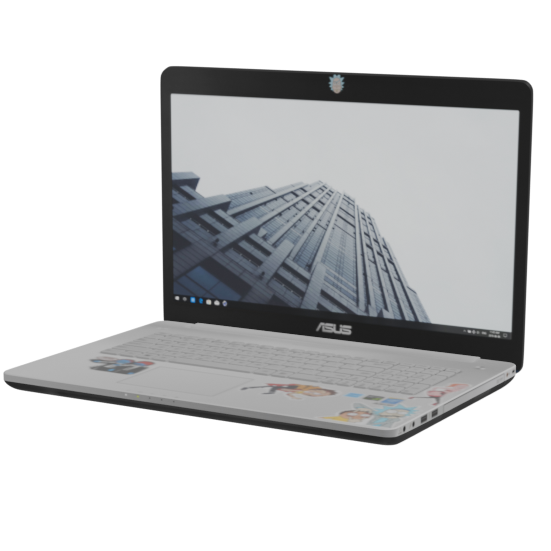} &&
            \includegraphics[height=0.3\textwidth]{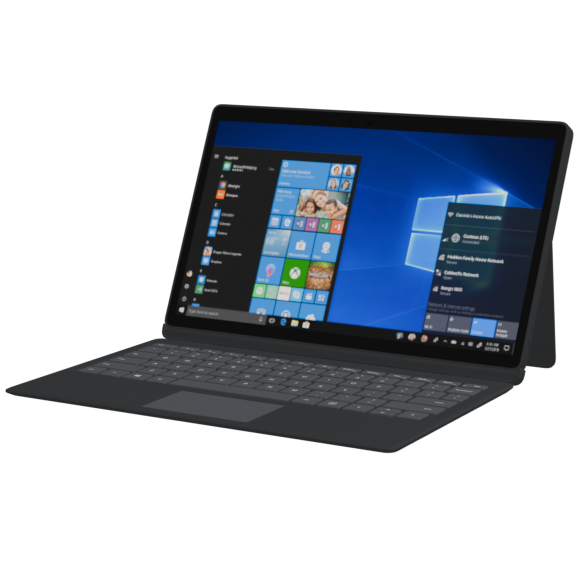} &&
            \includegraphics[height=0.3\textwidth]{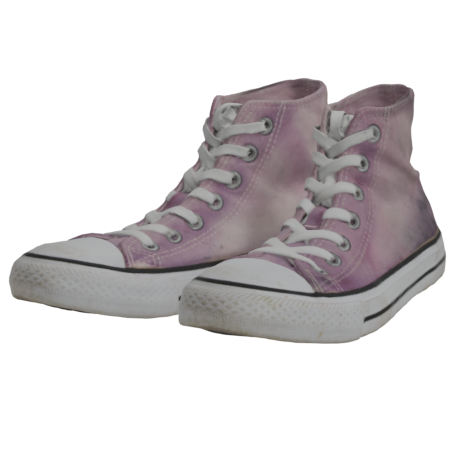} &&
            \includegraphics[height=0.3\textwidth]{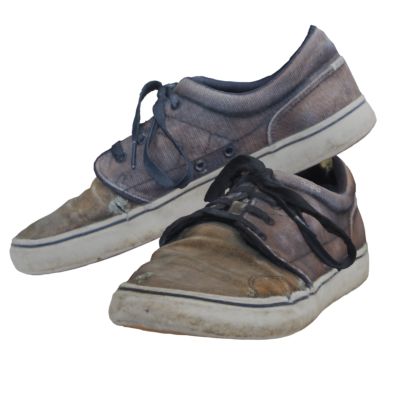} &&
            \includegraphics[height=0.3\textwidth]{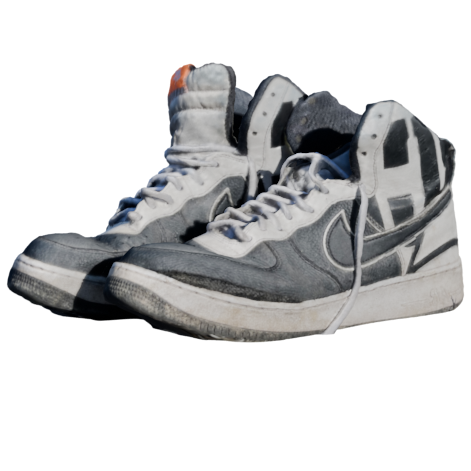} &&
            \includegraphics[height=0.3\textwidth]{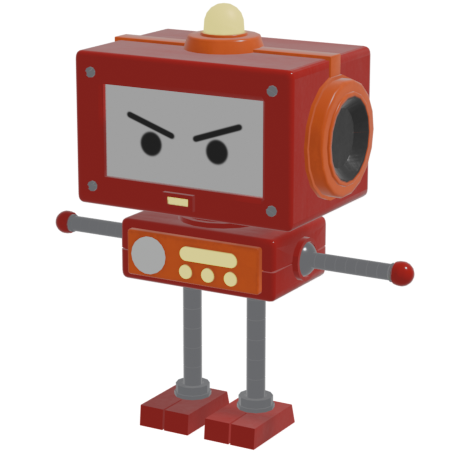} &&
            \includegraphics[height=0.3\textwidth]{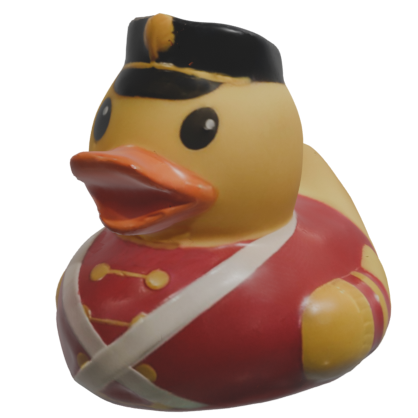} &&
            \includegraphics[height=0.3\textwidth]{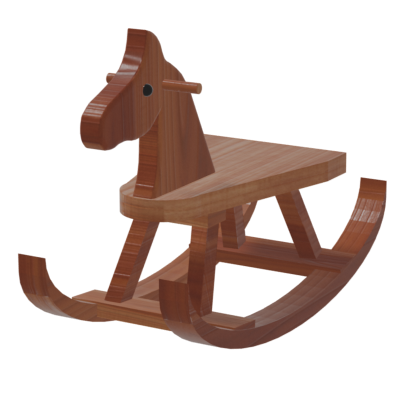}
            \\
            \addlinespace[4mm]
            \multicolumn{5}{c}{\Huge\textbf{Laptop}} && \multicolumn{5}{c}{\Huge\textbf{Shoes}} &&
            \multicolumn{5}{c}{\Huge\textbf{Toy}} \\
            \addlinespace[2mm]
        \end{tabular}
    }
    \caption{Sample input images of personalized targets from \ours dataset. We include three instances from various object categories within the dataset.}
    \label{fig:dataset_samples}
\end{figure}

\tit{Input References}
The input images for each target personalized object are generated by rendering the 3D mesh of the object in an isolated setting.
Specifically, the input images are not expected to match the camera specification of the navigating agent~\cite{krantz2022instance}.
The digital camera undergoes a 30-degree yaw rotation to capture a favorable perspective of the objects. Each instance is then rotated 180 degrees in yaw to view its reverse side, followed by a 90-degree pitch rotation to observe the object from above. This procedure produces a set of three input images for each target object.
An illustration of the acquired reference images is displayed in Fig.~\ref{fig:dataset_samples}.
Moving on to the textual references, manually annotated descriptions are produced for each target personalized object with the scope of highlighting the details that allow the agent to distinguish it from other instances of the same category. 
Specifically, we provide three descriptions for each personalized object in the \ours dataset. 
To annotate the descriptions, we provided two object instances at a time to the annotators, asking them to describe one of the two objects in such a way that it is distinguishable from the other. This procedure results in a total of 960 unique words and an average of 10.7 words per description. Additional samples of input references are included in the Appendix.

\tit{Scenes}
The benchmark defined by the \shorttask dataset is situated in the indoor photo-realistic scenes (\eg, apartments, offices, houses) within the semantically-annotated subset~\cite{yadav2023habitat} of Habitat-Matterport3D (HM3D)~\cite{ramakrishnan2021hm3d} which consists of 145 environments for the training split and 36 for validation set. However, one validation scene is ignored as it represents an art gallery and has no suitable spawnable surfaces. HM3D was selected due to its status as the largest publicly available dataset of semantically annotated indoor spaces with photo-realistic quality for embodied navigation.

\begin{figure}[!t]
    \centering
    \resizebox{.95\linewidth}{!}{
        \setlength{\tabcolsep}{.1em}
        \begin{tabular}{rr p{2mm} rr }
            \includegraphics[height=0.215\textwidth]{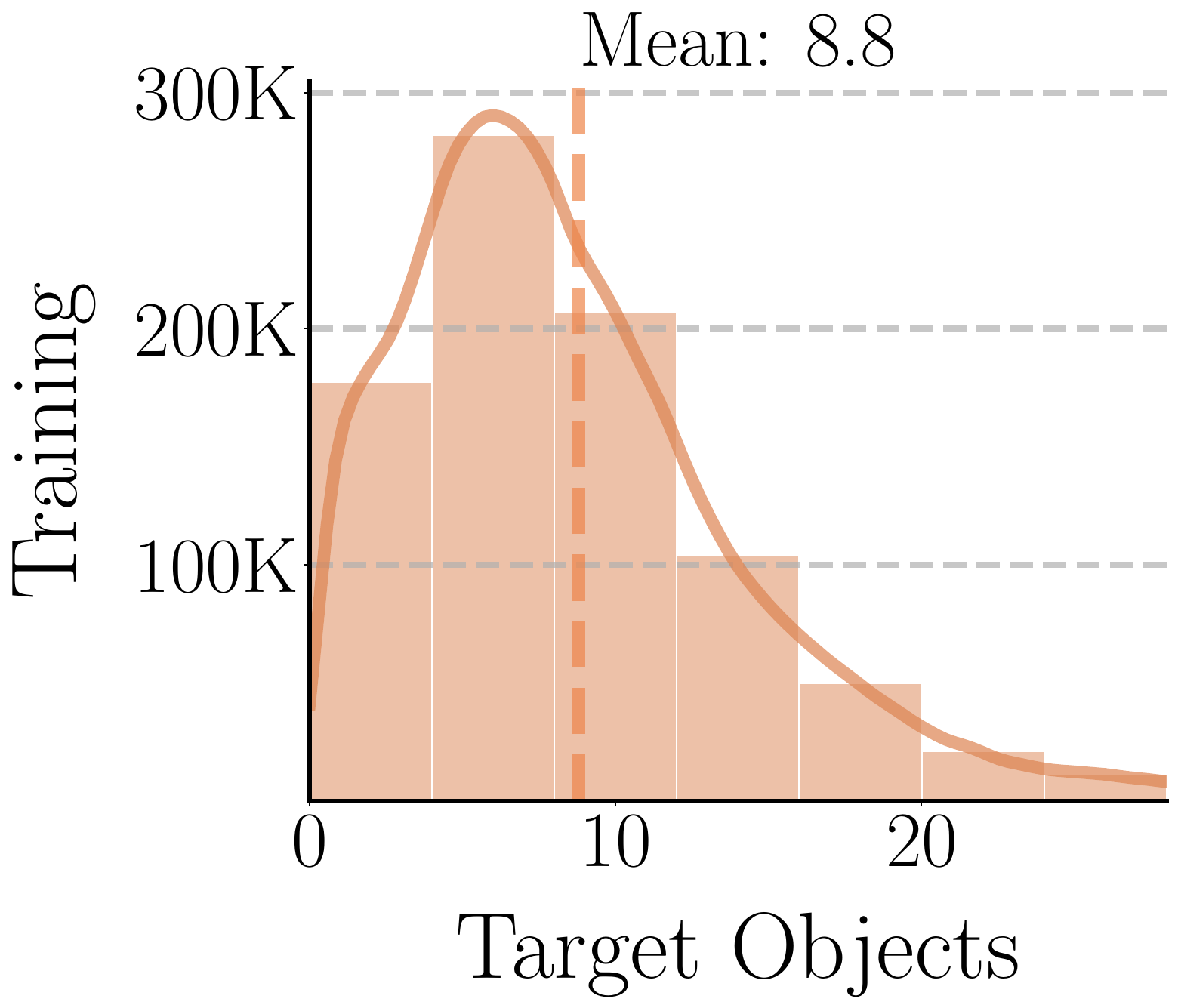} &
            \includegraphics[height=0.215\textwidth]{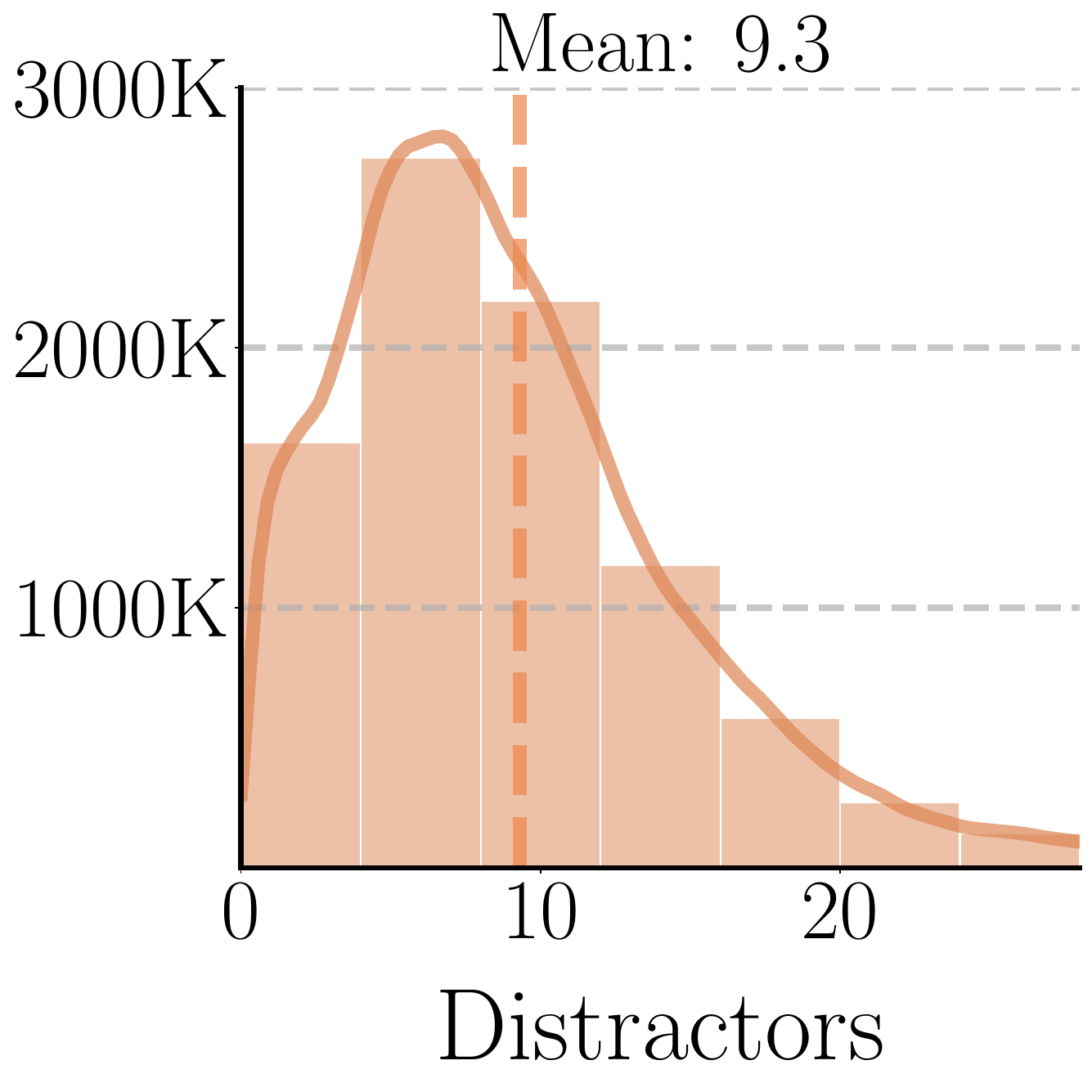} & &
            \includegraphics[height=0.215\textwidth]{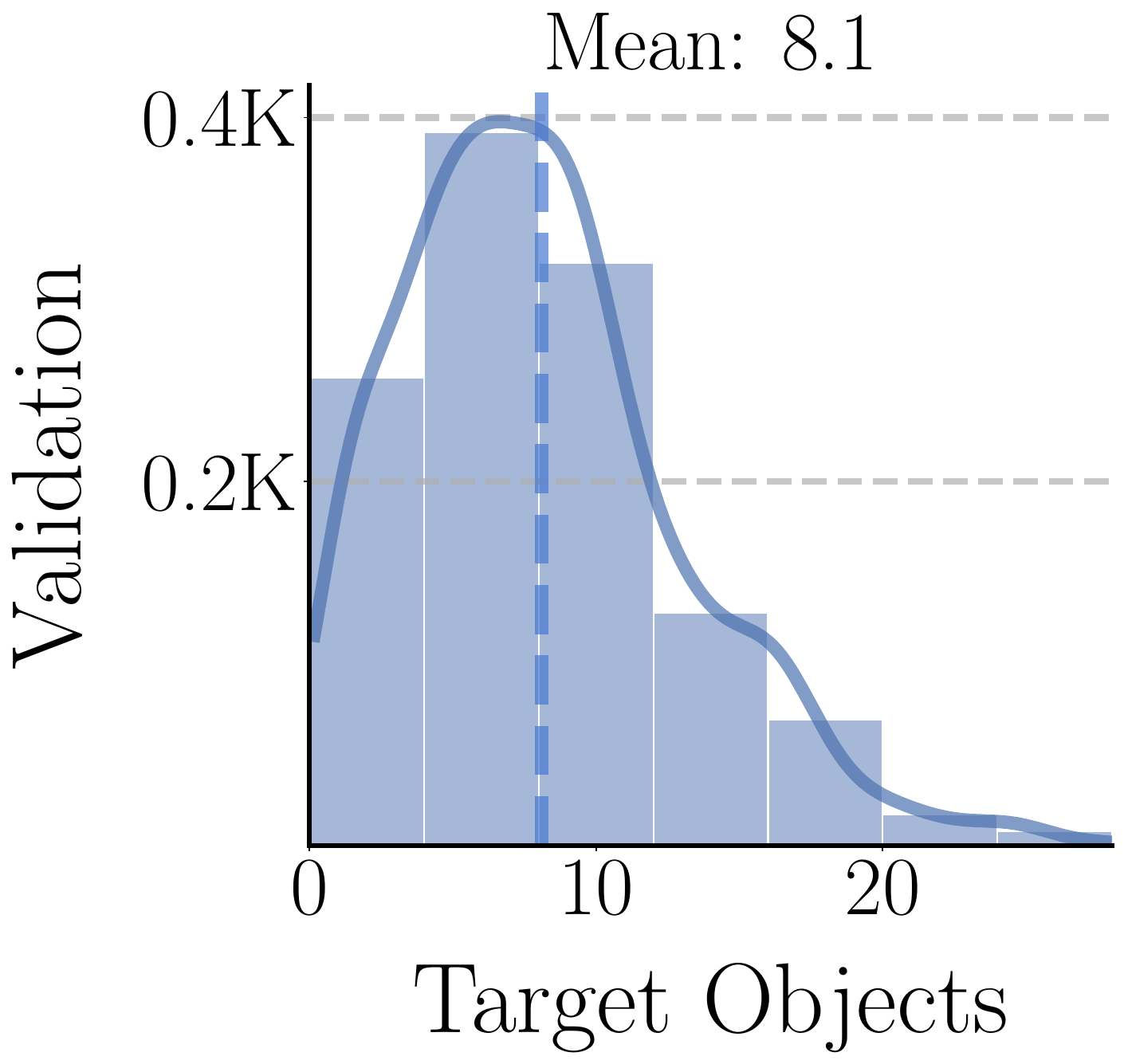} &
            \includegraphics[height=0.215\textwidth]{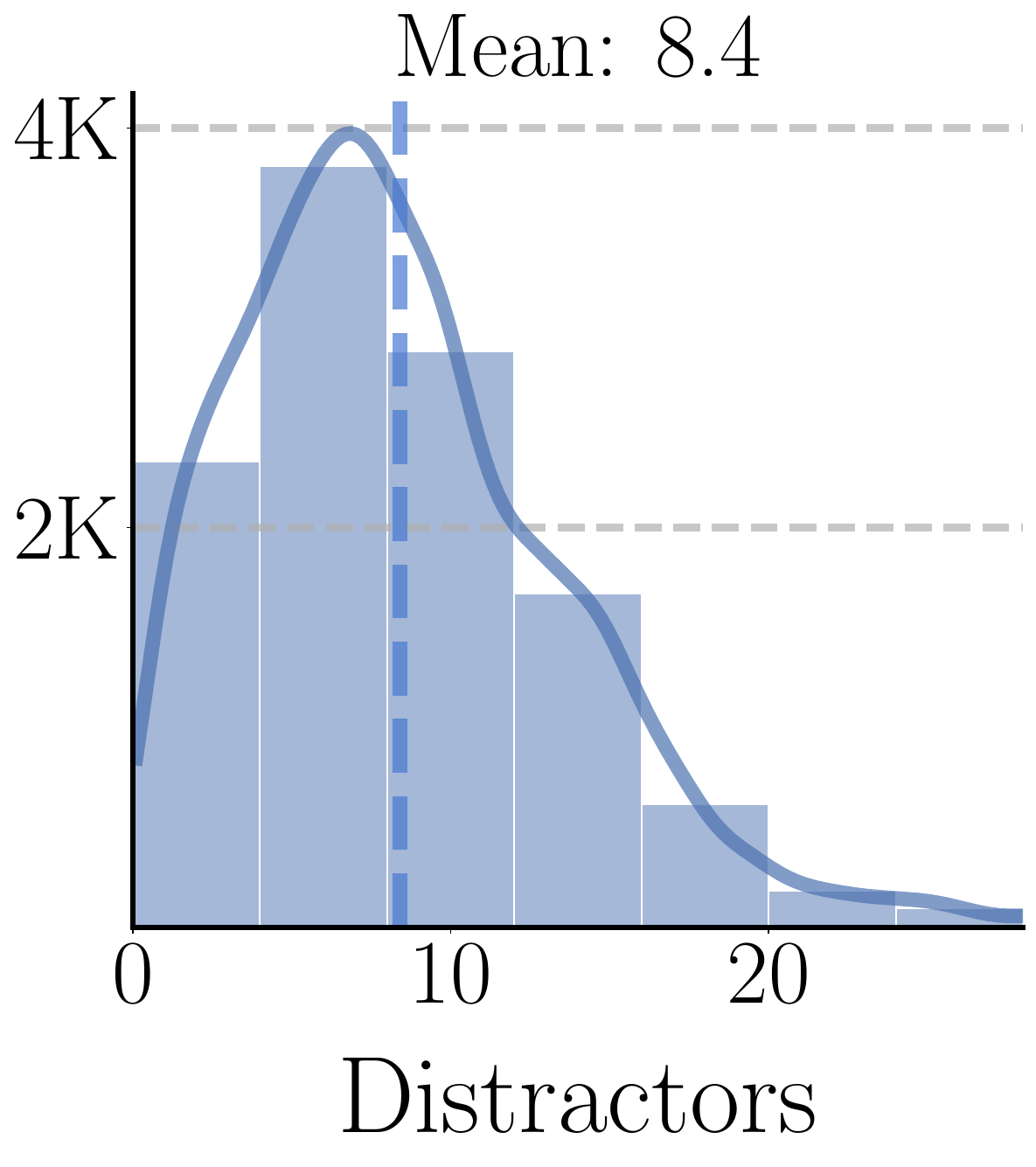} \\
        \end{tabular}
    }
    \caption{Plots of the distance statistics for the splits of \ours dataset. The episodes of the training (\textcolor{orange}{orange}) and validation splits (\textcolor{RoyalBlue}{blue}) are presented in terms of geodesic distance from the start position to the target object (left) and to all the distractors (right). All the distances are plotted in meters, and the mean value of each plot is shown on top.} 
    \label{fig:distances_plots}
    \vspace{-.2cm}
\end{figure}

\tit{Episode Generation}
During the generation of the dataset, the bounding boxes of the surfaces in the environment are extracted using the semantic annotations of the scene.
To obtain the bounding box from the texture, we extracted the point cloud 3D model of each scene and ensured that each point retained its associated annotation color.
Subsequently, points were clustered by annotation color to create the bounding box associated with each piece of furniture.
The spawning position of each object is selected by sampling from the positions of a curated set of suitable surface macro-categories included in the semantic annotations of HM3D. The surface categories selected for the creation of the dataset are: \textit{armchair, bed, bench, cabinet, piano, rug, sofa, table}. These specific surfaces are chosen because of the high probability of personalized objects being positioned on top.

In each episode of the \ours dataset, a single instance of a specific category is chosen as the target object. Consequently, up to $6$ instances belonging to the same category, and up to 13 objects from other categories, are added to the environment as distractors. All additional objects placed in the environment are constrained to be on the same level/floor as the agent by selecting spawnable surfaces with a bounding box position within $0.5$ meters from the starting position of the agent along the vertical axis. For each environment in the training split a set of 400 episodes is sampled for each one of the possible categories.
For the generation of the validation split each target category is used twice.
Finally, episodes where the target object is not reachable by an agent following the shortest path are removed from the dataset. Refer to the Appendix for more details on dataset generation.

The resulting dataset for \shorttask is defined by a total of $865,519$ generated episodes for the training split, while the validation split contains $1,193$ episodes.
The geodesic distances of the target and distractors from the starting position of the agent in the episodes of \ours are shown in Fig.~\ref{fig:distances_plots}. In the figure, the distribution of the distances of targets and distractors significantly overlap, hence prior information on the target object distance is hardly exploitable.
\section{Baselines}
\label{sec:baselines}

\begin{figure}[!t]
    \centering
    \resizebox{.99\linewidth}{!}{
        \includegraphics[height=\linewidth]{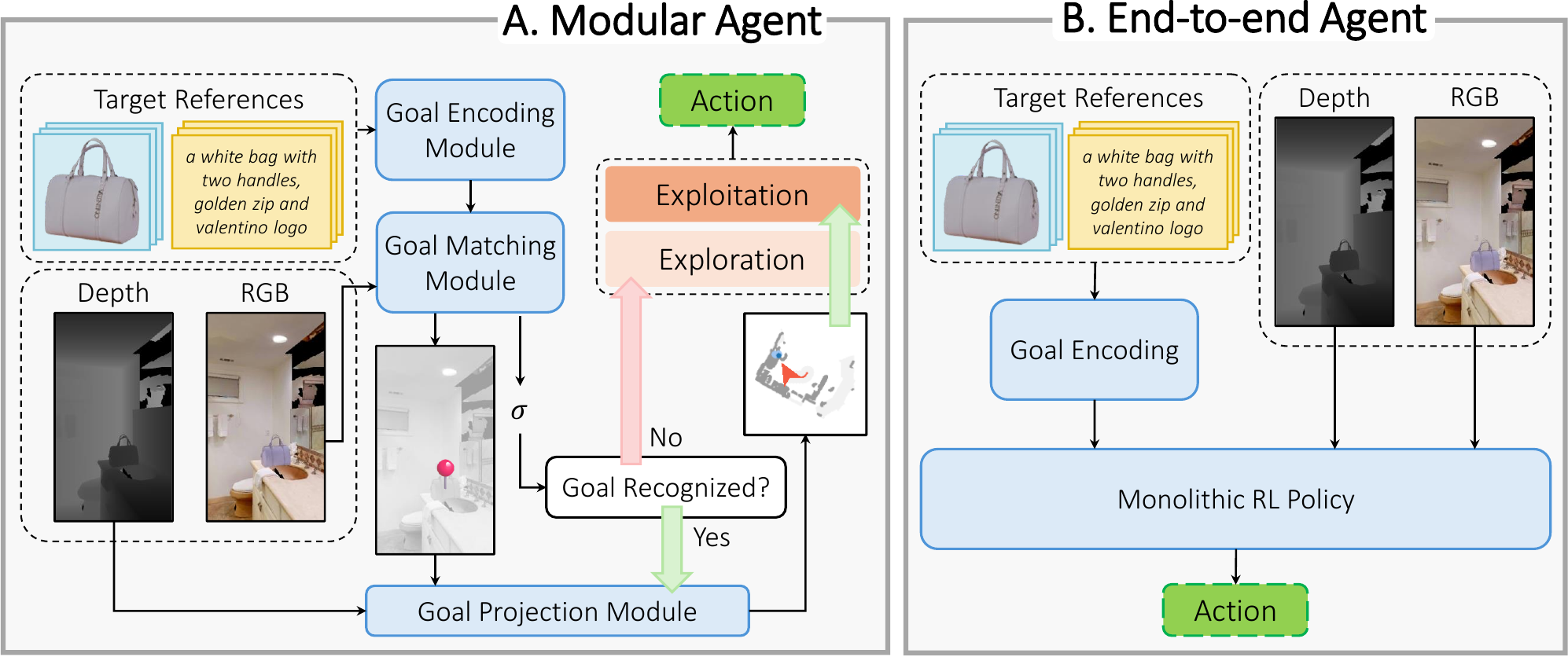}
        }
    \caption{Overview of the baselines designed for the \shorttask task: modular agent (on the left) and end-to-end agent based on a monolithic reinforcement learning-based policy (on the right).}
\vspace{-.2cm}
\label{fig:overview}
\end{figure}

In this section, we present the set of approaches that are revisited and tested on our introduced \ours dataset. These methods are recent object-driven methods and can be grouped into two categories: (i) \textbf{modular agents} that decouple the navigation task into specialized sub-modules and (ii) \textbf{end-to-end agents} based on a monolithic policy trained using reinforcement learning. Fig.~\ref{fig:overview} shows an overview of these two approaches. We refer to the Appendix for more details on the implementation of the baselines.

\subsection{Modular Agents}
\label{subsec:modular_agents}

In recent years, modular agents gathered an increasing interest in various embodied settings. These agents tackle the high-level navigation tasks by decoupling them into a chain of specialized sub-modules, each of which handles a smaller task. Specifically, Chaplot~\etal~\cite{chaplot2020object} proposed SemExp, a modular agent designed for the ObjectNav task composed of three main modules: exploration, object detection, and exploitation. The core idea is that the agent explores as much as possible the unseen environment while the detection module localizes the semantic objects in the acquired observations. Inspired by this approach, Mod-IIN~\cite{krantz2023navigating} and CLIP on Wheels (CoW)~\cite{gadre2022cow} adapt the detection module to handle specific instances and open-vocabulary targets, respectively. For our modular agent baselines, we consider the same exploration and exploitation modules used in these previous works, while changing and adapting the object detection module for the \shorttask task.

\tit{Exploration Module} The exploration module is entitled to explore the unseen areas of the environment with the scope of encountering the target object. As in Mod-IIN and CoW, we adopt a frontier-based exploration (FBE~\cite{yamauchi1997frontier}) approach. The agent builds an occupancy map of the environment during navigation, and at every time step, if the goal is not detected, the unexplored frontier on the map which is closest to the agent is selected as the current goal.

\tit{Object Detection Module} The object detection module receives the visual or textual references and the current RGB observation of the agent. Then, it is tasked with providing (i) a \textbf{matching score} that, whether it exceeds a certain matching threshold $\sigma$, determines that the goal has been recognized; and (ii) a \textbf{series of coordinates} on the observation which correspond to where the goal is located, that are used by the exploitation module to project the goal on a 2D map. We select three categories of approaches to implement this module:\vspace{-0.1cm}
\begin{itemize}[left=3mm]
    \item[\pin] \textbf{Keypoint Matching}: In this category, the visual target references and the current RGB observation are provided to a keypoint matching method. We tested SuperGlue~\cite{sarlin2020superglue}, following the approach proposed by Mod-IIN~\cite{krantz2023navigating}, and the framework introduced in IEVE~\cite{lei2024instance}. In particular, SuperGlue outputs a confidence score for each matched keypoint pair. We use the sum of these confidences as the matching score and the keypoints that exceed a given confidence threshold $\tau$ as the localization coordinates. Regarding the Exploration-Verification-Exploitation framework proposed in IEVE, we adapted some components to match the different requirements of our task. Specifically, we first collected an auxiliary dataset, which includes, for each goal in the training set, $10$ positive samples and one negative sample containing a distractor from the same category as the goal. We trained InternImage~\cite{wang2023internimage} to classify the $18$ categories of our dataset using the goal images of the training set. Instead of the InternImage segmentation model, since, to the best of our knowledge, no segmentation dataset contains all our categories, we adopted the open-vocabulary segmenter GroundedSAM~\cite{ren2024grounded}. For the image-matching step, we exploited LightGlue~\cite{lindenberger2023lightglue} on the keypoints extracted with DISK~\cite{tyszkiewicz2020disk} as in the original IEVE paper.
    \item[\pin] \textbf{Patch-level Matching}: A Vision Transformer (ViT~\cite{dosovitskiy2020vit}) encoder divides an image into patches and extracts patch-level embeddings. Hence, we extract a goal embedding from each reference and compute the cosine similarity with the patch-level feature vectors of the RGB observation. If at least a patch has a similarity that exceeds the matching threshold $\sigma$, the goal is considered detected. The center coordinates of these patches are used as the goal localization result. For the visual references, we employ DINO~\cite{caron2021emerging}, DINOv2~\cite{oquab2023dinov2}, and CLIP~\cite{radford2021learning} performing a region pooling over the reference objects to obtain goal feature vectors. For the textual references, a text-aligned multimodal encoder is required. Hence, we employ CLIP and, inspired by~\cite{gadre2022cow}, CLIP with gradient relevance~\cite{chefer2021transformer} (CLIP-Grad). We assume the mean embedding of the set of prompt templates used in CoW applied to the target descriptions as the target feature vector. 
    \item[\pin] \textbf{Detection Model}: We consider detection models that produce output regions according to a given reference. Specifically, we consider PerSAM~\cite{zhang2023personalize} (both in the standard and one-shot finetuned versions) and OWL~\cite{minderer2022simple}, which localize regions according to, respectively, visual and textual references. As in CoW, we exploit the output confidence to determine whether the goal has been detected and return the central coordinates of the region as the goal localization result.
\end{itemize}

\tit{Exploitation Module} The exploitation module takes control of the navigation when the goal is recognized in the current observation. After detecting the target object at a given location, the exploitation module is triggered and computes the route to reach the target object.
The goal position provided by the object detection module is projected into an occupancy map, and the Fast Marching Method~\cite{chaplot2020neural,sethian1996fast} is used to plan the path from the current position of the agent to the detected goal position. When the agent reaches the goal position, the `\textit{stop}' action is called to conclude the episode.

\subsection{End-to-End Agents}
\label{subsec:endtoend_agents}
In contrast to modular agents, end-to-end approaches train a neural network policy to process sensor input and predict the atomic actions needed to complete the required task. We consider two recent approaches for embodied navigation and adapt them for the \fulltask task: (i) ZSON~\cite{majumdar2022zson}, which pre-trains an ImageNav agent and evaluates downstream on ObjectNav leveraging the capabilities of CLIP multimodal embeddings; and (ii) RIM~\cite{chen2023object}, which employs a Transformer-based architecture~\cite{vaswani2017attention} that is trained using auxiliary tasks and uses a recursive implicit map that is updated during the navigation for the ObjectNav task. 
We finetune both approaches on \ours dataset. Specifically, ZSON is adapted to use image references as input during its ImageNav pretraining phase. 
While, for RIM, we employ two finetuning strategies: conditioning the navigation on textual features extracted from the reference descriptions and conditioning on visual features extracted from the image references. The features produced using both modalities of \ours references are extracted using CLIP.
\section{Experimental Evaluation}
\label{sec:experiments}

In this section, we present an experimental analysis of the selected baselines on the \shorttask task, discussing the set of metrics used to effectively evaluate the performances and the obtained results.

\begin{table}[!t]
    \centering
    \caption{Navigation results on \ours on the environments of HM3D dataset, considering the presence of distractors from the same category. \textbf{Bold} text denotes the best performance among each category of approaches.}
    \setlength{\tabcolsep}{.32em}
    \resizebox{\linewidth}{!}{
    \begin{tabular}{lcc c ccccc c ccccc}
        \toprule
        & & & & \multicolumn{5}{c}{\textbf{Navigation Metrics}} & & \multicolumn{5}{c}{\textbf{Detection Metrics}} \\
        \cmidrule{5-9} \cmidrule{11-14}
        & Backbone & Modality & & SR$\uparrow$ & SPL$\uparrow$ & CE$\downarrow$ & D2G$\downarrow$ & Steps & & \%Match$\uparrow$ & TM$\uparrow$ & CM$\downarrow$ & NM$\downarrow$ \\
        \midrule
        \rowcolor{light_gray}
        \textit{Modular Agents} & & & & & & & & & & & & & \\
        \addlinespace[1mm]
        \hspace{0.3cm}CLIP~\cite{radford2021learning} & ViT-B/16 & Textual & & 3.10 & 1.82 & 9.31 & 7.94 & 503.1 & & 62.95 & 20.07 & 22.07 & 57.86 \\
        \hspace{0.3cm}CLIP-Grad~\cite{gadre2022cow} & ViT-B/32 & Textual & & 4.53 & 2.42 & 6.95 & 7.94 & 465.8 & & 77.95 & 4.65 & 7.21 & 84.14\\
        \hspace{0.3cm}OWL~\cite{gadre2022cow,minderer2022simple} & ViT-B/32 & Textual & & 7.29 & 3.36 & 12.66 & 7.90 & 871.7 & & 22.97 & \textbf{62.60} & 32.88 & 4.52\\
        \cmidrule{1-14}
        \hspace{0.3cm}SuperGlue~\cite{krantz2023navigating,sarlin2020superglue} & - & Visual & & 3.27 & 1.28 & 7.38 & 8.36 & 804.0 & & 29.42 & 16.96 & 3.44 & 79.60 \\ 
        \hspace{0.3cm}IEVE~\cite{lei2024instance} & - & Visual & & 3.52 & 3.07 & 12.25 & 7.73 & 712.1 & & 30.03 & 32.39 & 16.01 & 51.60 \\
        \hspace{0.3cm}PerSAM~\cite{zhang2023personalize} & ViT-B/16 & Visual & & 2.77 & 1.81 & 6.53 & 8.20 & 362.5 & & \textbf{81.98} & 1.15 & 10.43 & 88.42\\
        \hspace{0.3cm}PerSAM-F~\cite{zhang2023personalize} & ViT-B/16 & Visual & & 1.93 & 1.28 & \textbf{5.63} & 8.12 & 321.3 & & 36.13 & 0.60 & 13.48 & 85.92 \\
        \hspace{0.3cm}DINO~\cite{caron2021emerging} & ViT-B/16 & Visual & & 4.02 & 1.71 & 6.88 & 8.28 & 826.0 & & 23.89 & 62.73 & \textbf{1.36} & 35.91 \\ 
        \hspace{0.3cm}CLIP~\cite{radford2021learning} & ViT-B/16 & Visual & & 9.64 & 5.39 & 13.33 & 7.79 & 623.5 & & 58.51 & 32.53 & 16.35 & 51.12 \\
        \hspace{0.3cm}DINOv2~\cite{oquab2023dinov2} & ViT-B/14 & Visual & & \textbf{14.84} & \textbf{7.94} & 26.15 & \textbf{7.28} & 658.7 && 55.74 & 55.33 & 42.00 & \textbf{2.67} \\
        \midrule
        \rowcolor{light_gray}
        \textit{End-to-end Agents} & & & & & & & & & & & & & \\
        \addlinespace[1mm]
        \hspace{0.3cm}RIM~\cite{chen2023object} & ResNet-50 & Textual & & 7.12 & 6.67 & \textbf{10.44} & 8.43 & 409.3 & & - & - & - & -\\
        \midrule
        \hspace{0.3cm}RIM~\cite{chen2023object} & ResNet-50 & Visual & & 8.80 & 6.80 & 13.41 & 8.48 & 402.1 & & - & - & - & -\\
        \hspace{0.3cm}ZSON~\cite{majumdar2022zson} & ResNet-50 & Visual & & \textbf{9.14} & \textbf{7.18} & 21.12 & \textbf{7.00} & 389.9 & & - & - & - & -\\
        \bottomrule
    \end{tabular}
    }
    \vspace{-0.2cm}
\label{tab:pin_main}
\end{table}

\subsection{Evaluation Metrics}
Traditional metrics for object-driven embodied navigation are \textbf{success rate} (SR) and \textbf{success rate weighted by path length} (SPL). SR is the ratio between the number of episodes where the agent successfully reaches the target object within a maximum distance of 1 meter and the total number of episodes, while SPL weighs the success rate with the length of the path taken by the agent. Moreover, we report the \textbf{average number of steps} taken by the agent and the \textbf{average distance from the goal} (D2G) at the end of each episode.
The agent designed for tackling the \shorttask task should be able to distinguish whether the target object is present in the current observation while exploring the unseen environment and correctly localize it, within the timesteps budget $T$ (set to 1,000). The main challenge is represented by distractor instances belonging to the same category as the target object. Hence, we introduce the \textbf{category error} (CE) metric, which measures the percentage of episodes in which the agent stopped within one meter from instances belonging to the same category of the goal.

In modular agents, the ability to detect the correct instance resides in having large matching scores when the target is present in the observation and small scores when the target is absent. Since in these agents it is possible to determine whether a given observation matches, we compute four additional metrics: the \textbf{percentage of episodes with at least a detected match} (\%Match), the \textbf{percentage of matched observations} that contain the \textbf{target object} (TM), an \textbf{instance of the same category of the target} (CM), or \textbf{no relevant objects} (NM). 

\begin{table}[!t]
    \centering
    \caption{Navigation results on \ours on the environments of HM3D dataset, without considering the presence of distractors from the same category of the target. \textbf{Bold} text denotes the best performance among each category of approaches.}
    \setlength{\tabcolsep}{.32em}
    \resizebox{0.85\linewidth}{!}{
    \begin{tabular}{lcc c cccc c ccc}
        \toprule
        & & & & \multicolumn{4}{c}{\textbf{Navigation Metrics}} & & \multicolumn{3}{c}{\textbf{Detection Metrics}} \\
        \cmidrule{5-8} \cmidrule{10-12}
        & Backbone & Modality & & SR$\uparrow$ & SPL$\uparrow$ & D2G$\downarrow$ & Steps & & \%Match$\uparrow$ & TM$\uparrow$ & NM$\downarrow$ \\
        \midrule
        \rowcolor{light_gray}
        \textit{Modular Agents} & & & & & & & & & & & \\
        \addlinespace[1mm]
        \hspace{0.3cm}CLIP~\cite{radford2021learning} & ViT-B/16 & Textual & & 3.35 & 1.86 & 8.01 & 516.5 & & \textbf{61.86} & 22.83 & 77.17 \\
        \hspace{0.3cm}OWL~\cite{gadre2022cow,minderer2022simple} & ViT-B/32 & Textual & & 8.22 & 3.18 & 7.88 & 929.9 & & 13.83 & 93.91 & 6.09 \\
        \midrule
        \hspace{0.3cm}CLIP~\cite{radford2021learning} & ViT-B/16 & Visual & & 11.15 & 5.92 & 7.65 & 666.2 & & 52.56 & 35.57 & 64.43 \\
        \hspace{0.3cm}DINOv2~\cite{oquab2023dinov2} & ViT-B/14 & Visual & & \textbf{23.13} & \textbf{11.61} & \textbf{6.62} & 784.5 & & 38.64 & \textbf{96.09} & \textbf{3.91}\\
        \midrule
        \rowcolor{light_gray}
        \textit{End-to-end Agents} & & & & & & & & & & & \\
        \addlinespace[1mm]
        \hspace{0.3cm}RIM~\cite{chen2023object} & ResNet-50 & Textual & & 7.46 & 6.87 & 7.94 & 487.1 & & - & - & - \\
        \midrule
        \hspace{0.3cm}RIM~\cite{chen2023object} & ResNet-50 & Visual & & 10.35 & 7.53 & 7.75 & 475.9 & & - & - & - \\
        \hspace{0.3cm}ZSON~\cite{majumdar2022zson} & ResNet-50 & Visual & & \textbf{10.39} & \textbf{8.00} & \textbf{6.91} & 460.1 & & - & - & - \\
        \bottomrule
    \end{tabular}
    }
\vspace{-.2cm}
\label{tab:pin_textual_without_distractors}
\end{table}

\subsection{Experimental Results}
\tinytit{\fulltask Experiments}
In Table~\ref{tab:pin_main}, we present the results on the \shorttask task. Among modular agents, DINOv2 performs best according to SR and SPL.
The high values of TM, CM, and CE show that the obtained matches usually refer to the same category of the target instance. The same reasoning can be applied to OWL for the modular agents using textual references. However, OWL produces fewer matches as can be noted from the \%Match metric. Models such as SuperGlue, PerSAM, and PerSAM-F, which exhibit low SR and TM, have also a corresponding high NM, demonstrating that they are not able to provide significant matching scores for distinguishing the correct instances or even the correct categories.
It is noteworthy that SuperGlue struggles to match the instances of \ours, which are represented on a neutral background, contrary to InstanceImageNav~\cite{krantz2023navigating}, where the reference image is a photo of the object in the same context in which it is located.
Regarding PerSAM and PerSAM-F, the results show that the feature space of SAM~\cite{kirillov2023segany} is not informative enough to understand whether an instance is present in an observation. IEVE shows an improvement with respect to the other image-matching modular agent, based on SuperGlue. This is motivated by the fact that IEVE, differently from other image-matching approaches, combines LightGlue with a semantic detector, allowing the agent to focus only on observations that contain objects of the target category. This behavior is confirmed by the increased numbers of target matches, category matches, and category errors.

Moreover, end-to-end agents tend to perform worse than modular agents. This can be attributed to the imitation training performed using the ground-truth trajectory to the goal. Since in the \shorttask task the target instances can be placed in multiple locations, it is not possible to exploit prior semantic knowledge about the estimated location of the target instance. Moreover, end-to-end agents tend to struggle in backtracking and in recovering the navigation when moving in the wrong direction. This behavior can also be noted from the path length, which for end-to-end agents is shorter than modular agents, that continue the exploration until the whole environment is observed.

\tinytit{Ablation on Category Distractors}
In Table~\ref{tab:pin_textual_without_distractors}, we introduce an ablation study in which we remove the distractors belonging to the same category of the target instance. Overall, metrics for all the agents improve because the presence of these distractors represents the core challenge of the \shorttask task. In particular, DINOv2 improves by 8.29 with respect to the main experiments, demonstrating that it embeds strong semantic correspondence properties among the same category, but that it is not trivial to identify a threshold that clearly distinguishes specific instances. The impact of same-category distractors on end-to-end agents is minor since they are finetuned to identify the correct instance.
\section{Conclusion}
\label{sec:conclusion}

In this work, we presented the task of \fulltask (\shorttask) in which the agent is required to locate and navigate toward a specific target instance. Additionally, we release \ours, a task-specific dataset built by injecting a set of additional photo-realistic objects in the scenes of HM3D. Finally, we perform an extensive analysis of recent navigation methods adapted for the proposed task. Experimental results demonstrate that the new challenges in the recognition of specific instances introduced in our proposed task are still far from being addressed.
This benchmark sets a novel testbed for future work on embodied navigation toward personalized instances.
\acksection

This work has been conducted under a research grant co-funded by Leonardo S.p.A. and supported by the EU Horizon project ``ELIAS - European Lighthouse of AI for Sustainability'' (No. 101120237), the project ``Personalized Robotics as Service Oriented Applications (PERSEO)'' funded under the Marie Sklodowska-Curie Action Horizon 2020 (No. 955778), and the PNRR project ``Fit for Medical Robotics (Fit4MedRob)'' funded by the Italian Ministry of University and Research.

\newpage
{
\small
\bibliography{main}
\bibliographystyle{plain}
}

\newpage
\setcounter{table}{0}
\setcounter{figure}{0}

\colorlet{punct}{red!60!black}
\definecolor{delim}{RGB}{20,105,176}
\colorlet{numb}{black}
\lstdefinelanguage{json}{
    basicstyle=\small\ttfamily,
    numbers=left,
    numberstyle=\scriptsize,
    stepnumber=1,
    numbersep=6pt,
    float=t,
    showstringspaces=false,
    breaklines=true,
    frame=single,
    linewidth=.97\textwidth,
    xleftmargin=0.45cm,
    backgroundcolor=\color{lighter_gray},
    literate=
     *{0}{{{\color{numb}0}}}{1}
      {1}{{{\color{numb}1}}}{1}
      {2}{{{\color{numb}2}}}{1}
      {3}{{{\color{numb}3}}}{1}
      {4}{{{\color{numb}4}}}{1}
      {5}{{{\color{numb}5}}}{1}
      {6}{{{\color{numb}6}}}{1}
      {7}{{{\color{numb}7}}}{1}
      {8}{{{\color{numb}8}}}{1}
      {9}{{{\color{numb}9}}}{1}
      {:}{{{\color{punct}{:}}}}{1}
      {,}{{{\color{punct}{,}}}}{1}
      {\{}{{{\color{delim}{\{}}}}{1}
      {\}}{{{\color{delim}{\}}}}}{1}
      {[}{{{\color{delim}{[}}}}{1}
      {]}{{{\color{delim}{]}}}}{1},
}

\title{\bigpin Personalized Instance-based Navigation Toward User-Specific Objects in Realistic Environments \\ \large \vspace{2mm}\textnormal{Supplemental Material}\vspace{-2mm}}

\author{%
  Luca~Barsellotti\thanks{Equal contribution.} \quad Roberto~Bigazzi$^{*}$ \\ \textbf{Marcella~Cornia} \quad \textbf{Lorenzo~Baraldi} \quad \textbf{Rita~Cucchiara}\\
  % Department of Engineering ``E. Ferrari''\\
  University of Modena and Reggio Emilia, Italy \\
  \texttt{\{firstname.lastname\}@unimore.it} \\
  % examples of more authors
  % \And
  % Coauthor \\
  % Affiliation \\
  % Address \\
  % \texttt{email} \\
  % \AND
  % Coauthor \\
  % Affiliation \\
  % Address \\
  % \texttt{email} \\
  % \And
  % Coauthor \\
  % Affiliation \\
  % Address \\
  % \texttt{email} \\
  % \And
  % Coauthor \\
  % Affiliation \\
  % Address \\
  % \texttt{email} \\
  \vspace{-2mm}
  \\Project page: \href{https://aimagelab.github.io/pin}{aimagelab.github.io/pin}
}

% \begin{document}

\maketitle

\appendix
\renewcommand{\thesubsection}{\thesection.\alph{subsection}}
\renewcommand{\thesection}{\Alph{section}}
\renewcommand{\thetable}{\Alph{table}}%
\renewcommand{\thefigure}{\Alph{figure}}%
\renewcommand{\thelstlisting}{\Alph{lstlisting}}

\setcounter{footnote}{0}

\vspace{-5mm}
\section{Dataset and Codebase Release}
\label{sec:code_release}
The dataset and codebase of our work are released at the following link\footnote{\href{https://github.com/aimagelab/pin}{https://github.com/aimagelab/pin}}. We provide the instructions to download the assets contained in the \ours dataset and the codebase to run the main experiments on the \fulltask (\shorttask) task.

\section{Limitations}
A limitation of this work is related to the visual appearance of some of the object instances in the \ours dataset. 
For example, the Habitat simulator's~\cite{savva2019habitat} rendering can cause a deterioration in the texture quality of some objects, failing to accurately reproduce them in the environment.
Moreover, instances with very small or detailed components can also exhibit a degradation in their visual fidelity when instantiated in the simulator. Consequently, as the agent moves farther from these objects, their details become less discernible. 
As a direct consequence, detecting small target objects is a critical challenge for navigation agents tackling the \shorttask task.

This behavior is showcased in Sec.~\ref{sec:hard_cases}, where agents tackling the \shorttask task in the episodes of \ours dataset face significant challenges in successfully detecting instances of inherently small object categories. 
In fact, despite agents such as the modular agent with DINOv2~\cite{oquab2023dinov2} showcase good performance on the overall \shorttask task, detecting small objects represents one of the main limitations of current object-driven agents, as they can only be recognized when the robot is close to them. 

A possible future improvement could involve designing novel exploration policies that aim to bring the robot closer to surfaces where the target might be placed while leveraging different detection criteria that take into consideration the scale of the observed objects.

\section{Broader Impacts}
The introduction of the \fulltask (\shorttask) task and the accompanying \ours dataset has the potential to advance the field of visual navigation and Embodied AI.
The \shorttask task fills the limitations of the current datasets for embodied navigation by requiring agents to distinguish between multiple instances of objects from the same category, thereby enhancing their precision and robustness in real-world scenarios. This advancement can lead to more capable and reliable robotic assistants and autonomous systems, especially in household settings.
Moreover, the \ours dataset serves as a comprehensive benchmark for the development and evaluation of novel algorithms in object-driven navigation. 
By providing a challenging and extensive dataset, we encourage the research community to develop innovative approaches and solutions.

\section{Additional \fulltask Details}
\label{subsec:additional_navigation_details}

\tit{Configurations}
In addition to the task definition details provided in Sec.~\ref{subsec:task_definition} of the main paper, relevant hyperparameters employed for executing each episode of the \ours dataset are presented in Table~\ref{tab:configurations}. 

\begin{table}[!t]
    \centering
    \caption{Configuration of the main parameters used for executing each episode of the \shorttask task contained in the \ours dataset.}
    \label{tab:configurations}
    \resizebox{\linewidth}{!}{
        \renewcommand{\arraystretch}{1.1} % Default value: 1
        \begin{tabular}{L{3.3cm}c c L{3.3cm}c c L{3.3cm}c}
            \toprule
            \addlinespace[2mm]
            \multicolumn{2}{l}{\textbf{\textit{Action Space }}} & & \multicolumn{2}{l}{\textbf{\textit{Episode Configuration}}} & & \multicolumn{2}{l}{\textbf{\textit{Depth Sensor}}} \\
            \hspace{0.3cm}\textit{forward step} & 0.25 & & \hspace{0.3cm}\textit{success distance} & 1.0 & & \hspace{0.3cm}\textit{width} & 360 \\
            \hspace{0.3cm}\textit{turn angle} & 30 & & \hspace{0.3cm}\textit{max episode steps} & 1000 & & \hspace{0.3cm}\textit{height} & 640 \\
            \hspace{0.3cm}\textit{tilt angle} & 30 & & \multicolumn{2}{l}{\textbf{\textit{RGB Sensor}}} & & \hspace{0.3cm}\textit{hfov} & 42 \\
            \multicolumn{2}{l}{\textbf{\textit{Agent Configuration}}} & & \hspace{0.3cm}\textit{width} & 360 & & \hspace{0.3cm}\textit{position} & [0, 1.31, 0] \\
            \hspace{0.3cm}\textit{visual sensors} & rgb, depth & & \hspace{0.3cm}\textit{height} & 640 & & \hspace{0.3cm}\textit{min depth} & 0.5 \\
            \hspace{0.3cm}\textit{height} & 1.41 & & \hspace{0.3cm}\textit{hfov} & 42 & & \hspace{0.3cm}\textit{max depth} & 5.0 \\
            \hspace{0.3cm}\textit{radius} & 0.17 & & & & & \\
            \hspace{0.3cm}\textit{position} & [0, 1.31, 0] & & & & & \\
            \addlinespace[1mm]
            \bottomrule
        \end{tabular}
    }
    \vspace{-0.2cm}
\end{table}

\begin{figure}[!t]
    \centering
    \normalsize
    \resizebox{\linewidth}{!}{
        \setlength{\tabcolsep}{.06em}
        \begin{tabular}{c p{0mm} c p{0mm} c p{2mm}p{2mm} c p{2mm}p{2mm} c p{2mm}p{2mm} c}
            \addlinespace[2mm]
            % Row 1
            \includegraphics[height=0.3\textwidth]{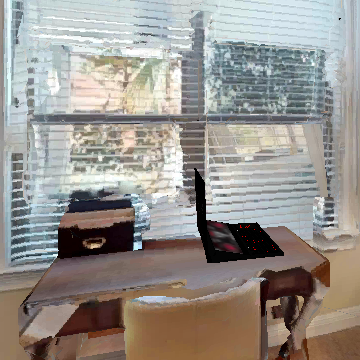} & &
            \includegraphics[height=0.3\textwidth]{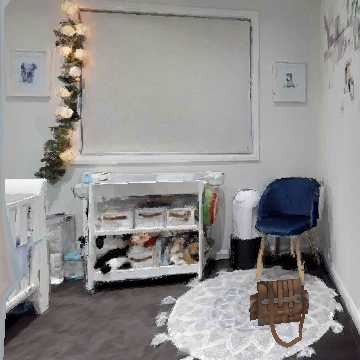} & &
            \includegraphics[height=0.3\textwidth,width=0.3\textwidth]{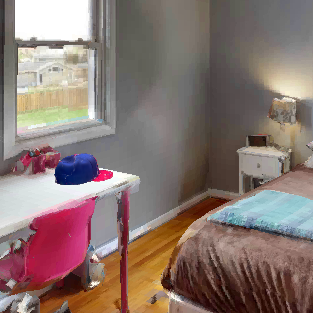} & & &
            \includegraphics[height=0.3\textwidth]{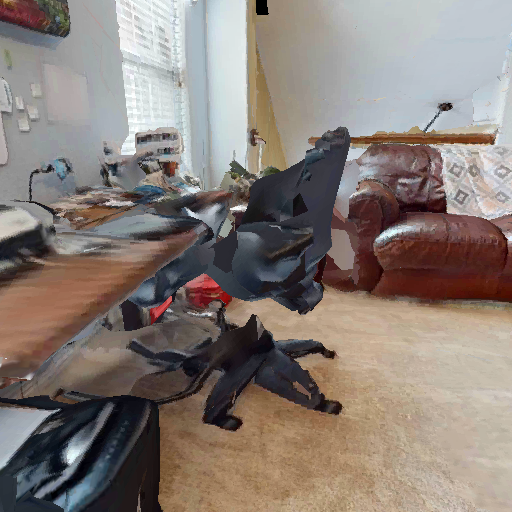} &&&
            \includegraphics[height=0.3\textwidth]{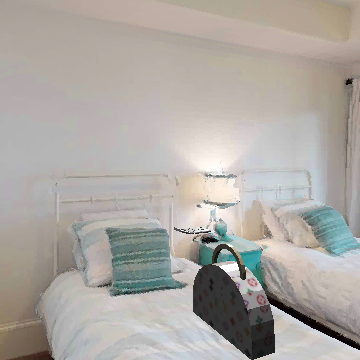} &&& \includegraphics[height=0.3\textwidth,width=0.3\textwidth]{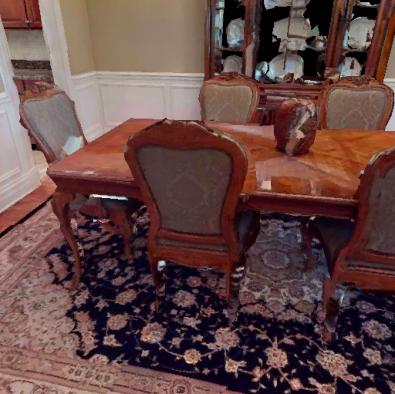} \\
            % Row 2
            \includegraphics[height=0.3\textwidth]{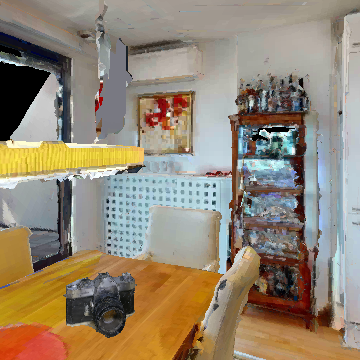} &&
            \includegraphics[height=0.3\textwidth,width=0.3\textwidth]{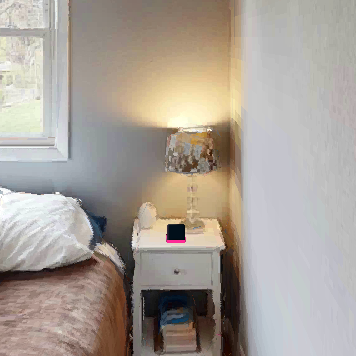} && \includegraphics[height=0.3\textwidth,width=0.3\textwidth]{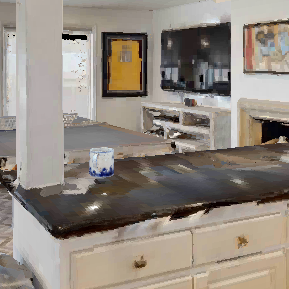} & & &
            \includegraphics[height=0.3\textwidth]{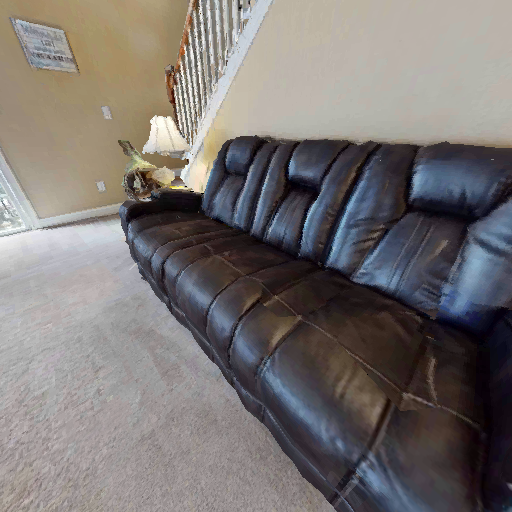} &&&
            \includegraphics[height=0.3\textwidth]{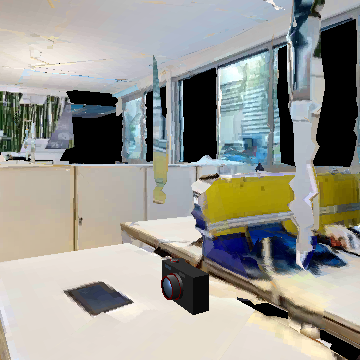}  &&& \includegraphics[height=0.3\textwidth,width=0.3\textwidth]{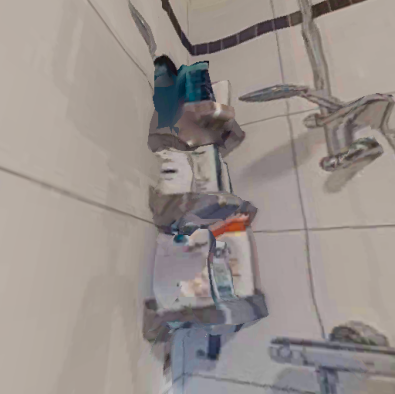} \\
            % Row 3
            \includegraphics[height=0.3\textwidth]{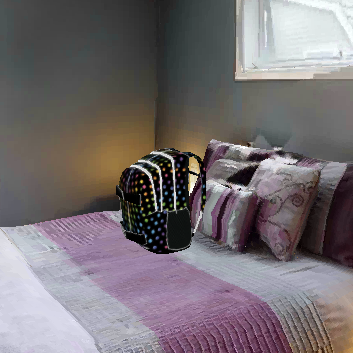} && \includegraphics[height=0.3\textwidth,width=0.3\textwidth]{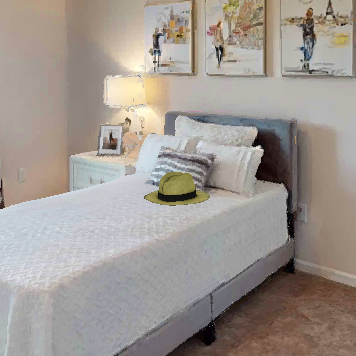} && \includegraphics[height=0.3\textwidth,width=0.3\textwidth]{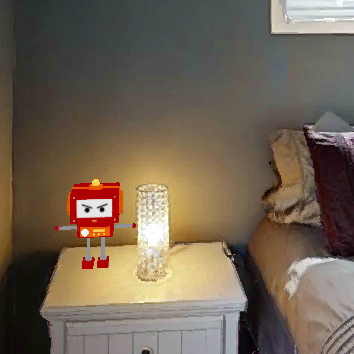} & & &
            \includegraphics[height=0.3\textwidth]{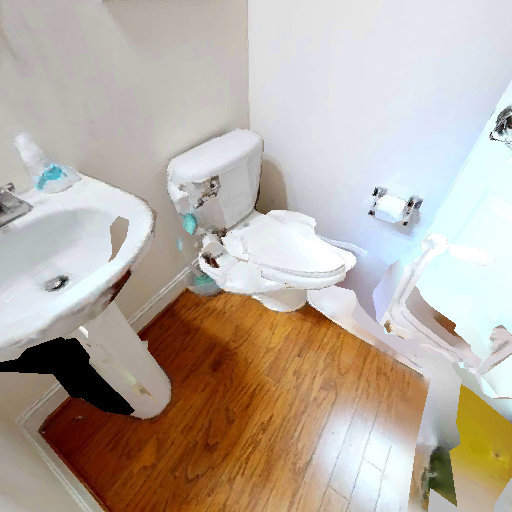} &&&
            \includegraphics[height=0.3\textwidth]{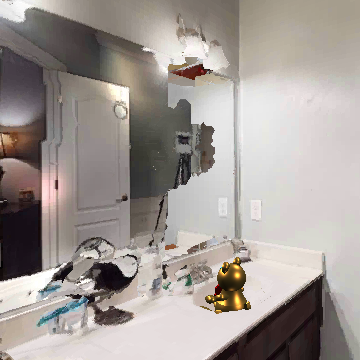}  &&& \includegraphics[height=0.3\textwidth,width=0.3\textwidth]{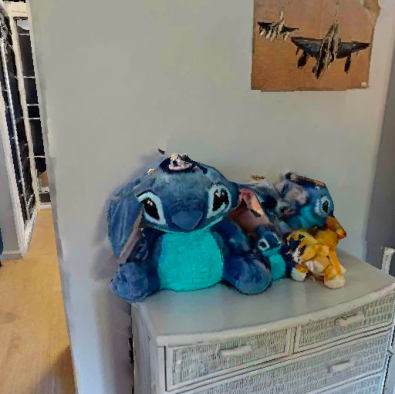} \\
            \addlinespace[2mm]
            \multicolumn{5}{c}{\Large\textbf{\ours (Ours)}} &&& \textbf{\Large InstanceImageNav} &&& \textbf{\Large MultiON} &&& \textbf{\Large GOAT-Bench} \\
        \end{tabular}
    }
    \caption{Comparison of observations depicting different target objects of \ours dataset with the target objects of InstanceImageNav, MultiON, and GOAT-Bench datasets.}
    \vspace{-2mm}
    \label{fig:vs_instance_supp}
\end{figure}

The configuration used for a \shorttask episode comprises a maximum duration of $1,000$ time steps, with the agent's action space defined by discrete forward steps of $0.25$ m, a turn angle of $30$\textdegree, and a head tilt angle of $30$\textdegree. 
Each episode is considered successful if the position of the agent is within $1$ meter from the position of the target object, and it predicts the `\textit{stop}' action before the end of the time step budget. 
The configurations used for the navigation experiments reflect the settings employed to simulate the camera sensors and space occupation of the HelloRobot Stretch\footnote{\href{https://hello-robot.com/stretch}{https://hello-robot.com/stretch}} platform.

\tit{Comparison with Object-oriented Tasks}
In addition to Fig.~\ref{fig:vs_instance} of the main paper, in Fig.~\ref{fig:vs_instance_supp} we showcase additional examples of goal objects captured in the embodied setting for different object-driven datasets. The target objects belonging to the \ours dataset are compared with InstanceImageNav~\cite{krantz2022instance}, MultiON~\cite{wani2020multion}, and GOAT-Bench~\cite{khanna2024goat} datasets. It is noticeable that injecting photo-realistic objects allows to have targets that do not present artifacts or reconstruction errors, which is common for InstanceImageNav and GOAT-Bench target objects. Furthermore, when comparing the target objects of \ours with those in the MultiON dataset, it is noticeable that the \ours objects exhibit a more photo-realistic visual quality.

\tit{Comparison with ProcTHOR}
ProcTHOR~\cite{deitke2022procthor} is a framework built on AI2-THOR~\cite{kolve2017ai2} to procedurally generate interactive environments, enabling the evaluation of data augmentation and large-scale training in different Embodied AI tasks. \ours is a dataset designed specifically to study the newly introduced \shorttask task, in which the agent is tasked with finding a specific instance according to target images or textual descriptions.

ProcTHOR includes 1,633 instances across 108 object categories, with the ability to vary brightness, colors, materials, and object states. These categories include several household objects, covering generic objects, such as `\textit{pen}' or `\textit{apple}', objects that can be personal, such as `\textit{mug}' and `\textit{watch}', and large objects that are unlikely to change their placement in the environment, such as `\textit{fridge}', and `\textit{window}'. \ours presents 18 object categories that can be personal, with the specific purpose of accompanying the task in which the agent has to distinguish instances belonging to the same category. All the categories represent objects that can be moved frequently in the environment and do not have a predefined location.

As well as most procedural datasets, ProcTHOR sacrifices realism in favor of interactivity, scalability, and customizability. \ours, as a task-specific dataset, favors photo-realistic environments and objects. Indeed, it is the first instance-based navigation dataset based on both photo-realistic environments and injected objects, that can be moved frequently and with multimodal targets. Interactivity with the objects is out of scope for this work, however, the addition of external objects paves the way for possible future enhancements where interactivity is needed.

\section{Additional \ours Dataset Details}
\label{subsec:additional_dataset_details}

\begin{figure}[!t]
    \centering
    \normalsize
    \resizebox{.95\linewidth}{!}{
        \setlength{\tabcolsep}{.05em}
        \begin{tabular}{c p{1mm} c p{1mm} c p{10mm} c p{1mm} c p{1mm} c p{10mm} c p{1mm} c p{1mm} c}
            % Row 1
            \includegraphics[height=0.3\textwidth]{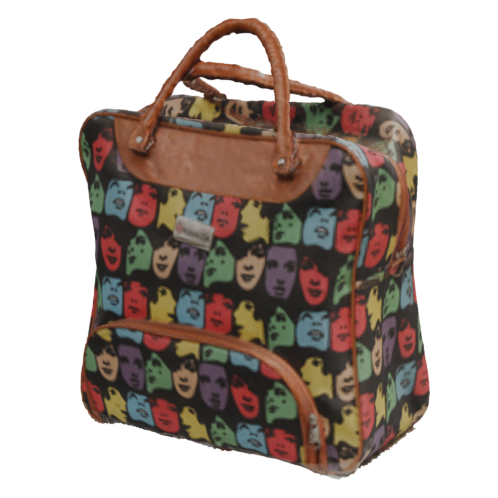} &&
            \includegraphics[height=0.3\textwidth]{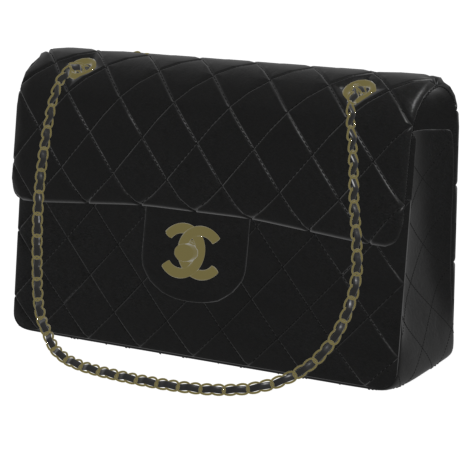} &&
            \includegraphics[height=0.3\textwidth]{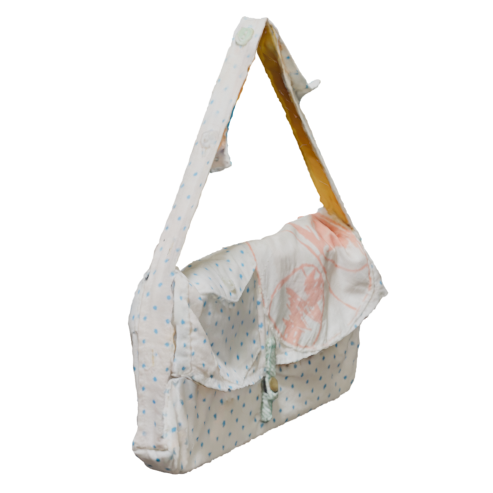} &&
            \includegraphics[height=0.3\textwidth]{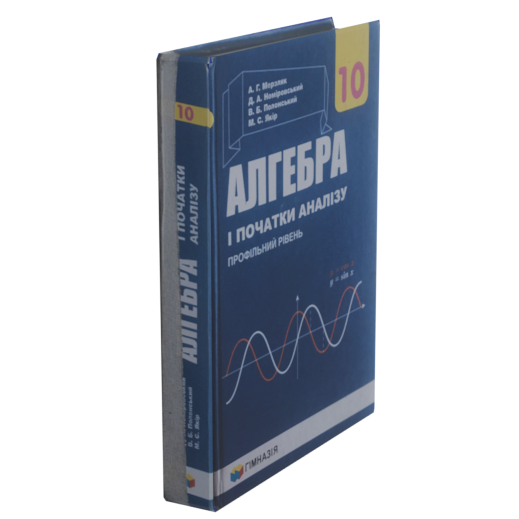} &&
            \includegraphics[height=0.3\textwidth]{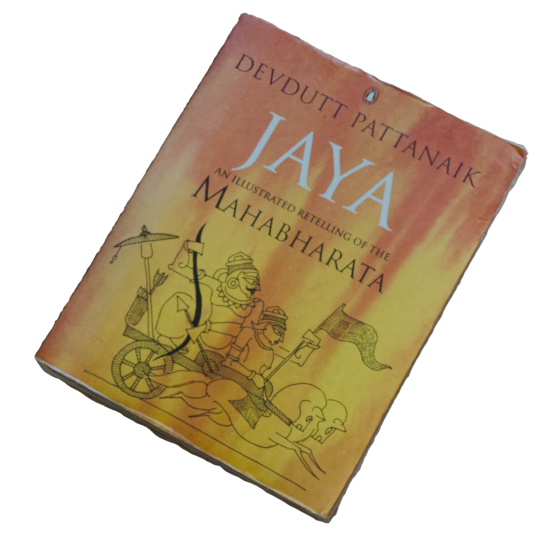} &&
            \includegraphics[height=0.3\textwidth]{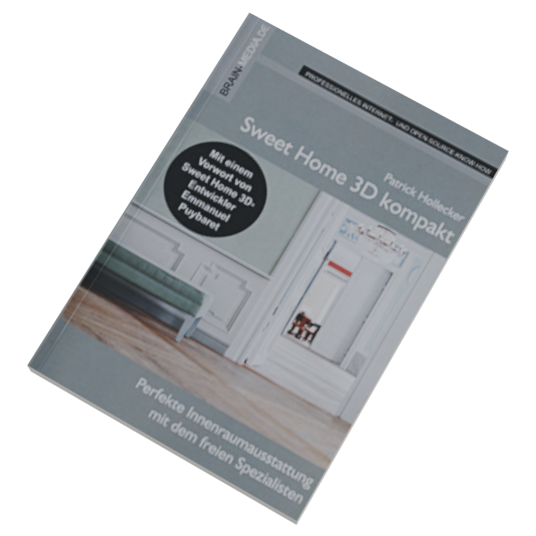} &&
            \includegraphics[height=0.3\textwidth]{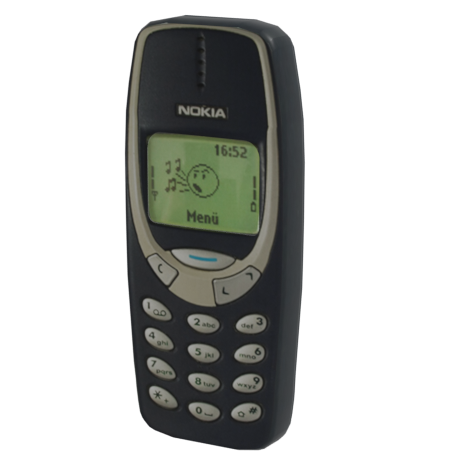} &&
            \includegraphics[height=0.3\textwidth]{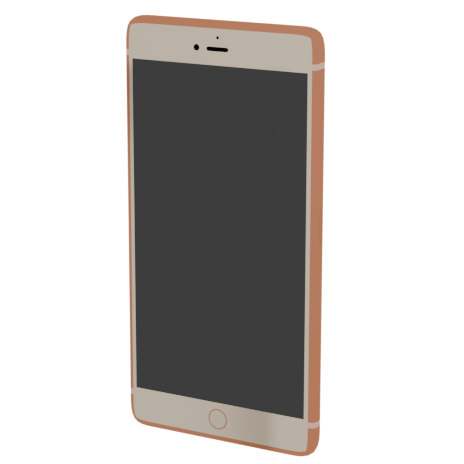} &&
            \includegraphics[height=0.3\textwidth]{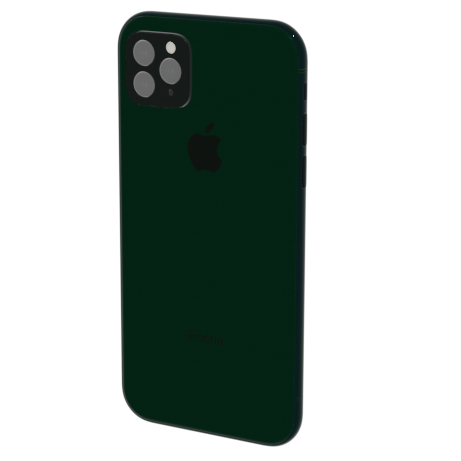}
            \\
            \addlinespace[6mm]
            \multicolumn{5}{c}{\Huge\textbf{Bag}} && \multicolumn{5}{c}{\Huge\textbf{Book}} &&
            \multicolumn{5}{c}{\Huge\textbf{Cellphone}} \\
            \addlinespace[12mm]
            \includegraphics[height=0.3\textwidth]{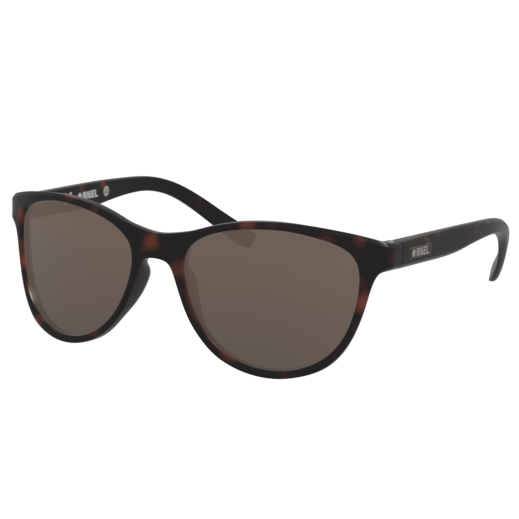} &&
            \includegraphics[height=0.3\textwidth]{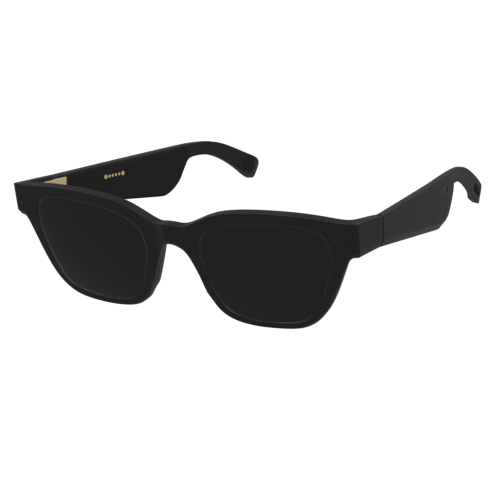} &&
            \includegraphics[height=0.3\textwidth]{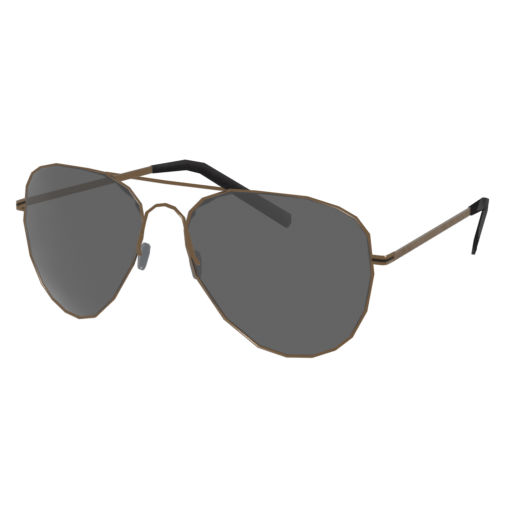} &&
            \includegraphics[height=0.3\textwidth]{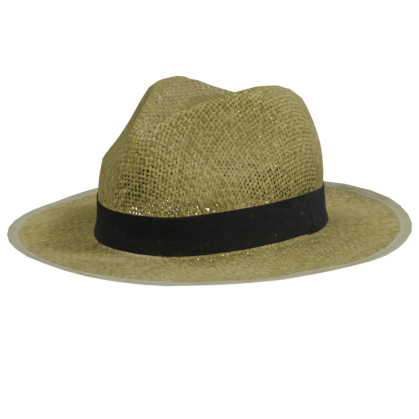} &&
            \includegraphics[height=0.3\textwidth]{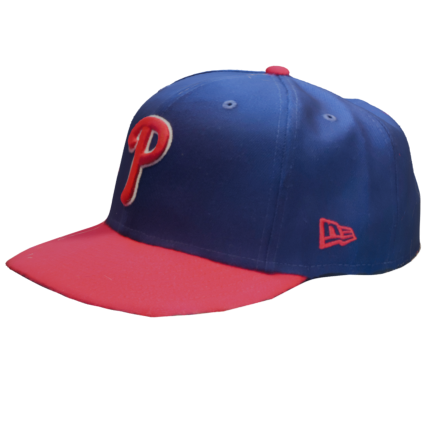} &&
            \includegraphics[height=0.3\textwidth]{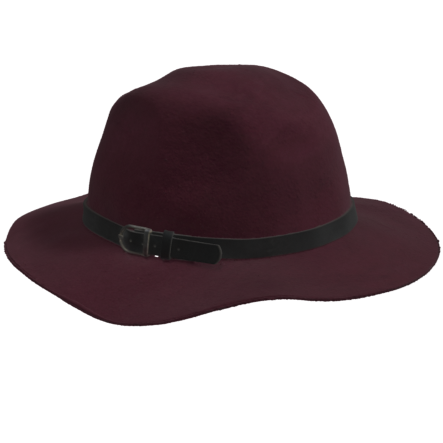} &&
            \includegraphics[height=0.3\textwidth]{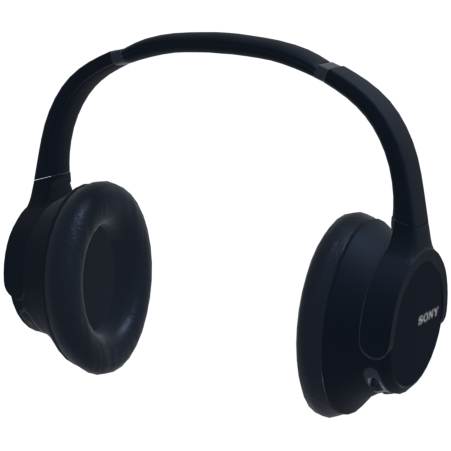} &&
            \includegraphics[height=0.3\textwidth]{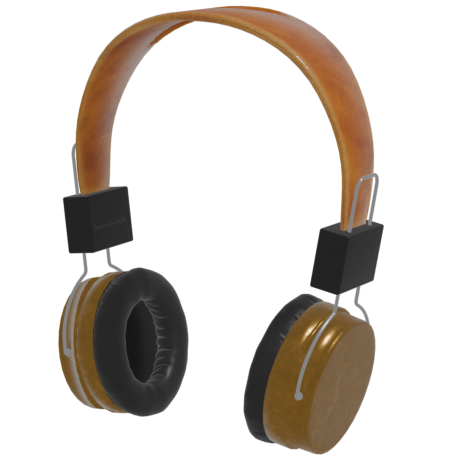} &&
            \includegraphics[height=0.3\textwidth]{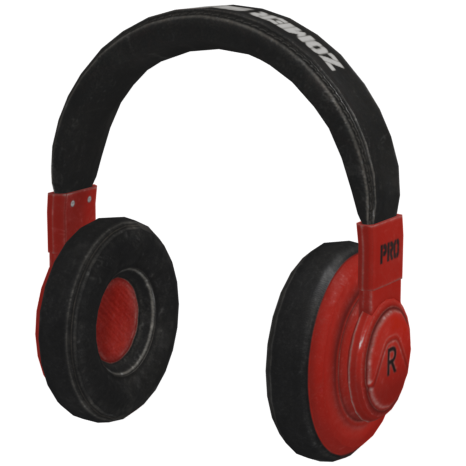}
            \\
            \addlinespace[4mm]
            \multicolumn{5}{c}{\Huge\textbf{Eyeglasses}} && \multicolumn{5}{c}{\Huge\textbf{Hat}} &&
            \multicolumn{5}{c}{\Huge\textbf{Headphones}} \\
            \addlinespace[12mm]
            \includegraphics[height=0.3\textwidth]{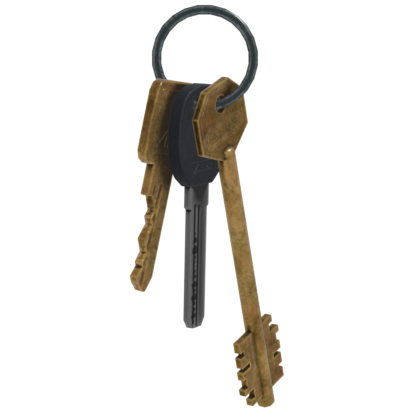} &&
            \includegraphics[height=0.3\textwidth]{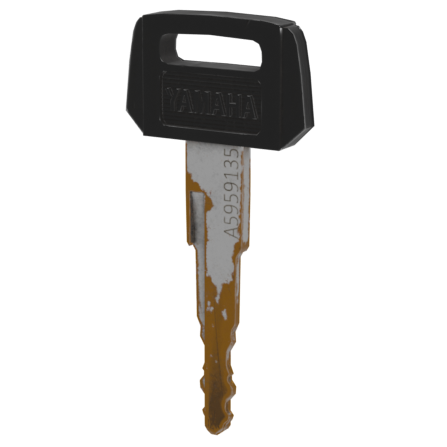} &&
            \includegraphics[height=0.3\textwidth]{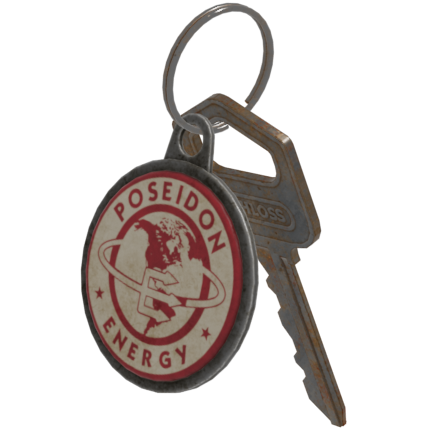} &&
            \includegraphics[height=0.3\textwidth]{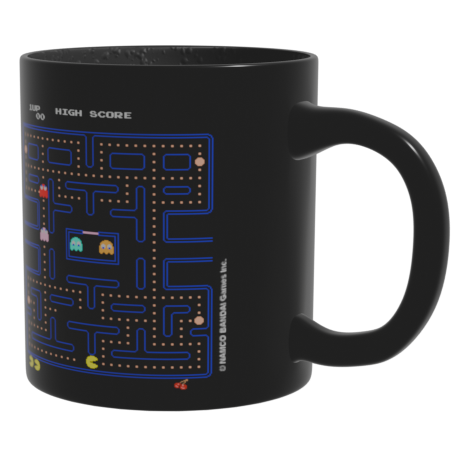} &&
            \includegraphics[height=0.3\textwidth]{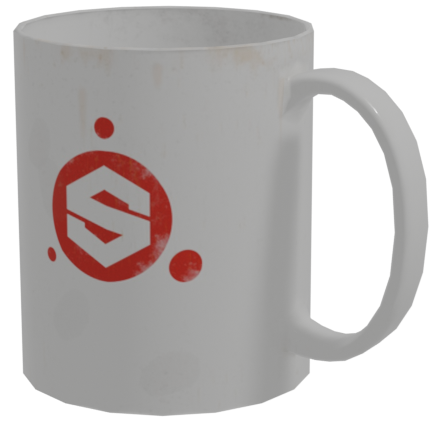} &&
            \includegraphics[height=0.3\textwidth]{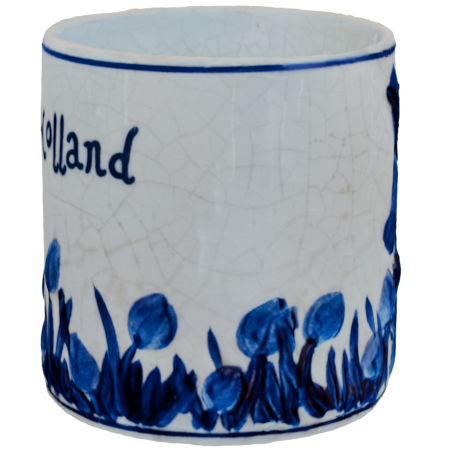} &&
            \includegraphics[height=0.3\textwidth]{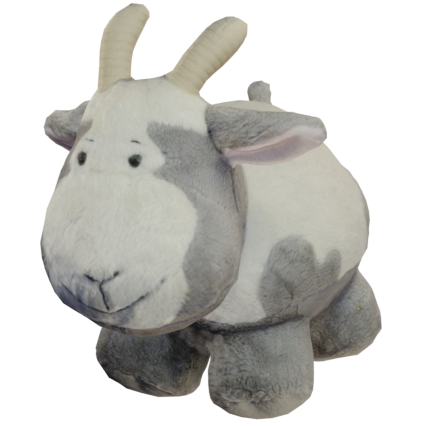} &&
            \includegraphics[height=0.3\textwidth]{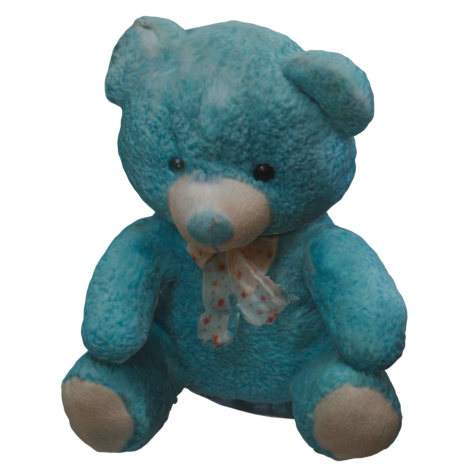} &&
            \includegraphics[height=0.3\textwidth]{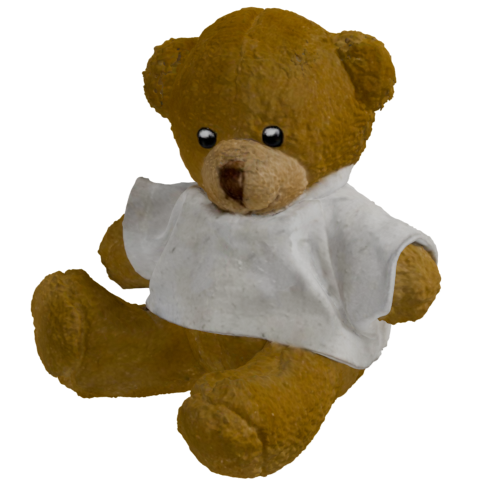}
            \\
            \addlinespace[4mm]
            \multicolumn{5}{c}{\Huge\textbf{Keys}} && \multicolumn{5}{c}{\Huge\textbf{Mug}} &&
            \multicolumn{5}{c}{\Huge\textbf{Teddy Bear}} \\
            \addlinespace[12mm]
            \includegraphics[height=0.3\textwidth]{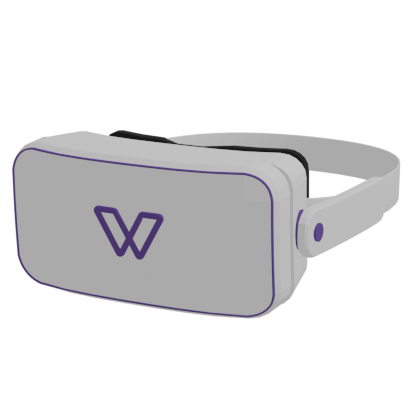} &&
            \includegraphics[height=0.3\textwidth]{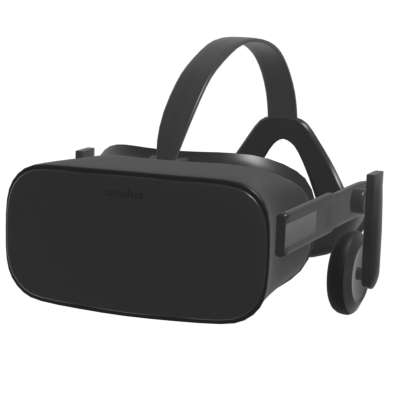} &&
            \includegraphics[height=0.3\textwidth]{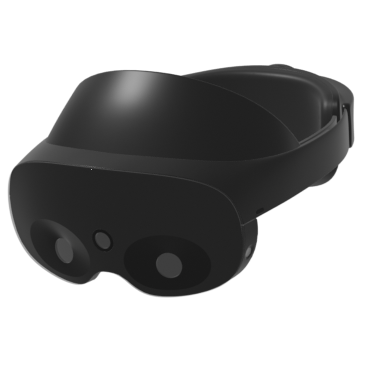} &&
            \includegraphics[height=0.3\textwidth]{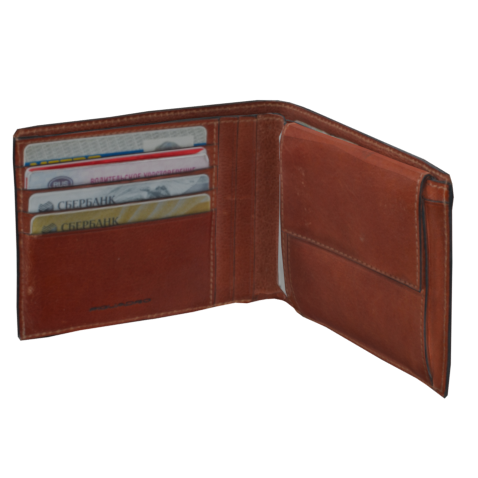} &&
            \includegraphics[height=0.3\textwidth]{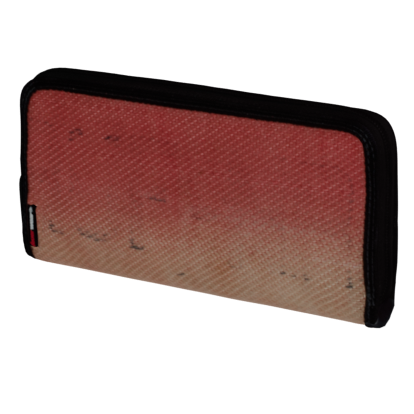} &&
            \includegraphics[height=0.3\textwidth]{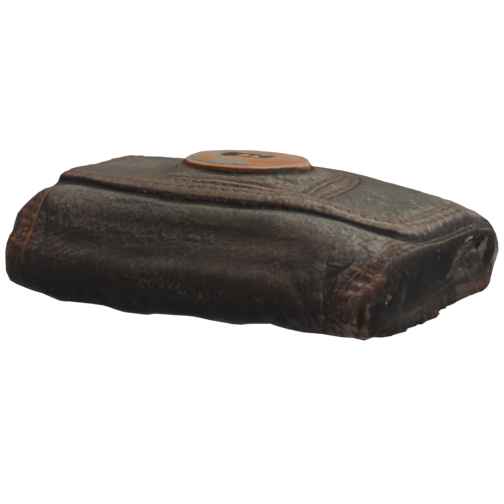} &&
            \includegraphics[height=0.3\textwidth]{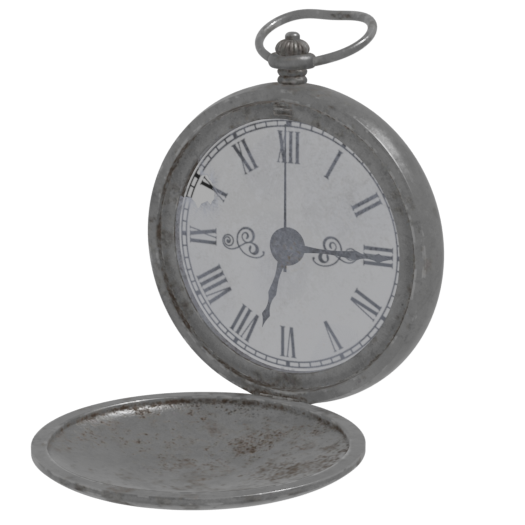} &&
            \includegraphics[height=0.3\textwidth]{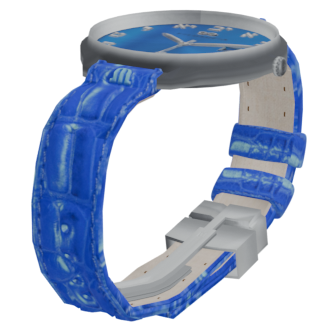} &&
            \includegraphics[height=0.3\textwidth]{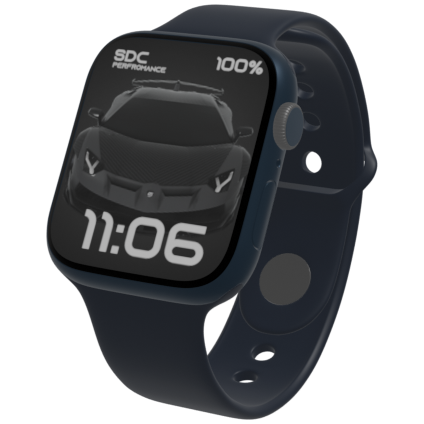}
            \\
            \addlinespace[4mm]
            \multicolumn{5}{c}{\Huge\textbf{Visor}} && \multicolumn{5}{c}{\Huge\textbf{Wallet}} &&
            \multicolumn{5}{c}{\Huge\textbf{Watch}} \\
            \addlinespace[4mm]
        \end{tabular}
    }
    \caption{Sample frontal visual references of personalized targets from \ours dataset. We include three instances for each object category, considering the categories not included in Fig.~\ref{fig:dataset_samples} of the main paper.}
    \label{fig:dataset_samples_suppl}
    \vspace{-.2cm}
\end{figure}

\tit{Additional Reference Samples} To better visualize the content of \ours dataset, in Fig.~\ref{fig:dataset_samples_suppl} we illustrate additional samples of the acquired visual references for the categories that are not included in Fig.~\ref{fig:dataset_samples} of the main paper. 

\begin{figure}[!t]
    \centering
    \small
    \resizebox{\linewidth}{!}{
        \setlength{\tabcolsep}{.1em}
        \begin{tabular}{lll p{0.4cm} L{9.4cm}}

            \toprule
            \addlinespace[1mm]

            \multirow{3}{*}{\includegraphics[height=0.105\textwidth]{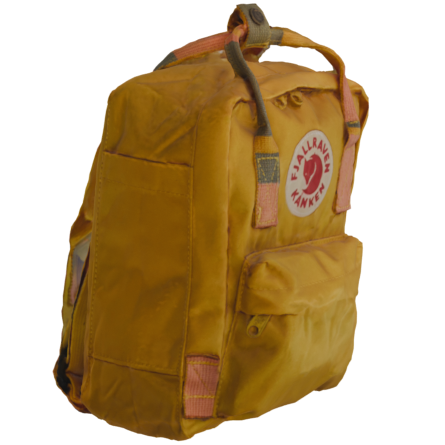}} &
            \multirow{3}{*}{\includegraphics[height=0.105\textwidth]{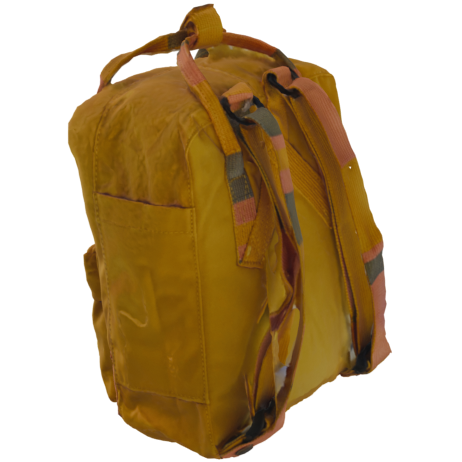}} &
            \multirow{3}{*}{\includegraphics[height=0.105\textwidth]{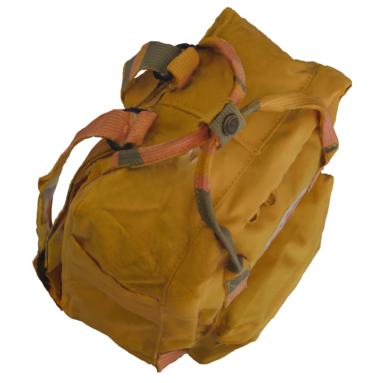}} 
            && \textit{a yellow kanken backpack with yellow straps on the top} \\
            \addlinespace[2mm]
            &&&& \textit{a yellow monochrome kanken backpack} \\
            \addlinespace[2mm]
            &&&& \textit{a photo of a yellow backpack with a strap and red circle on the front} \\

            \addlinespace[1mm]
            \midrule
            \addlinespace[2mm]

            \multirow{3}{*}{\includegraphics[height=0.105\textwidth]{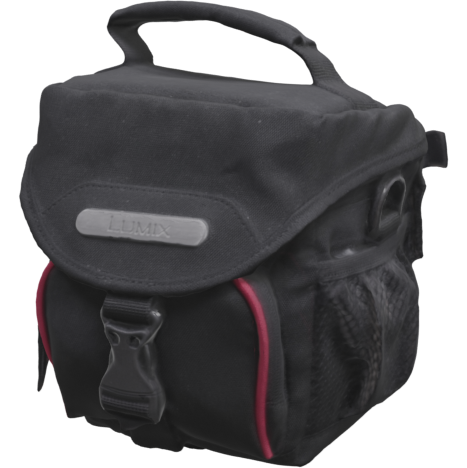}} &
            \multirow{3}{*}{\includegraphics[height=0.105\textwidth]{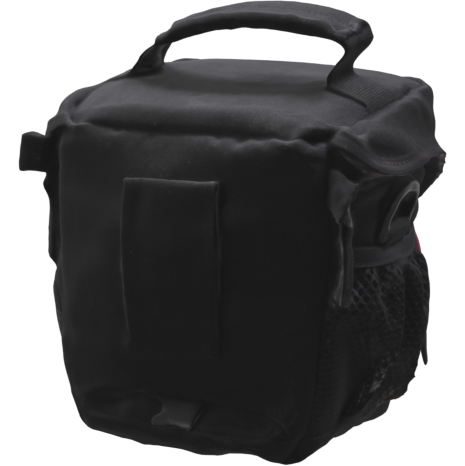}} &
            \multirow{3}{*}{\includegraphics[height=0.105\textwidth]{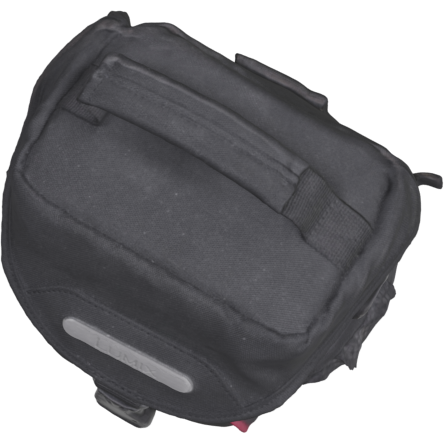}}
            && \textit{a black camera bag with a handle and a mesh pocket} \\
            \addlinespace[2mm]
            &&&& \textit{a black camera bag with a buckle and a small silver plate} \\
            \addlinespace[2mm]
            &&&& \textit{a black camera bag with two red laces, a silver plate and a black buckle in the middle front} \\

            \addlinespace[1mm]
            \midrule
            \addlinespace[2mm]
            
            \multirow{3}{*}{\includegraphics[height=0.105\textwidth]{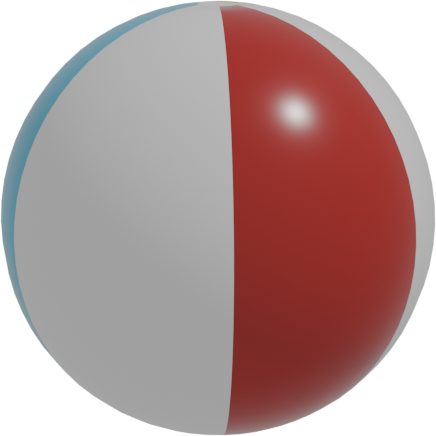}} &
            \multirow{3}{*}{\includegraphics[height=0.105\textwidth]{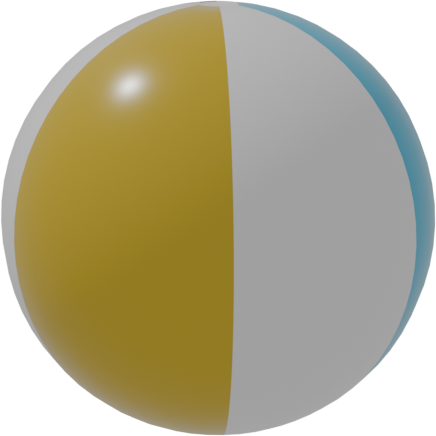}} &
            \multirow{3}{*}{\includegraphics[height=0.105\textwidth]{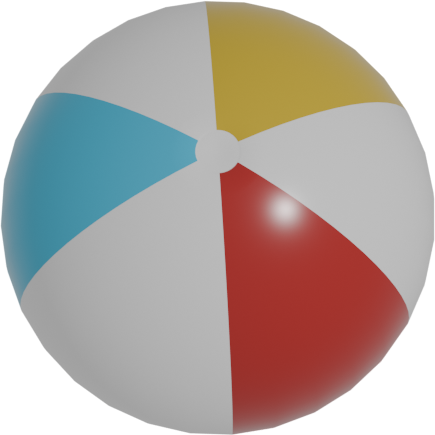}} 
            && \textit{a beach ball with alternated red, light blue and white slices} \\
            \addlinespace[2mm]
            &&&& \textit{an inflatable colored beach ball} \\
            \addlinespace[2mm]
            &&&& \textit{a beach ball with a multicolored design} \\
            
            \addlinespace[1mm]
            \midrule
            \addlinespace[2mm]
            
            \multirow{3}{*}{\includegraphics[height=0.105\textwidth]{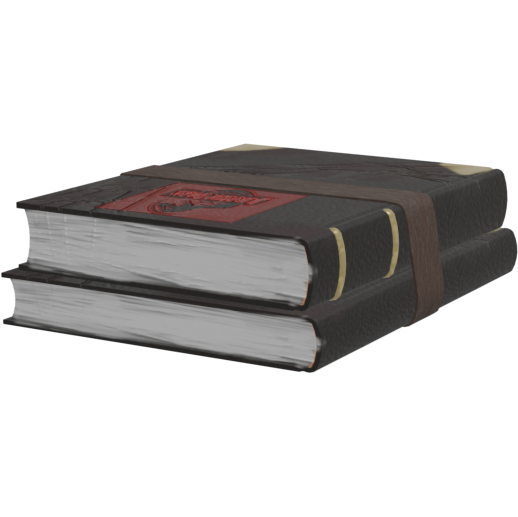}} &
            \multirow{3}{*}{\includegraphics[height=0.105\textwidth]{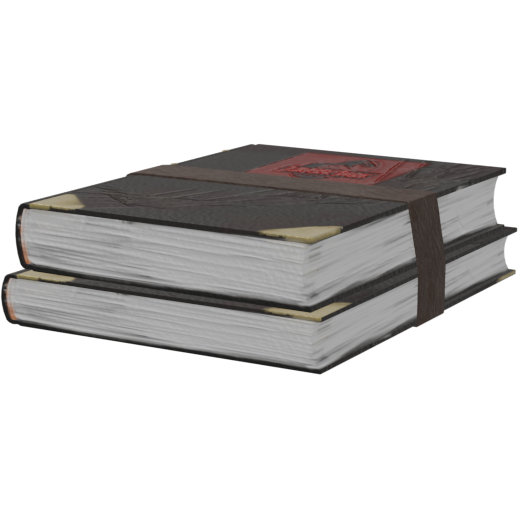}} &
            \multirow{3}{*}{\includegraphics[height=0.105\textwidth]{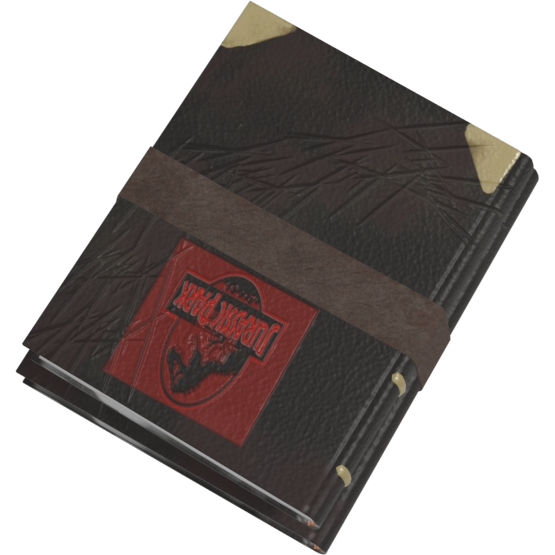}}
            && \textit{a stack of two books with a leather cover tied using a brown strap} \\
            \addlinespace[2mm]
            &&&& \textit{a brown book with yellowed pages with two straps and golden buckles on top} \\
            \addlinespace[2mm]
            &&&& \textit{two books tied together by a brown lace, with black leather covers and a red jurassic park logo} \\

            \addlinespace[1mm]
            \midrule
            \addlinespace[2mm]
            
            \multirow{3}{*}{\includegraphics[height=0.105\textwidth]{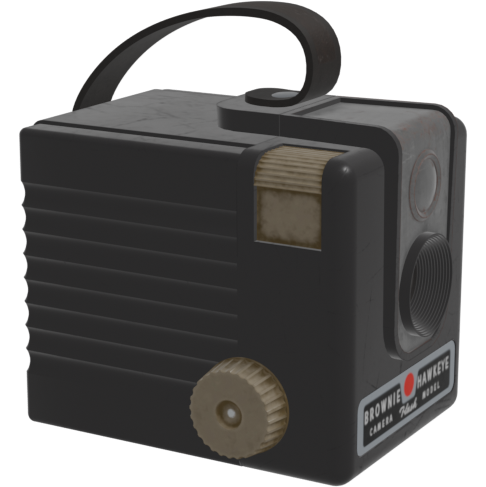}} &
            \multirow{3}{*}{\includegraphics[height=0.105\textwidth]{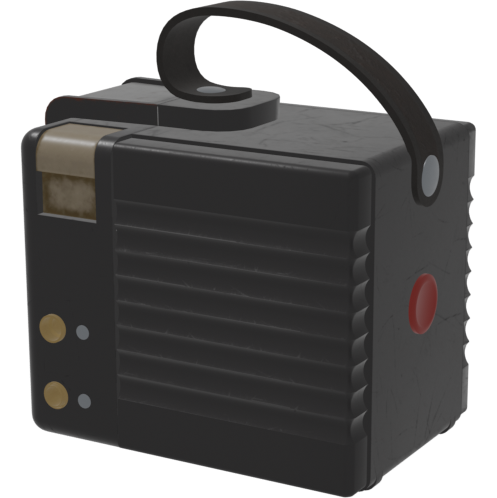}} &
            \multirow{3}{*}{\includegraphics[height=0.105\textwidth]{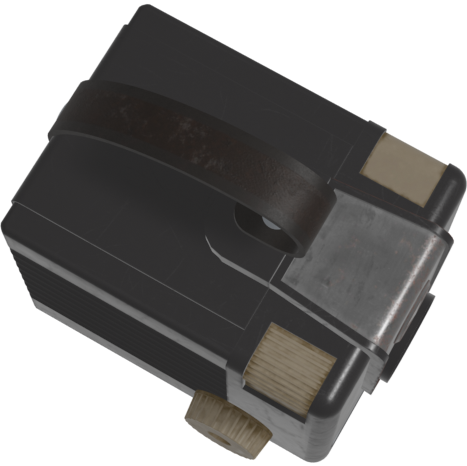}} 
            && \textit{a big black camera with a black handle and a wheel on the side} \\
            \addlinespace[2mm]
            &&&& \textit{a black cubic camera with a brown knob and a strap} \\
            \addlinespace[2mm]
            &&&& \textit{a kodak brownie hawkeye black flash camera, which is cube-shaped and has a black handle} \\
    
            \addlinespace[1mm]
            \midrule
            \addlinespace[2mm]
            
            \multirow{3}{*}{\includegraphics[height=0.105\textwidth]{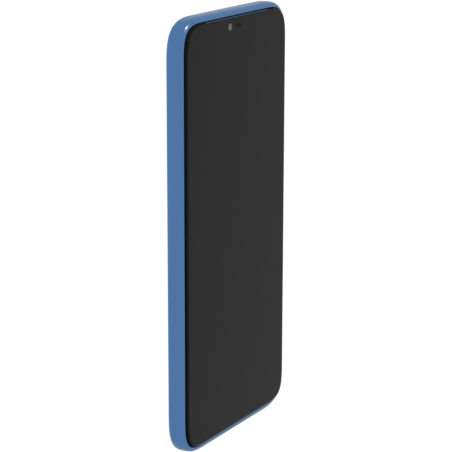}} &
            \multirow{3}{*}{\includegraphics[height=0.105\textwidth]{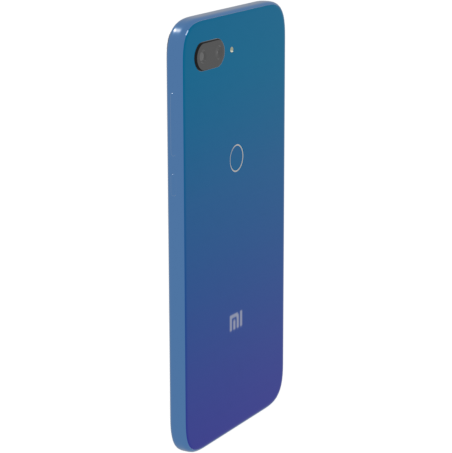}} &
            \multirow{3}{*}{\includegraphics[height=0.105\textwidth]{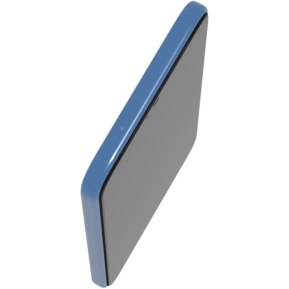}} 
            && \textit{a blue smartphone with a white text on the back} \\
            \addlinespace[2mm]
            &&&& \textit{a blue phone with a black screen} \\
            \addlinespace[2mm]
            &&&& \textit{a cellphone with a gradient blue to purple color, two lenses, a fingerprint reader and the xiaomi mi logo on the back} \\
            
            \addlinespace[1mm]
            \midrule
            \addlinespace[2mm]
            
            \multirow{3}{*}{\includegraphics[height=0.105\textwidth]{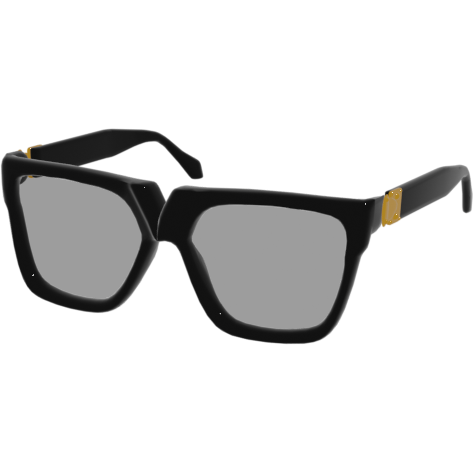}} &
            \multirow{3}{*}{\includegraphics[height=0.105\textwidth]{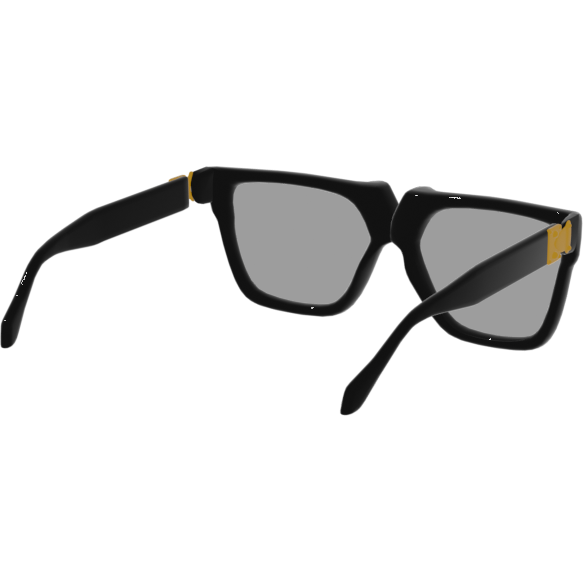}} &
            \multirow{3}{*}{\includegraphics[height=0.105\textwidth]{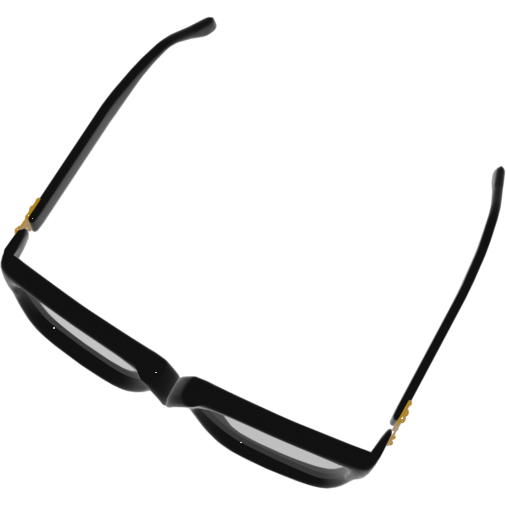}} 
            && \textit{a pair of black squared eyeglasses with a golden plate on the arms} \\
            \addlinespace[2mm]
            &&&& \textit{a pair of sunglasses with a black frame and gold detail} \\
            \addlinespace[2mm]
            &&&& \textit{a pair of black thick eyeglasses with squared frame and golden hinges} \\
            
            \addlinespace[1mm]
            \midrule
            \addlinespace[2mm]
            
            \multirow{3}{*}{\includegraphics[height=0.105\textwidth]{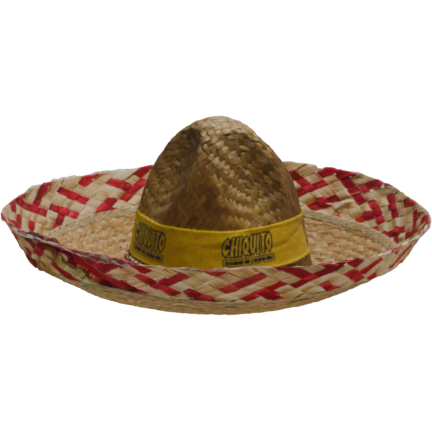}} &
            \multirow{3}{*}{\includegraphics[height=0.105\textwidth]{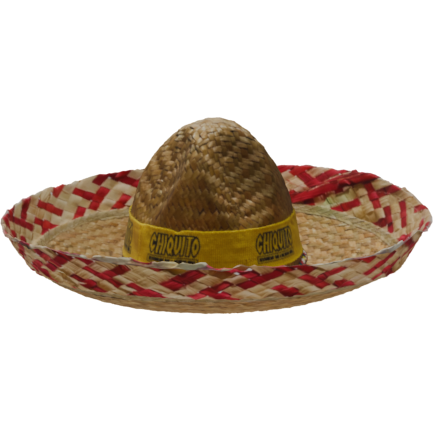}} &
            \multirow{3}{*}{\includegraphics[height=0.105\textwidth]{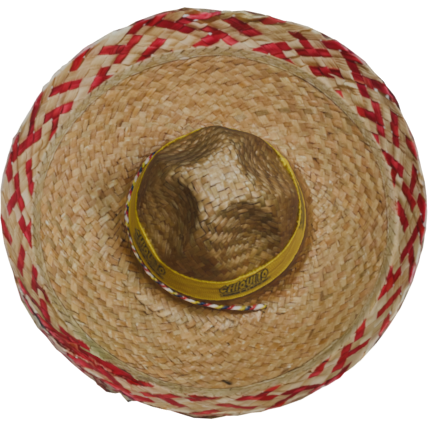}} 
            && \textit{a sombrero with red details and a yellow stripe} \\
            \addlinespace[2mm]
            &&&& \textit{a straw hat with a yellow ribbon around it } \\
            \addlinespace[2mm]
            &&&& \textit{a sombrero with red elements on the brim and a yellow stripe with chiquito written multiple times on top} \\

            \addlinespace[1mm]
            \midrule
            \addlinespace[2mm]
            
            \multirow{3}{*}{\includegraphics[height=0.105\textwidth]{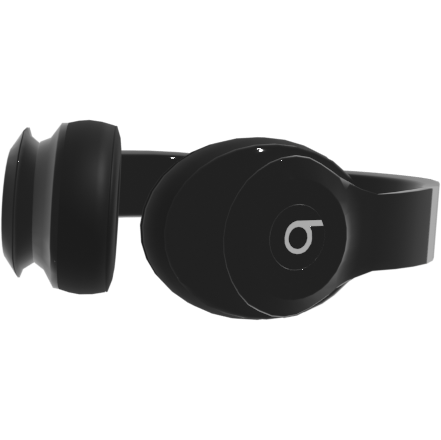}} &
            \multirow{3}{*}{\includegraphics[height=0.105\textwidth]{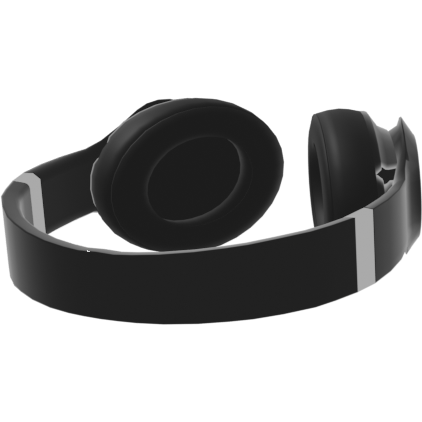}} &
            \multirow{3}{*}{\includegraphics[height=0.105\textwidth]{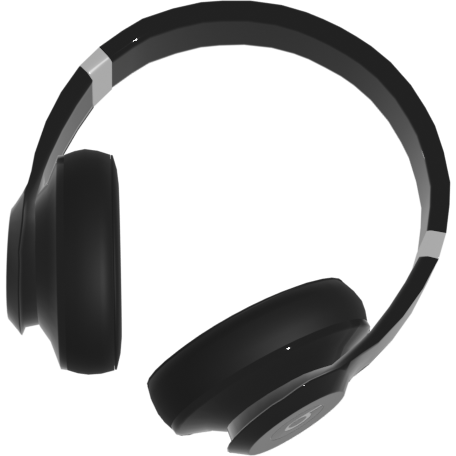}} 
            && \textit{a pair of black beats headphones} \\
            \addlinespace[2mm]
            &&&& \textit{a pair of headphones with a black band} \\
            \addlinespace[2mm]
            &&&& \textit{a pair of black headphones with the beats by dre logo on the ear cups and two gray lines on the headband} \\

            \addlinespace[1mm]
            \bottomrule
        \end{tabular}
    }
    \vspace{-.1cm}
    \caption{Visual reference images and textual reference descriptions of personalized targets from \ours dataset. The samples are taken from `\textit{backpack}', `\textit{bag}', `\textit{ball}', `\textit{book}', `\textit{camera}', `\textit{cellphone}', `\textit{eyeglasses}', `\textit{hat}', and `\textit{headphones}' object categories.}
    \label{fig:dataset_descriptions1}
    \vspace{-.2cm}
\end{figure}

\begin{figure}[!t]
    \centering
    \small
    \resizebox{\linewidth}{!}{
        \setlength{\tabcolsep}{.1em}
        \begin{tabular}{lll p{0.4cm} L{9.4cm}}
            \toprule
            \addlinespace[1mm]

            \multirow{3}{*}{\includegraphics[height=0.105\textwidth]{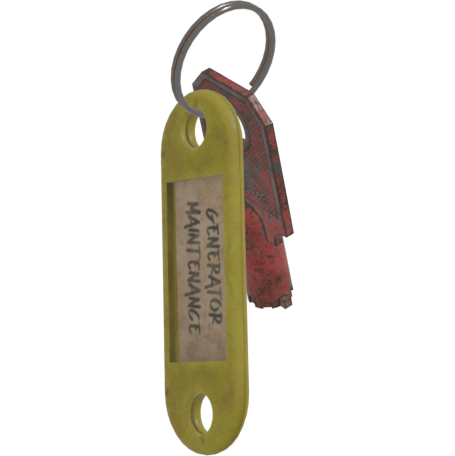}} &
            \multirow{3}{*}{\includegraphics[height=0.105\textwidth]{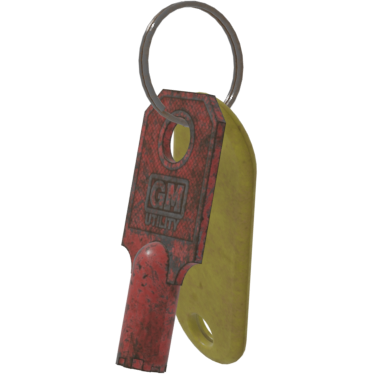}} &
            \multirow{3}{*}{\includegraphics[height=0.105\textwidth]{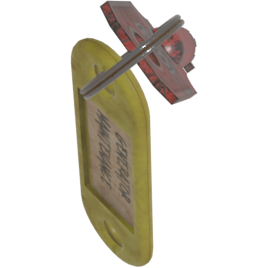}} 
            && \textit{a worn red key and a yellow keytag with a black text} \\
            \addlinespace[2mm]
            &&&& \textit{a yellow plastic tag with a red key} \\
            \addlinespace[2mm]
            &&&& \textit{a red rusty key and a yellow keytag with generator maintenance written on it} \\

            \addlinespace[1mm]
            \midrule
            \addlinespace[2mm]
            
            \multirow{3}{*}{\includegraphics[height=0.105\textwidth]{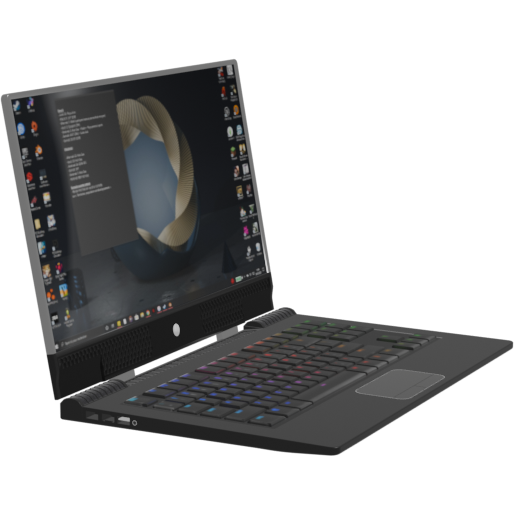}} & 
            \multirow{3}{*}{\includegraphics[height=0.105\textwidth]{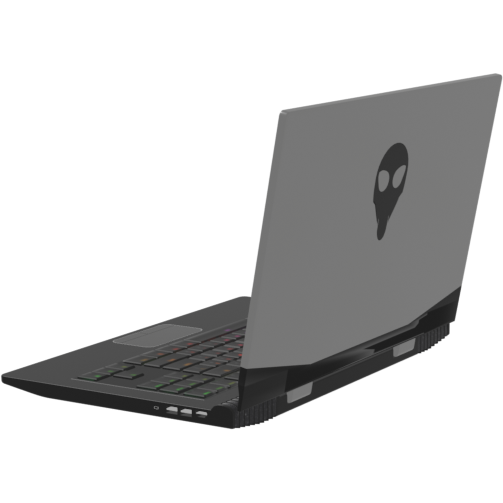}} & 
            \multirow{3}{*}{\includegraphics[height=0.105\textwidth]{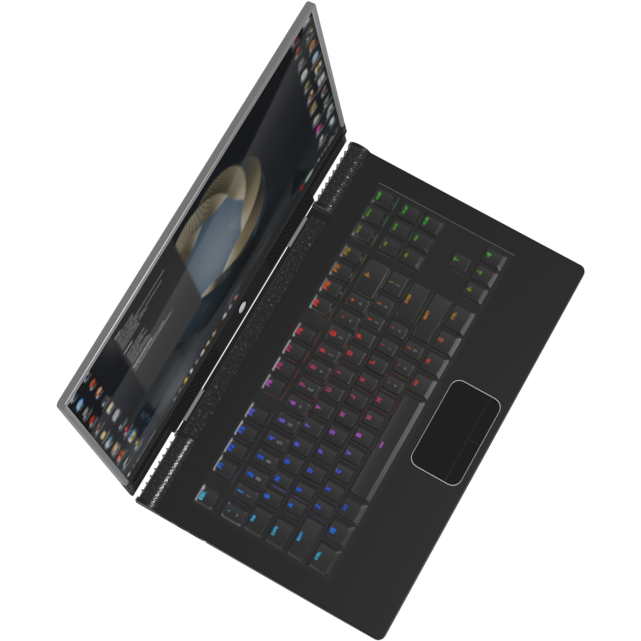}} 
            && \textit{a black and grey laptop with rgb keyboard} \\
            \addlinespace[2mm]
            &&&& \textit{a black and grey laptop with a alien head on the back} \\
            \addlinespace[2mm]
            &&&& \textit{a laptop having a gray top cover with an alien logo on the back, a black base panel and a rainbow colored keyboard} \\
            
            \addlinespace[1mm]
            \midrule
            \addlinespace[2mm]
            
            \multirow{3}{*}{\includegraphics[height=0.105\textwidth]{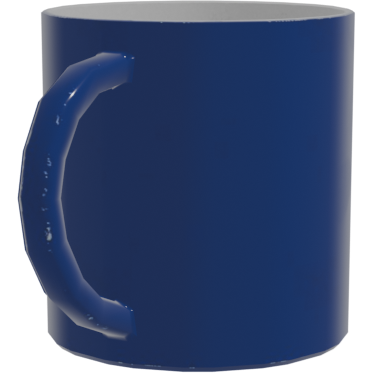}} & 
            \multirow{3}{*}{\includegraphics[height=0.105\textwidth]{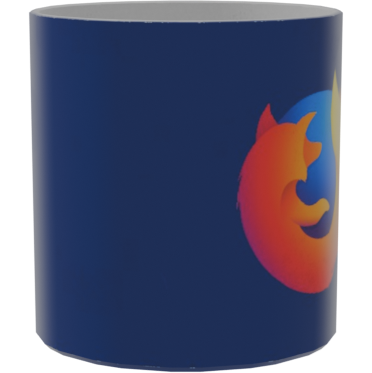}} & 
            \multirow{3}{*}{\includegraphics[height=0.105\textwidth]{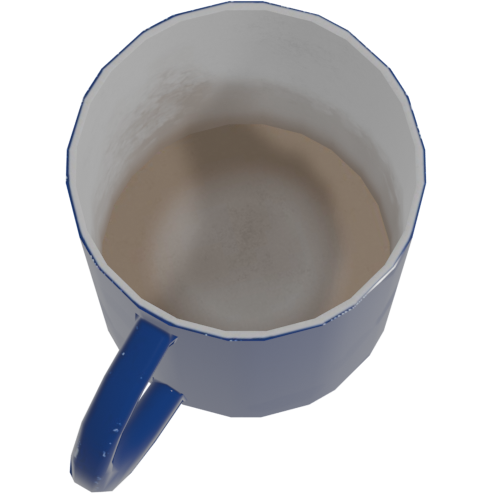}} 
            && \textit{a blue mug with a red fox logo on it} \\
            \addlinespace[2mm]
            &&&& \textit{a blue mug with a firefox logo on it} \\
            \addlinespace[2mm]
            &&&& \textit{a blue mug with the mozilla firefox logo, composed of a red fox around the globe, printed on it} \\
            
            \addlinespace[1mm]
            \midrule
            \addlinespace[2mm]
            
            \multirow{3}{*}{\includegraphics[height=0.105\textwidth] {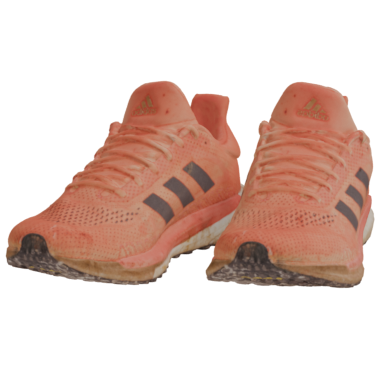}} & 
            \multirow{3}{*}{\includegraphics[height=0.105\textwidth]{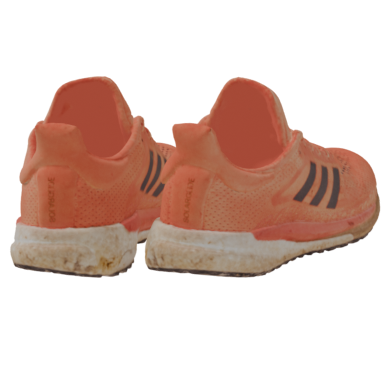}} & 
            \multirow{3}{*}{\includegraphics[height=0.105\textwidth]{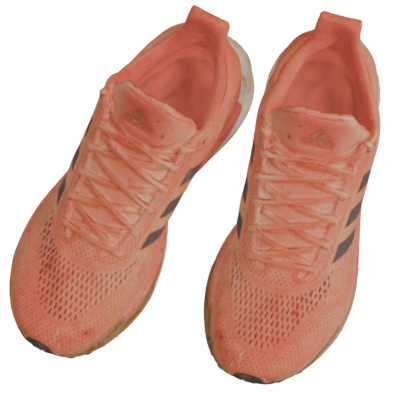}} 
            && \textit{a pair of orange adidas running shoes} \\
            \addlinespace[2mm]
            &&&& \textit{a pair of orange adidas sneakers with black stripes} \\
            \addlinespace[2mm]
            &&&& \textit{a pair of orange running shoes with orange laces, black adidas stripes, and white outsoles}
            \\
            
            \addlinespace[1mm]
            \midrule
            \addlinespace[2mm]
            
            \multirow{3}{*}{\includegraphics[height=0.105\textwidth] {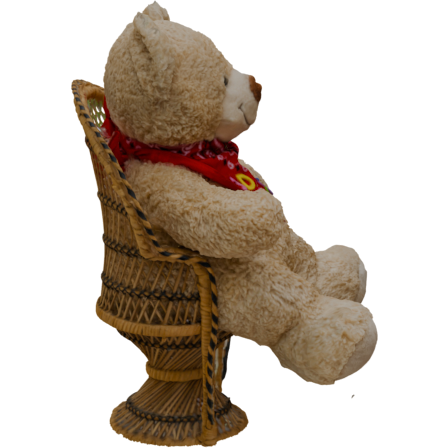}} & \multirow{3}{*}{\includegraphics[height=0.105\textwidth]{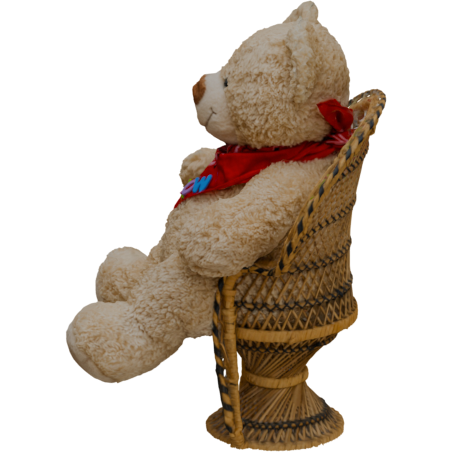}} & \multirow{3}{*}{\includegraphics[height=0.105\textwidth]{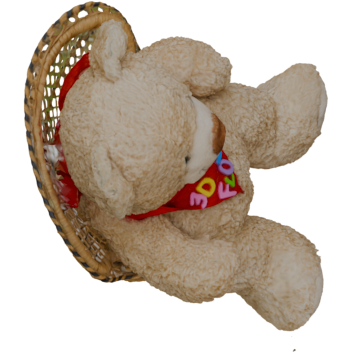}} 
            && \textit{a beige teddy bear with a red bandana on a wooden chair} \\
            \addlinespace[2mm]
            &&&& \textit{a teddy bear sitting in a wicker chair with a red bandana on its neck} \\
            \addlinespace[2mm]
            &&&& \textit{a cream-colored smiling teddy bear with a red scarf and sitting on a woven chair}
            \\
            
            \addlinespace[1mm]
            \midrule
            \addlinespace[2mm]

            \multirow{3}{*}{\includegraphics[height=0.105\textwidth] {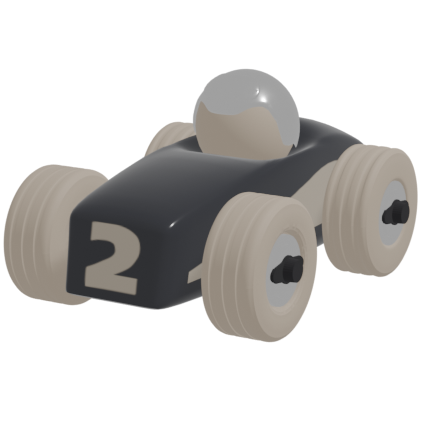}} & 
            \multirow{3}{*}{\includegraphics[height=0.105\textwidth]{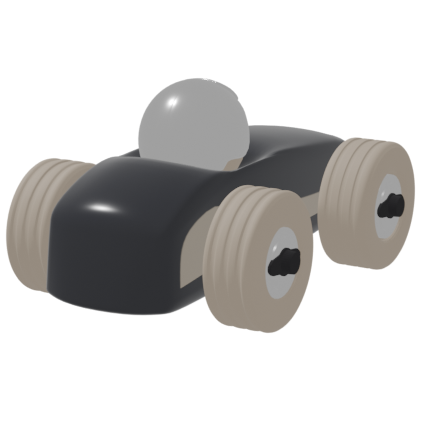}} & 
            \multirow{3}{*}{\includegraphics[height=0.105\textwidth]{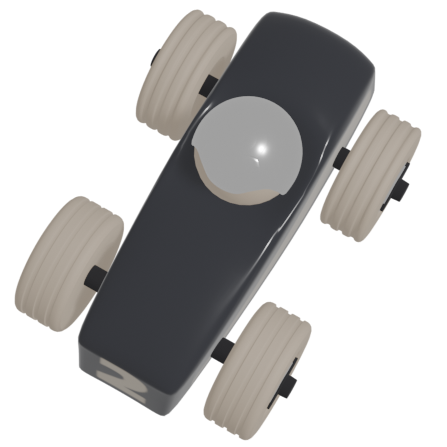}} 
            && \textit{a black and white toy car with a number 2 on the front} \\
            \addlinespace[2mm]
            &&&& \textit{a black and white toy car} \\
            \addlinespace[2mm]
            &&&& \textit{a black toy race car, with white wheels, a number 2 painted on the side, and a ball replacing the drive} \\
            
            \addlinespace[1mm]
            \midrule
            \addlinespace[2mm]
            
            \multirow{3}{*}{\includegraphics[height=0.105\textwidth]{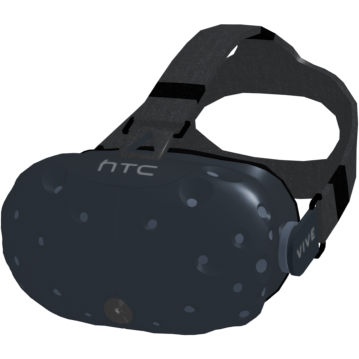}} & 
            \multirow{3}{*}{\includegraphics[height=0.105\textwidth]{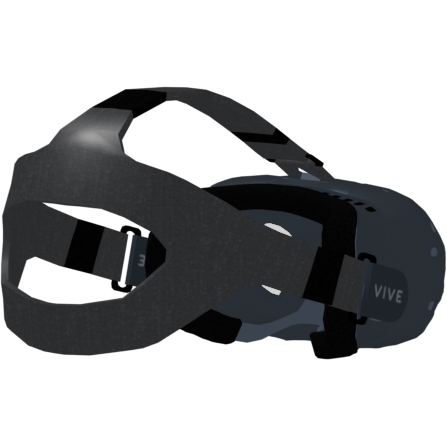}} & 
            \multirow{3}{*}{\includegraphics[height=0.105\textwidth]{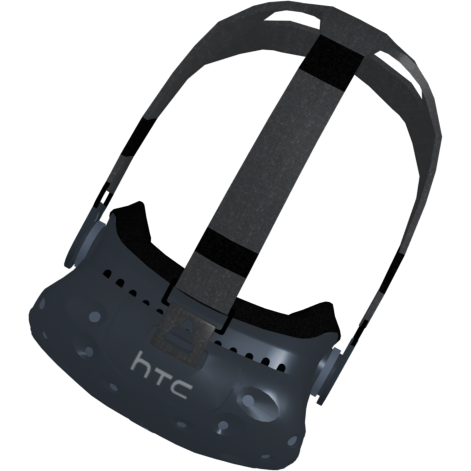}} 
            && \textit{a black htc visor with blue polka dots} \\
            \addlinespace[2mm]
            &&&& \textit{a blue rounded virtual reality headset with a black strap} \\
            \addlinespace[2mm]
            &&&& \textit{a htc visor having black bands and blue front side with light blue dots} \\
            
            \addlinespace[1mm]
            \midrule
            \addlinespace[2mm]
            
            \multirow{3}{*}{\includegraphics[height=0.105\textwidth]{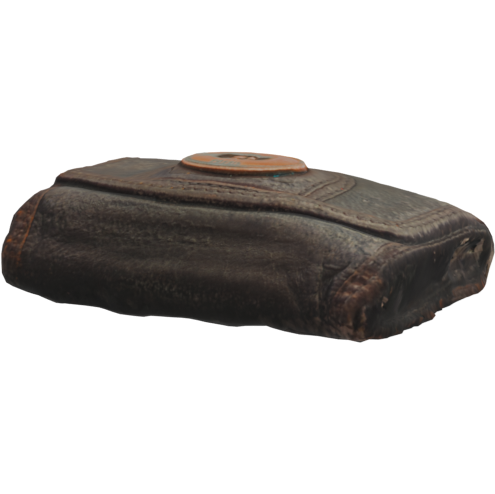}} & 
            \multirow{3}{*}{\includegraphics[height=0.105\textwidth]{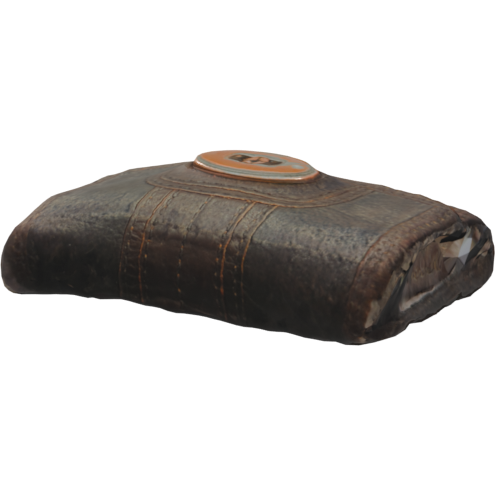}} & 
            \multirow{3}{*}{\includegraphics[height=0.105\textwidth]{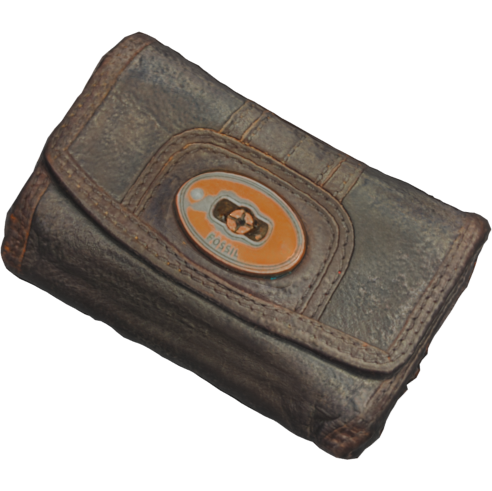}} 
            && \textit{a black leather wallet with an orange plate} \\
            \addlinespace[2mm]
            &&&& \textit{a brown leather wallet with a button on it} \\
            \addlinespace[2mm]
            &&&& \textit{a dark brown leather wallet having an orange patch with fossil written on it} \\
            
            \addlinespace[1mm]
            \midrule
            \addlinespace[2mm]
            
            \multirow{3}{*}{\includegraphics[height=0.105\textwidth]{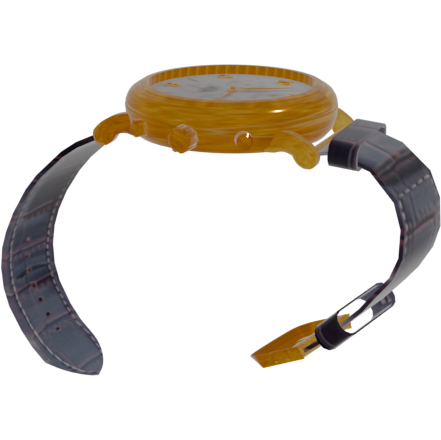}} & 
            \multirow{3}{*}{\includegraphics[height=0.105\textwidth]{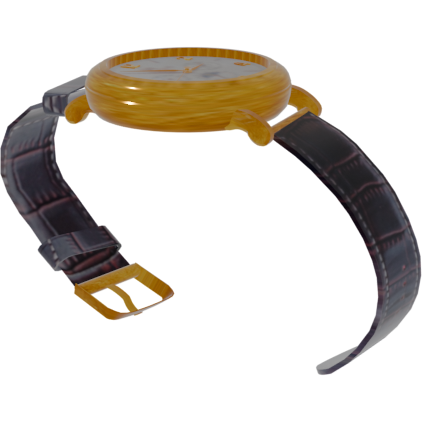}} & 
            \multirow{3}{*}{\includegraphics[height=0.105\textwidth]{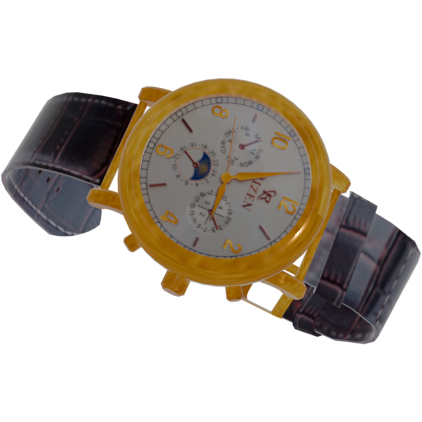}} 
            && \textit{a gold and grey watch with black leather strap} \\
            \addlinespace[2mm]
            &&&& \textit{a brown leather wallet with a button on it} \\
            \addlinespace[2mm]
            &&&& \textit{a rounded watch having a thick golden case, white dial and black leather band with a golden buckle} \\
            
            \addlinespace[1mm]
            \bottomrule
        \end{tabular}
    }
    \vspace{-.1cm}
    \caption{Visual reference images and textual reference descriptions of personalized targets from \ours dataset. The samples are taken from `\textit{keys}', `\textit{laptop}',  `\textit{mug}', `\textit{shoes}', `\textit{teddy bear}', `\textit{toy}', `\textit{visor}', `\textit{wallet}', and `\textit{watch}' object categories.}
    \label{fig:dataset_descriptions2}
    \vspace{-.2cm}
\end{figure}

Additionally, we present samples 
including both visual and textual modalities for the input references associated with some of the object instances of \ours dataset in Fig.~\ref{fig:dataset_descriptions1} and Fig.~\ref{fig:dataset_descriptions2}. In particular, we show the three views composing the set of visual references and the three manually annotated descriptions for the textual references.

\begin{table}[!t]
    \centering
    \caption{Statistics about the number of distractors placed in the episodes of the training and validation sets of \ours dataset. We consider the distractors belonging both to the same category of the target and to other categories.}
    \label{fig:dataset_stats_suppl}
    \resizebox{0.65\linewidth}{!}{
        \renewcommand{\arraystretch}{1.1} % Default value: 1
        \begin{tabular}{l c c c c c c c c}
            \toprule
            \# of Distractors &&\multicolumn{3}{c}{\textit{Same Object Category}} && \multicolumn{3}{c}{\textit{Other Categories}}\\
            \midrule
            && \textbf{Train} &&\textbf{Val} && \textbf{Train} && \textbf{Val}\\
            \cmidrule{3-5}\cmidrule{7-9}
            Max && 6 & & 3 && 13 && 10\\
            Average && 2.93 & & 2.90 && 7.75 && 7.19 \\
            Standard Deviation && 0.33 && 0.37 && 2.84 && 2.82\\
            \bottomrule
        \end{tabular}
    }
    \vspace{-0.2cm}
\end{table}

\tit{Object Selection and Distribution Criteria}
The scope of \shorttask is to provide a benchmark to evaluate an agent tasked with finding a specific object that can be located anywhere in an unexplored environment, where distractors of the same category are present; hence, the object categories are selected according to the following criteria: (i) objects that are highly customizable in terms of shapes, colors, and other visual aspects, (ii) objects that are frequently moved and can be placed anywhere, and (iii) objects of common use for which is reasonable to ask a robot to find.

\tit{Additional Information about Dataset Generation}
In Table~\ref{fig:dataset_stats_suppl}, we provide statistics on the number of distractors placed in the training and validation episodes of \ours dataset. 
During the generation of \shorttask episodes, a maximum number of distractors, both from the same category as the target instance and from other categories, is sampled from the set of available objects. The final number of additional objects in each episode is determined by the number of suitable surfaces and the available space on these surfaces. 
During the dataset generation process, objects are positioned above these surfaces and lowered until they contact the surface. 
If an object cannot be initially placed due to size constraints or collisions with other elements or walls, the placing process for that object is aborted, and another one is sampled from unused object instances. After the generation of the dataset of episodes, an additional assessment is performed through the Habitat simulator to remove the episodes containing objects that are not reachable from the starting position of the agent.


\begin{figure}[t]
    \centering
    \resizebox{.95\linewidth}{!}{
        \setlength{\tabcolsep}{.1em}
        \begin{tabular}{rr p{2mm} rr }
            \addlinespace[1mm]
            \includegraphics[height=0.215\textwidth]{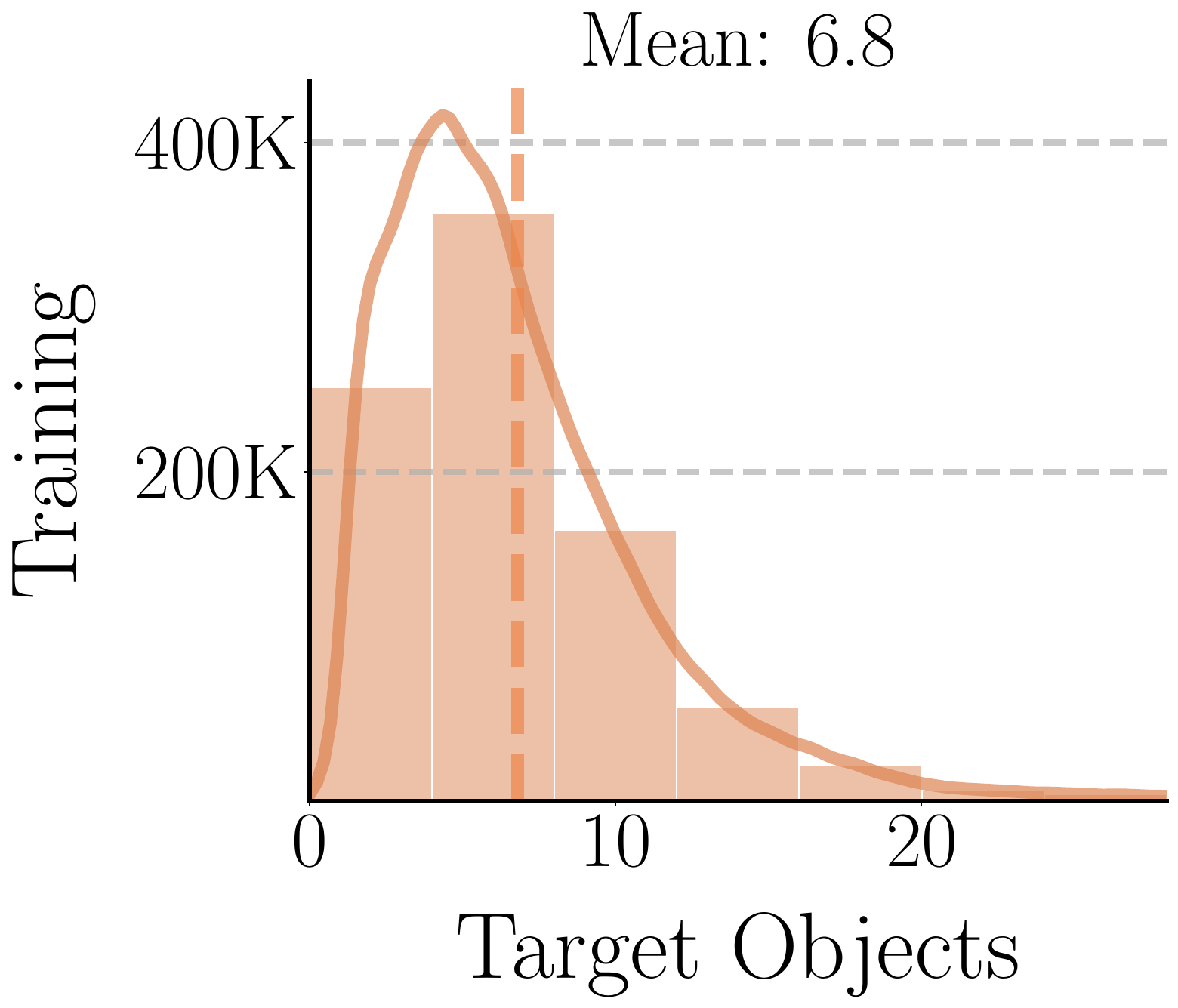} &
            \includegraphics[height=0.215\textwidth]{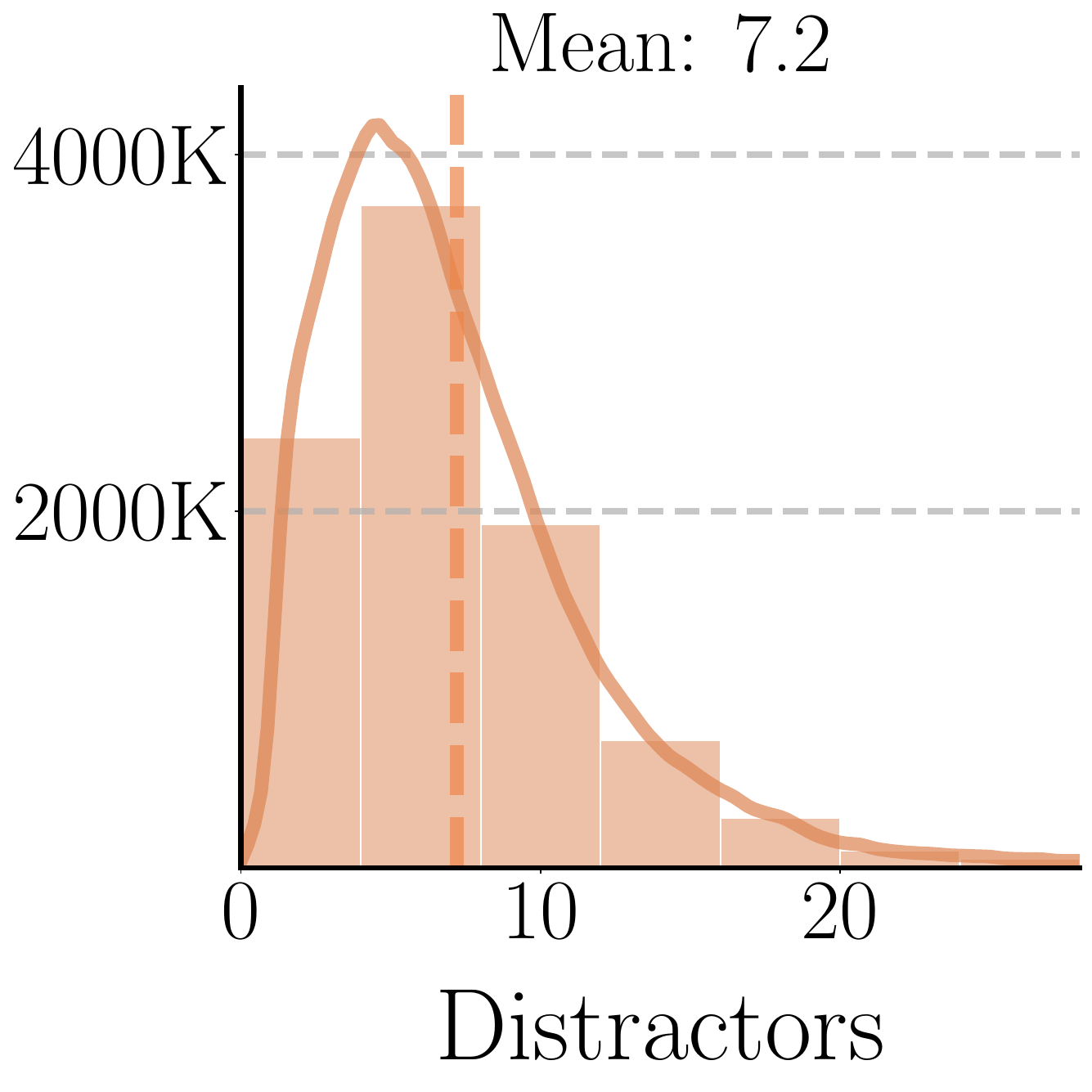} & &
            \includegraphics[height=0.215\textwidth]{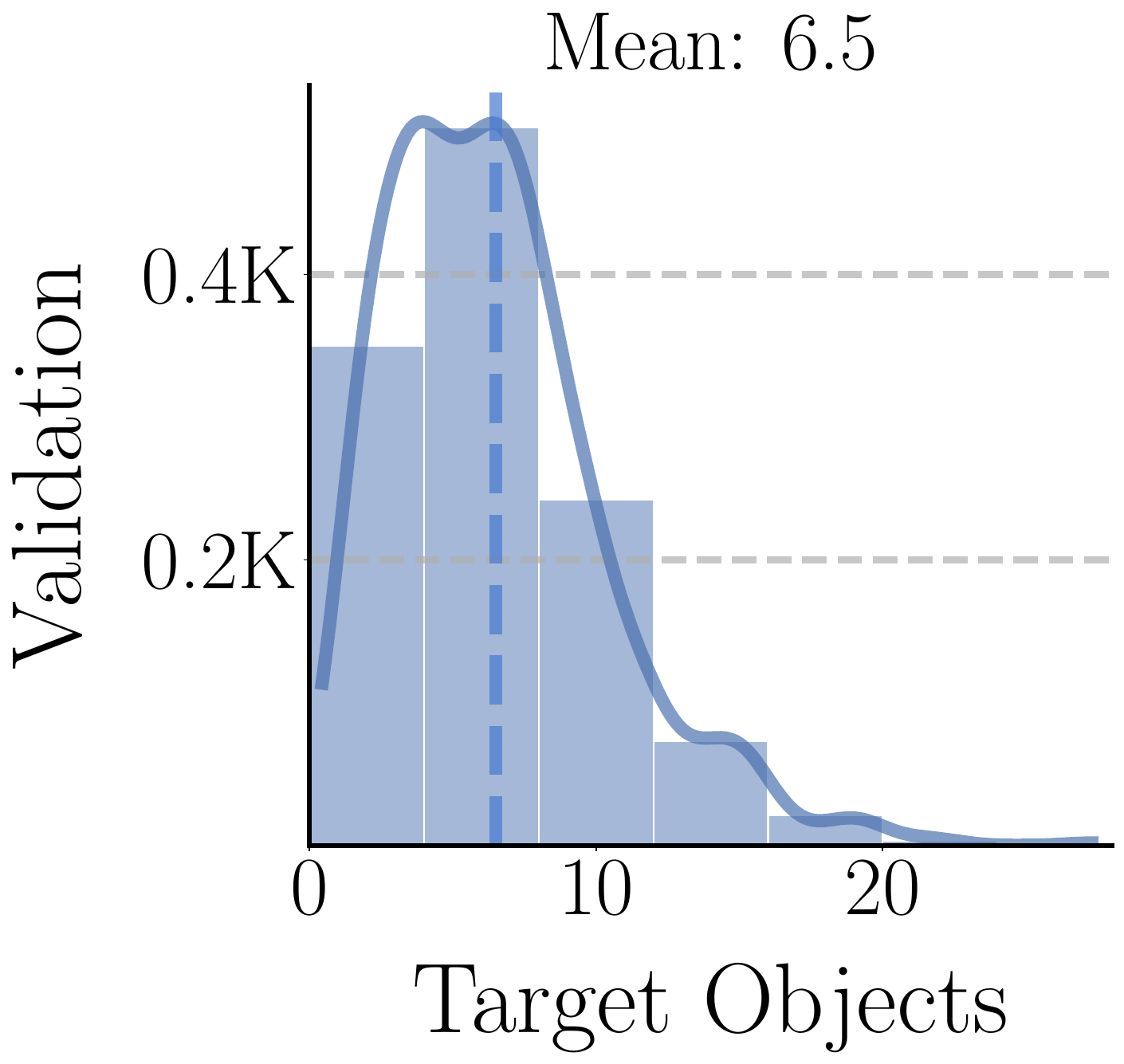} &
            \includegraphics[height=0.215\textwidth]{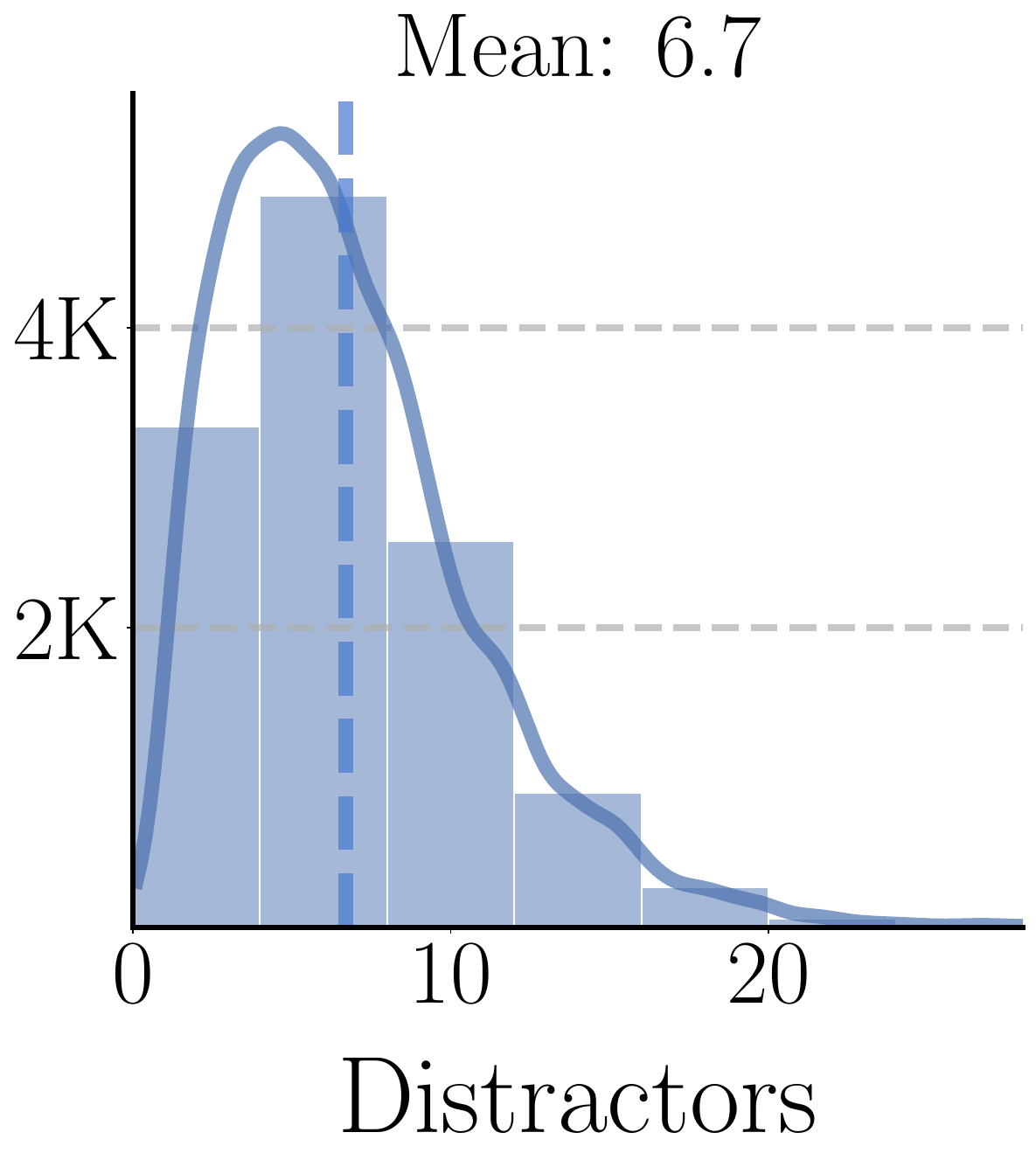} \\
        \end{tabular}
    }
    \vspace{-.2cm}
    \caption{Euclidean distances of the objects included in the episodes of training (\textcolor{orange}{orange}) and validation (\textcolor{RoyalBlue}{blue}) splits of \ours dataset. The plots consider the distances from the start position to the target object (left) and to all distractors (right). Distances are measured in meters, with the mean value for each plot displayed at the top.} 
    \label{fig:distances_plots_supp}
    \vspace{-.2cm}
\end{figure}
\begin{figure}[t]
    \centering
    \resizebox{.93\linewidth}{!}{
        \setlength{\tabcolsep}{.1em}
        \begin{tabular}{rrrr}
            \addlinespace[1mm]
            \includegraphics[height=0.215\textwidth]{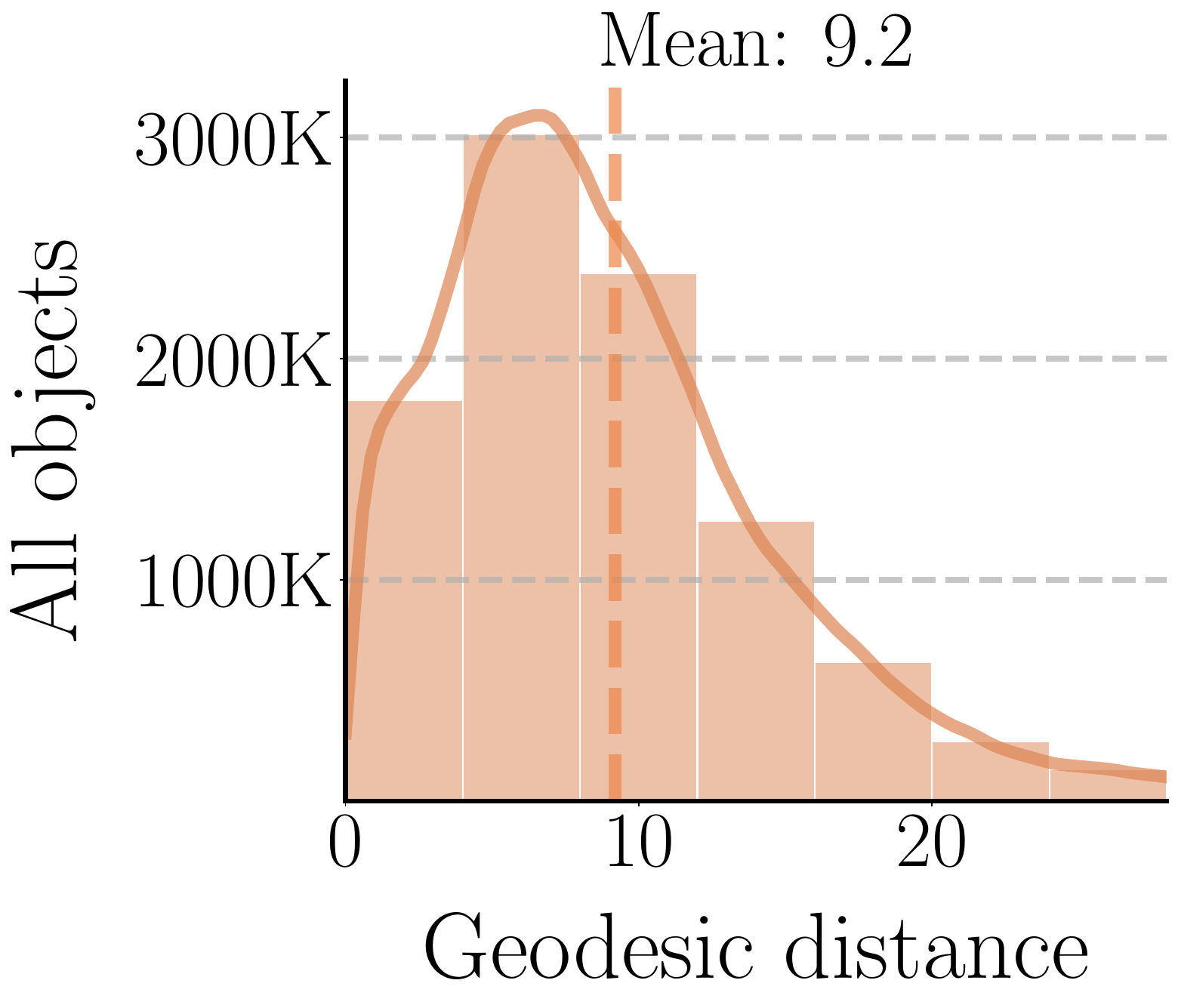} &
            \includegraphics[height=0.215\textwidth]{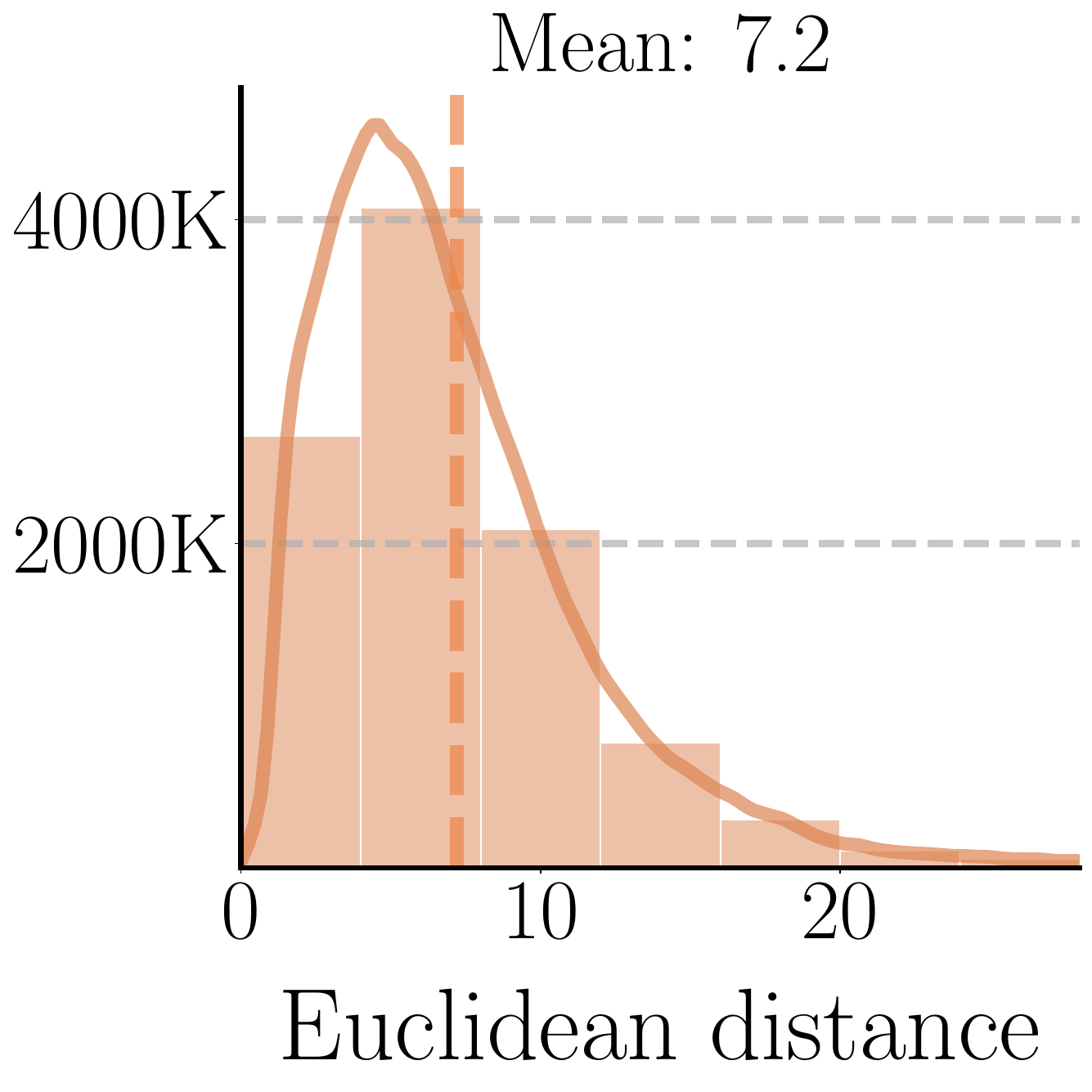} &
            \includegraphics[height=0.215\textwidth]{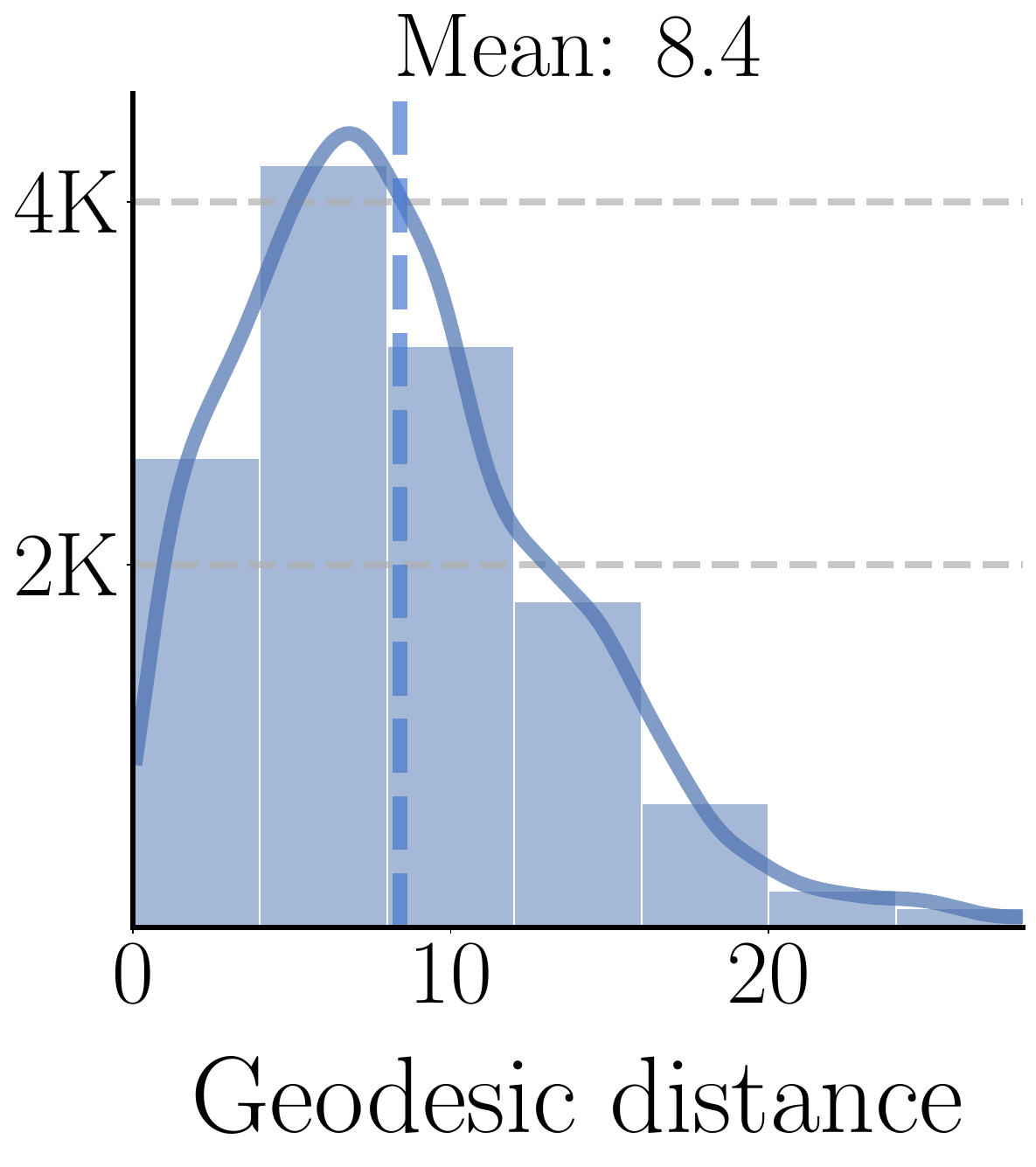} &
            \includegraphics[height=0.215\textwidth]{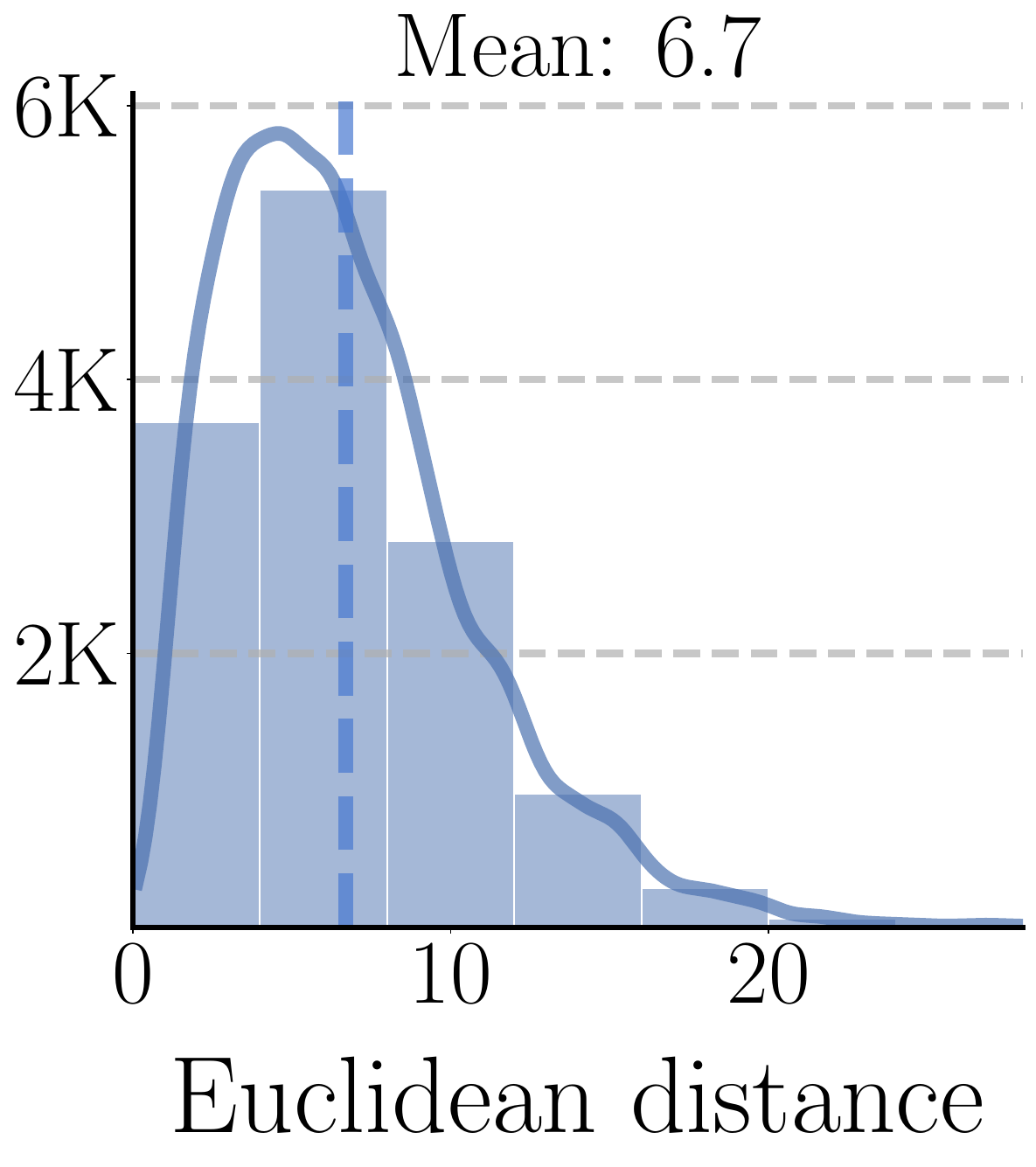} \\
        \end{tabular}
    }
    \vspace{-.2cm}
    \caption{Plots of the geodesic and Euclidean distances for all the objects placed in the episodes of \ours dataset. Training (\textcolor{orange}{orange}) and validation splits (\textcolor{RoyalBlue}{blue}) are presented in terms of distances from the start position to all the spawned additional objects. All the distances are plotted in meters, and the mean value of each plot is shown on top.}
    \label{fig:distances_plots_all_objs_supp}
    \vspace{-.2cm}
\end{figure}

\tit{Object Distances} In addition to Fig.~\ref{fig:distances_plots} of the main paper, Fig.~\ref{fig:distances_plots_supp} presents a plot depicting the Euclidean distances of target objects and distractors from the starting position of the agent in the episodes of both training and validation splits of \ours dataset. When considering the Euclidean distance, the distribution of the distances of the additional objects remains consistent with the geodesic distances presented in the main paper.
Furthermore, the plots of the distances of all additional objects (target instances and distractors) are presented in Fig.~\ref{fig:distances_plots_all_objs_supp}.

\begin{figure}[!t]
    \centering
    \Large
    \resizebox{\linewidth}{!}{
        \setlength{\tabcolsep}{.06em}
        \begin{tabular}{c p{4mm} c p{4mm} c c c c}
            \addlinespace[4mm]
            % Row 1
            \includegraphics[height=0.3\textwidth]{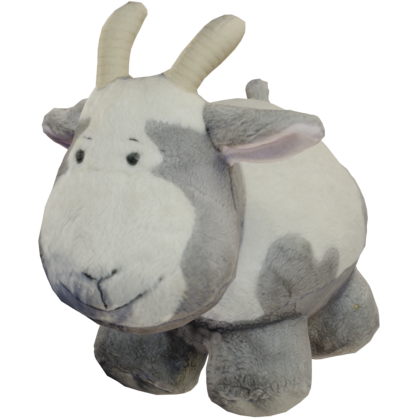} & &
            \includegraphics[height=0.3\textwidth]{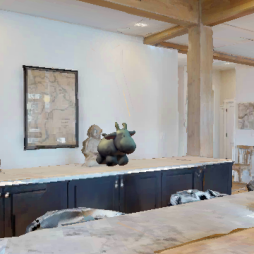} & &
            \includegraphics[height=0.3\textwidth,width=0.3\textwidth]{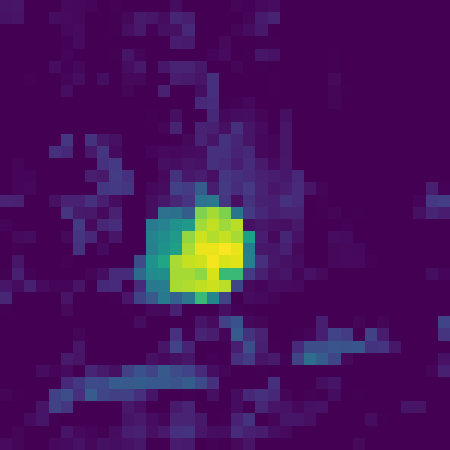} &
            \includegraphics[height=0.3\textwidth]{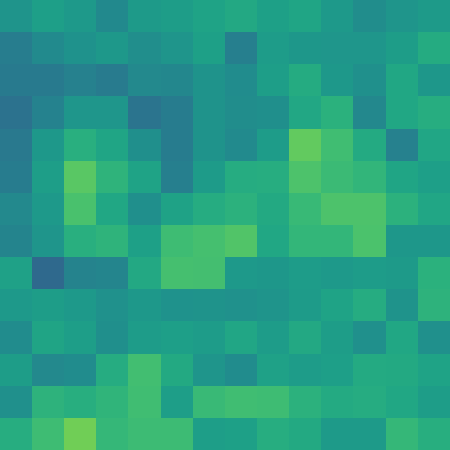} &
            \includegraphics[height=0.3\textwidth]{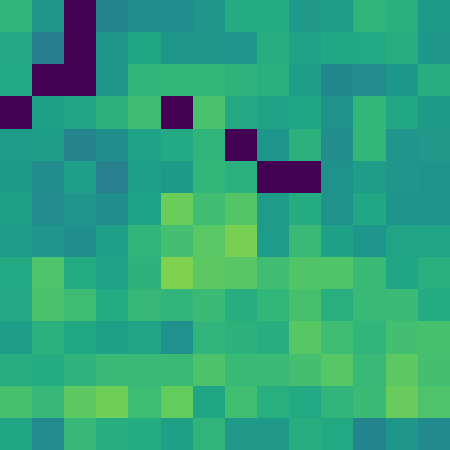} & \includegraphics[height=0.3\textwidth,width=0.3\textwidth]{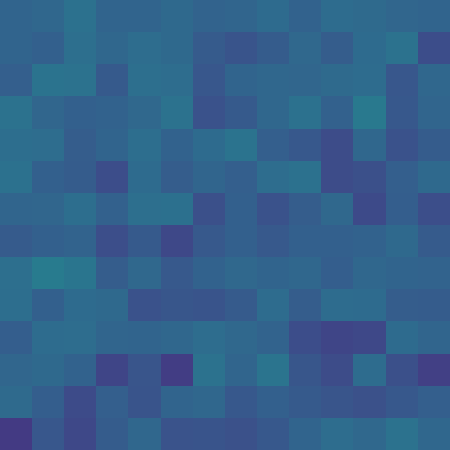} \\
            % Row 2
            \includegraphics[height=0.3\textwidth]{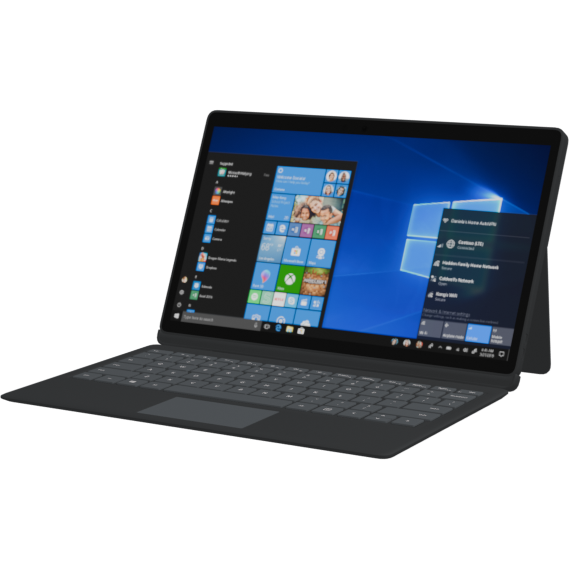} & &
            \includegraphics[height=0.3\textwidth]{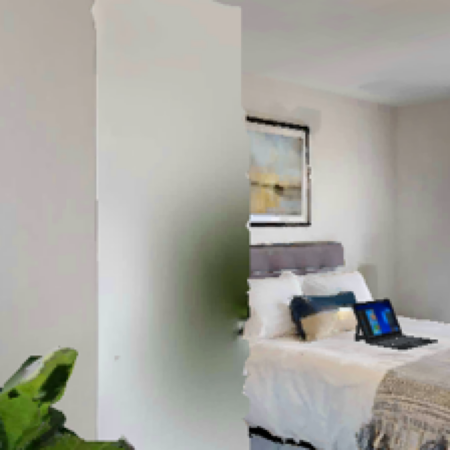} & & \includegraphics[height=0.3\textwidth,width=0.3\textwidth]{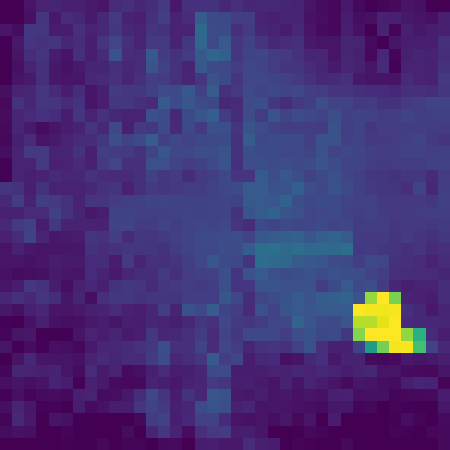} &
            \includegraphics[height=0.3\textwidth]{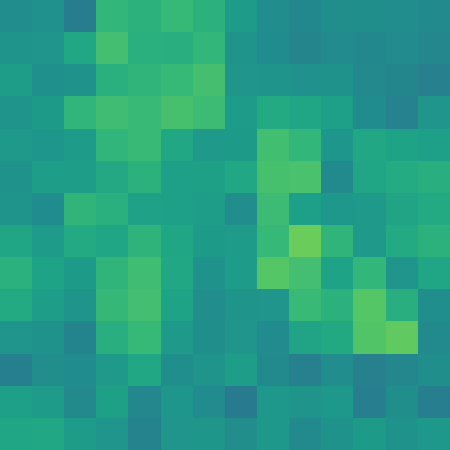} &
            \includegraphics[height=0.3\textwidth]{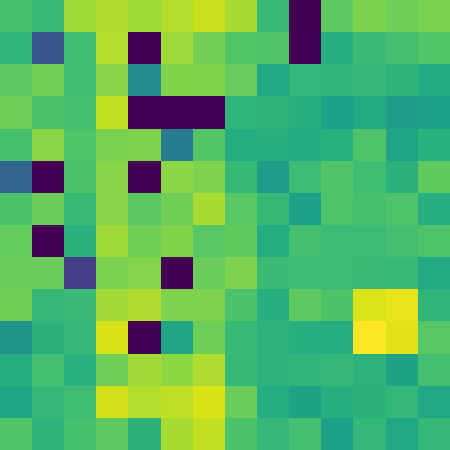} & \includegraphics[height=0.3\textwidth,width=0.3\textwidth]{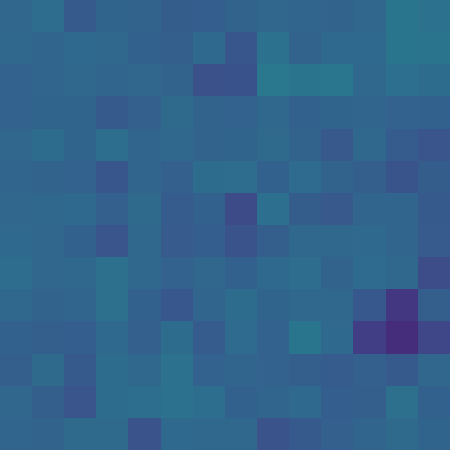} \\
            \addlinespace[2mm]
            \textbf{Reference} & &
            \textbf{Observation} & &
            \textbf{DINOv2} &
            \textbf{DINO} &
            \textbf{CLIP (Visual)} & 
            \textbf{CLIP (Textual)} \\

        \end{tabular}
    }
    % \vspace{-1mm}
    \caption{Comparison of the similarities between the patch-level features on two observations of an agent extracted with different backbones, DINOv2, DINO, CLIP with visual references, and CLIP with textual references, and the references. Purple values represent low similarity values, while yellow values represent high similarity values.}
    \vspace{-2mm}
    \label{fig:activation_comparison}
\end{figure}

\tit{Modular Agent Activations}
In Fig.~\ref{fig:activation_comparison} we present a comparison of the similarities computed between the patch-level features of different backbones on the observations of the agent and the references. In particular,  we show these similarities on DINOv2~\cite{oquab2023dinov2}, DINO~\cite{caron2021emerging}, CLIP with visual references, and CLIP with textual references~\cite{radford2021learning}. The resolution of the similarities extracted from DINOv2 is higher than the others since we employed the input resolution $518\times518$ on which the ViT-B/14 model has been trained, which corresponds to a grid of $37\times37$ patches, whereas DINO and CLIP are based on a ViT-B/16 backbone with $224\times224$ as input resolution. It is noteworthy that DINOv2 exhibits strong semantic localization properties, with high similarity values on the exact location of the image on which the target is observed. On the contrary, DINO and CLIP tend to exhibit less well-localized similarities. Moreover, CLIP with visual references has a high similarity on the patches corresponding to the laptop in the observation, whereas CLIP with textual references has a low similarity on the same patches.

\begin{figure}[!t]
    \huge
    \resizebox{\linewidth}{!}{
        \begin{tabular}{cc C{0mm} cc C{0mm} cc}
            \begin{subfigure}
                \centering
                \includegraphics[height=0.5\linewidth]{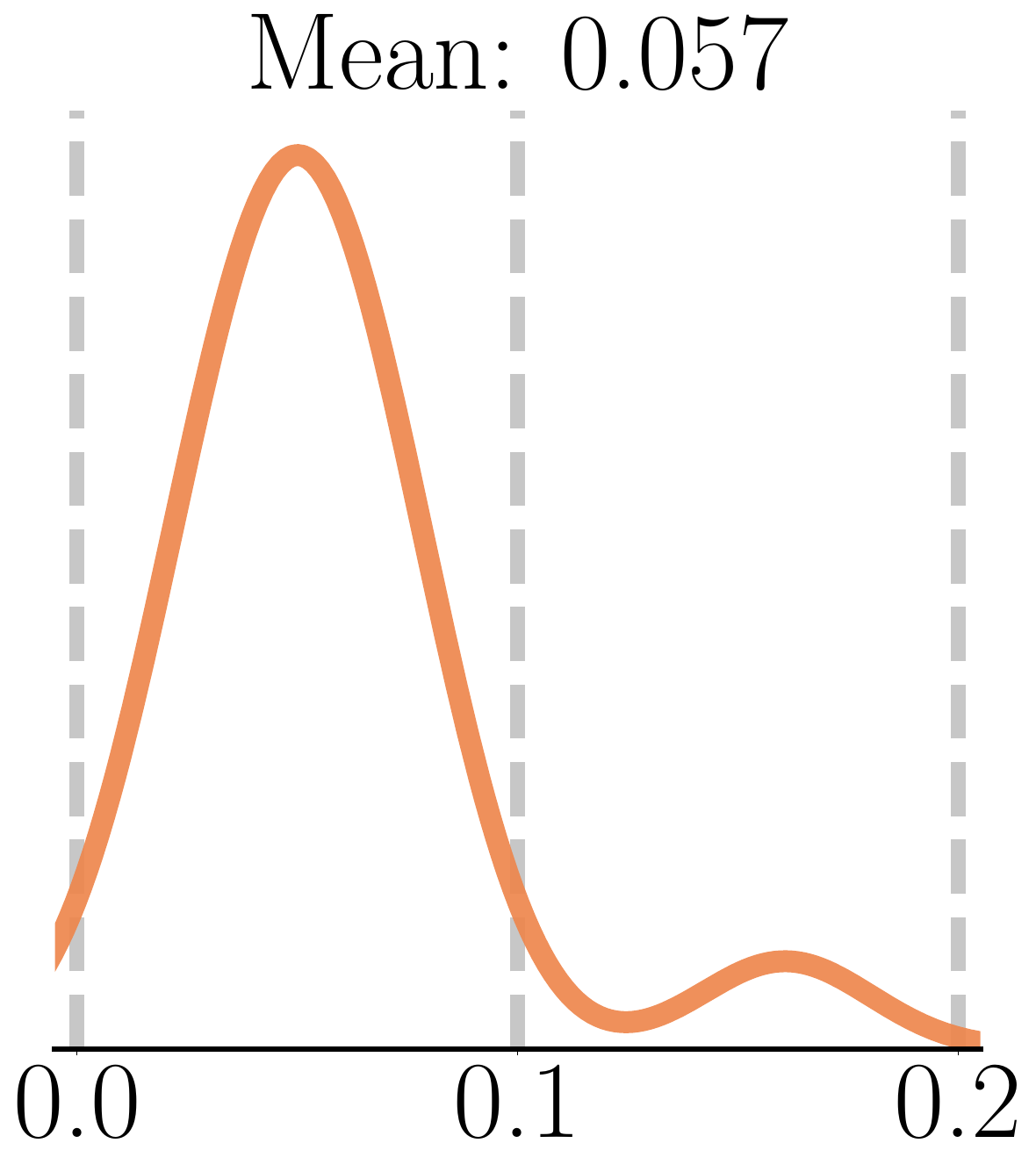}
            \end{subfigure} &
            \begin{subfigure}
                \centering
                \includegraphics[height=0.5\linewidth]{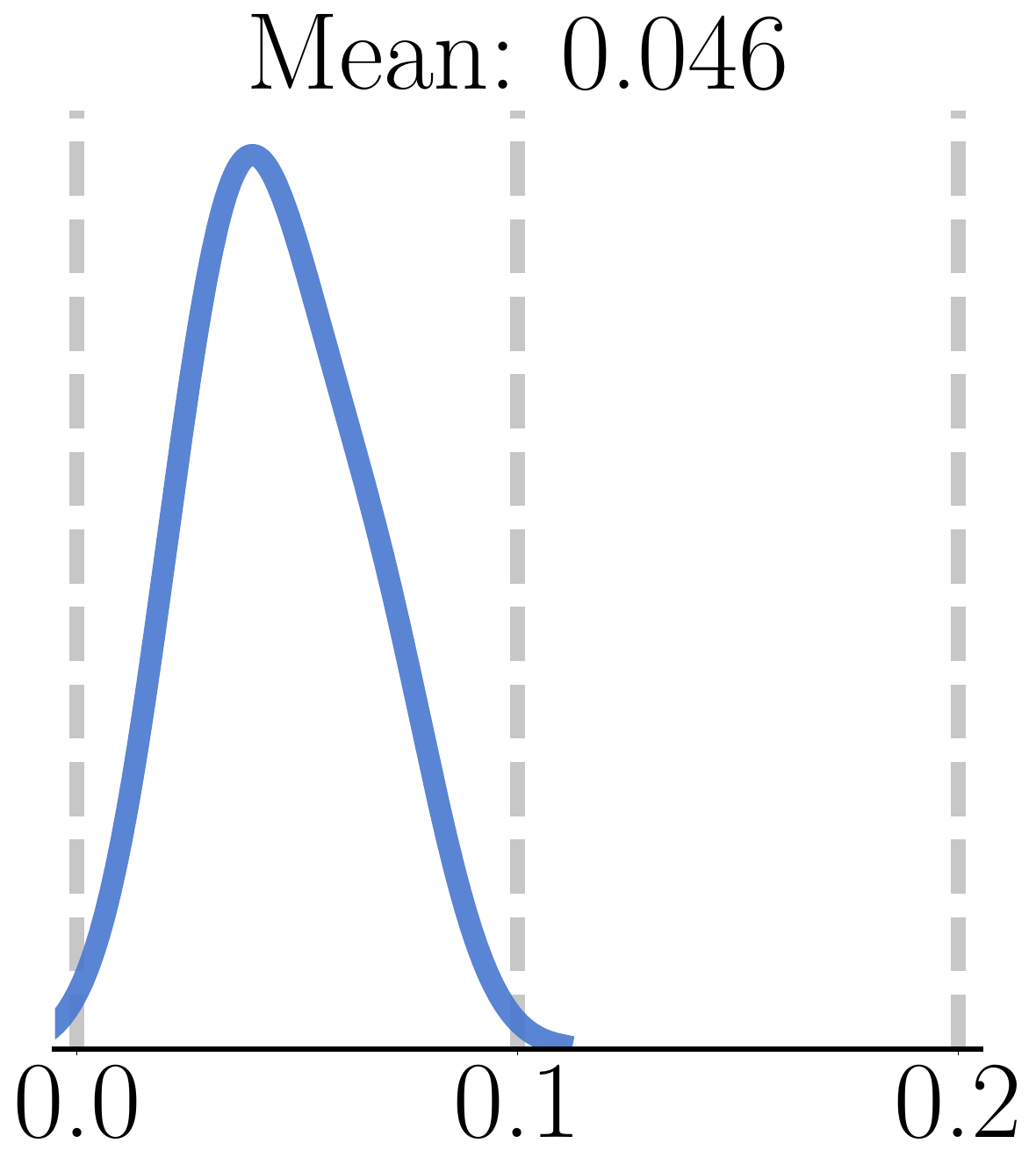}
            \end{subfigure} & &
            \begin{subfigure}
                \centering
                \includegraphics[height=0.5\linewidth]{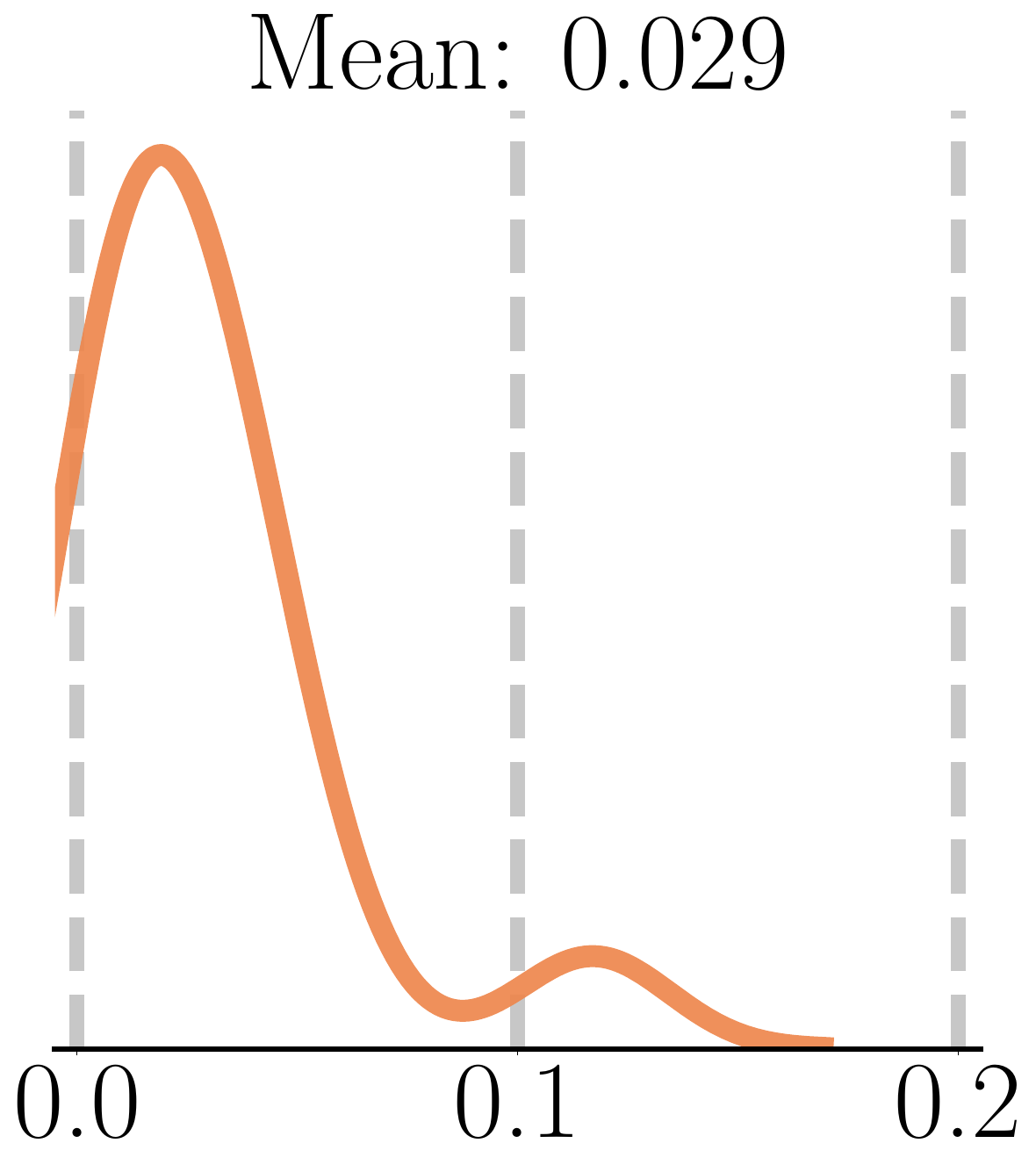}
            \end{subfigure} &
            \begin{subfigure}
                \centering
                \includegraphics[height=0.5\linewidth]{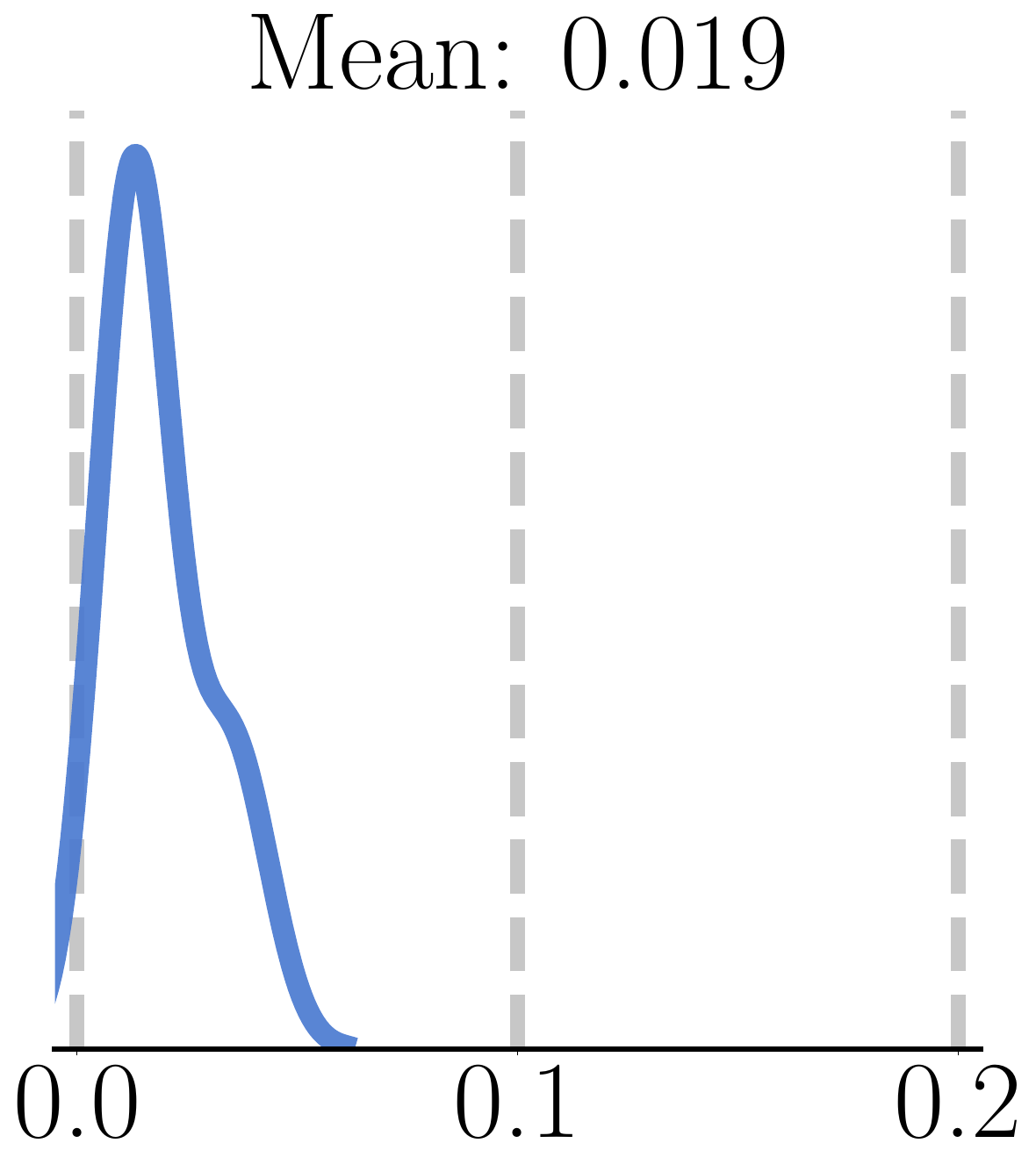}
            \end{subfigure} & &
            \begin{subfigure}
                \centering
                \includegraphics[height=0.5\linewidth]{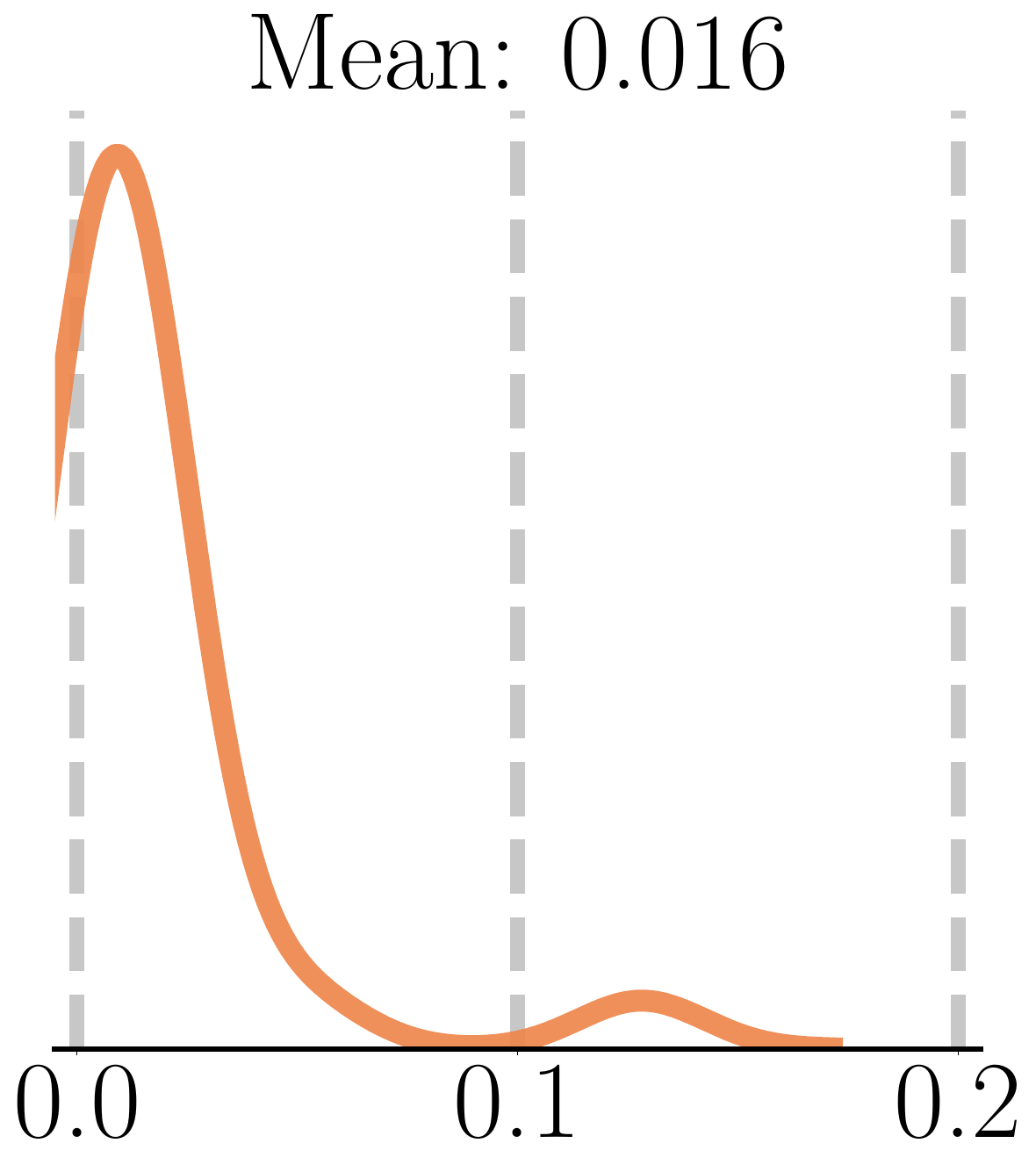}
            \end{subfigure} &
            \begin{subfigure}
                \centering
                \includegraphics[height=0.5\linewidth]{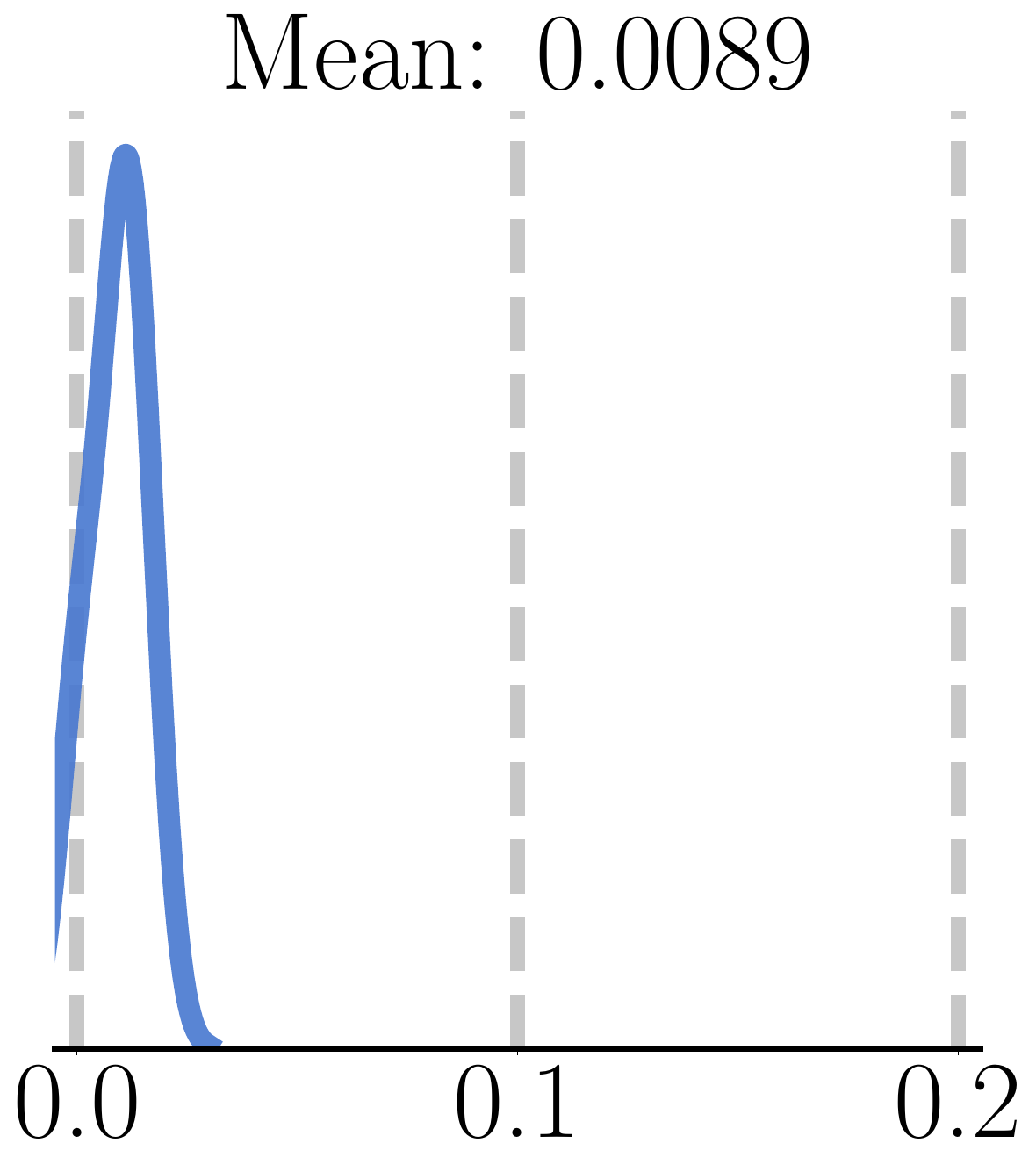}
            \end{subfigure} \\
            
            \multicolumn{2}{c}{\textbf{Backpack}} & & \multicolumn{2}{c}{\textbf{Bag}} & & \multicolumn{2}{c}{\textbf{Ball}} \\
            \addlinespace[5mm]
            
            \begin{subfigure}
                \centering
                \includegraphics[height=0.5\linewidth]{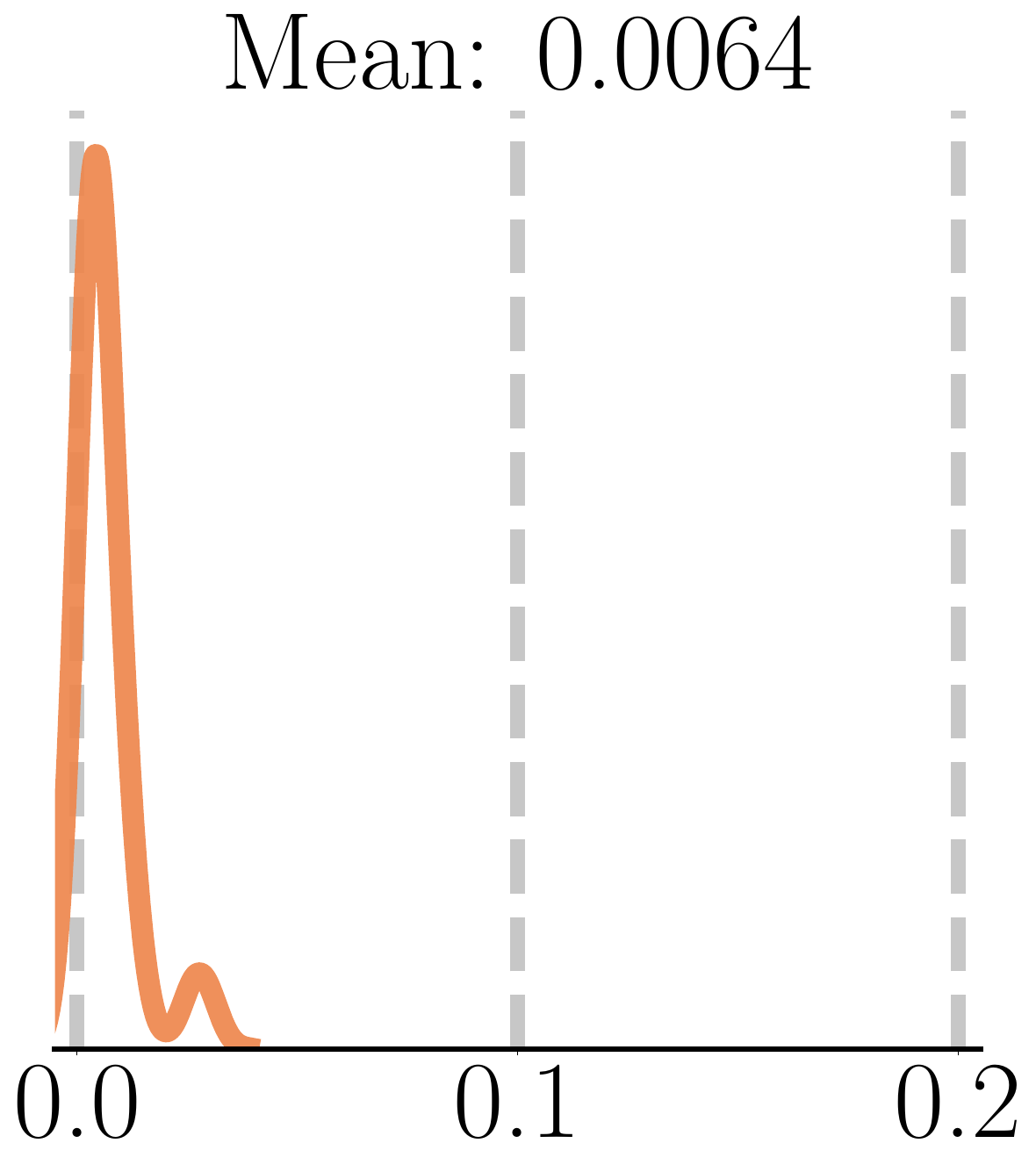}
            \end{subfigure} &
            \begin{subfigure}
                \centering
                \includegraphics[height=0.5\linewidth]{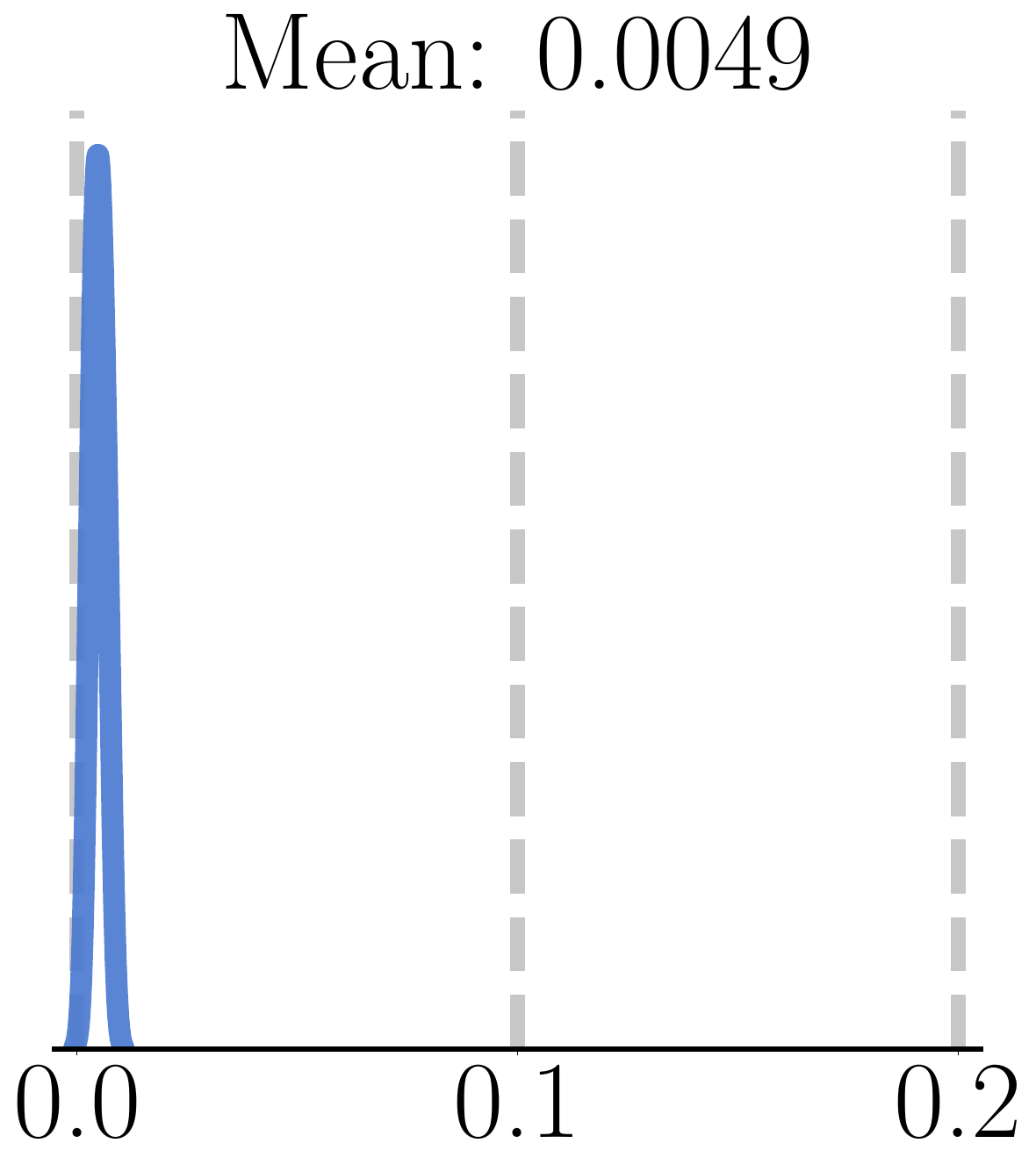}
            \end{subfigure} & &
            \begin{subfigure}
                \centering
                \includegraphics[height=0.5\linewidth]{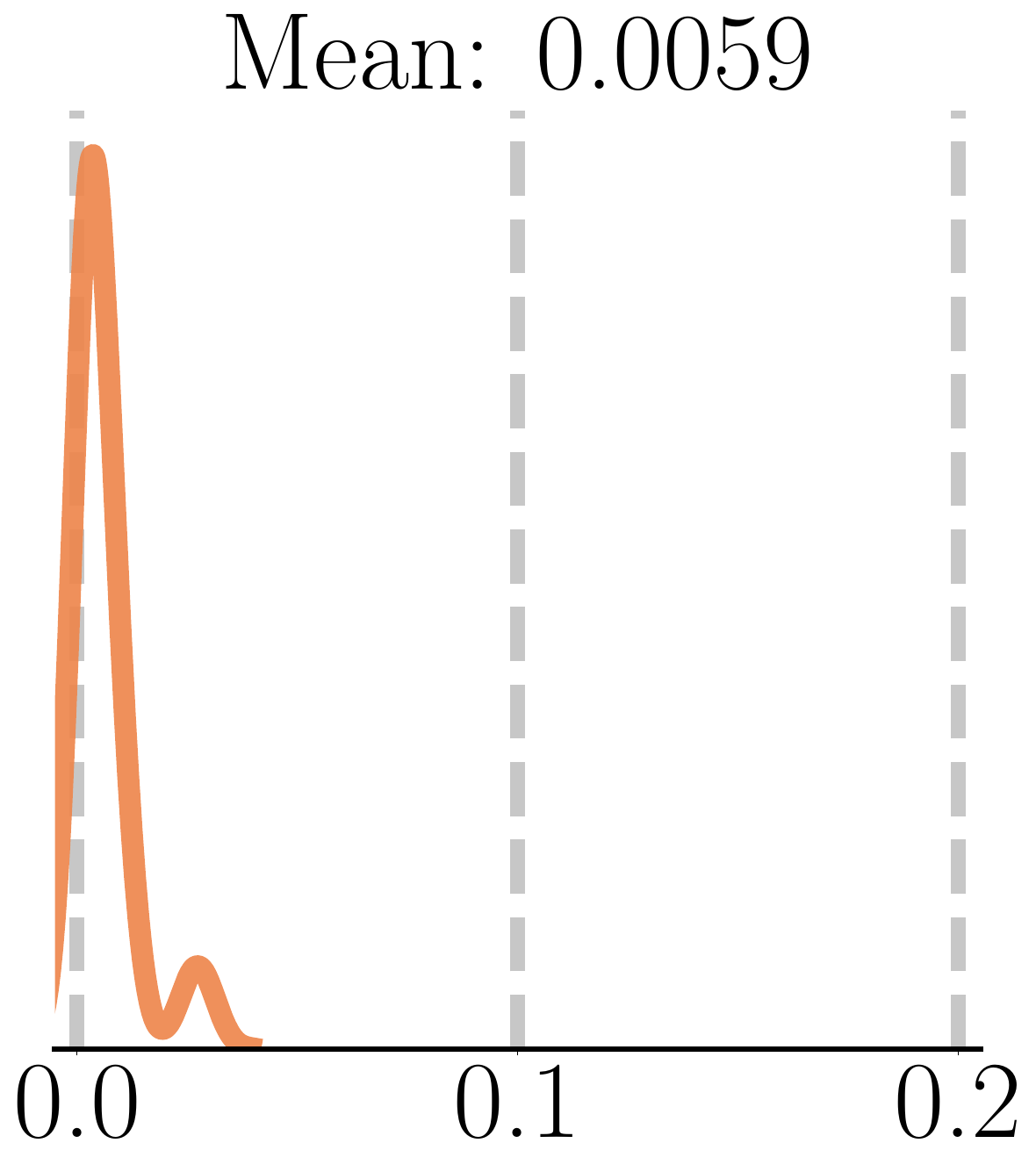}
            \end{subfigure} &
            \begin{subfigure}
                \centering
                \includegraphics[height=0.5\linewidth]{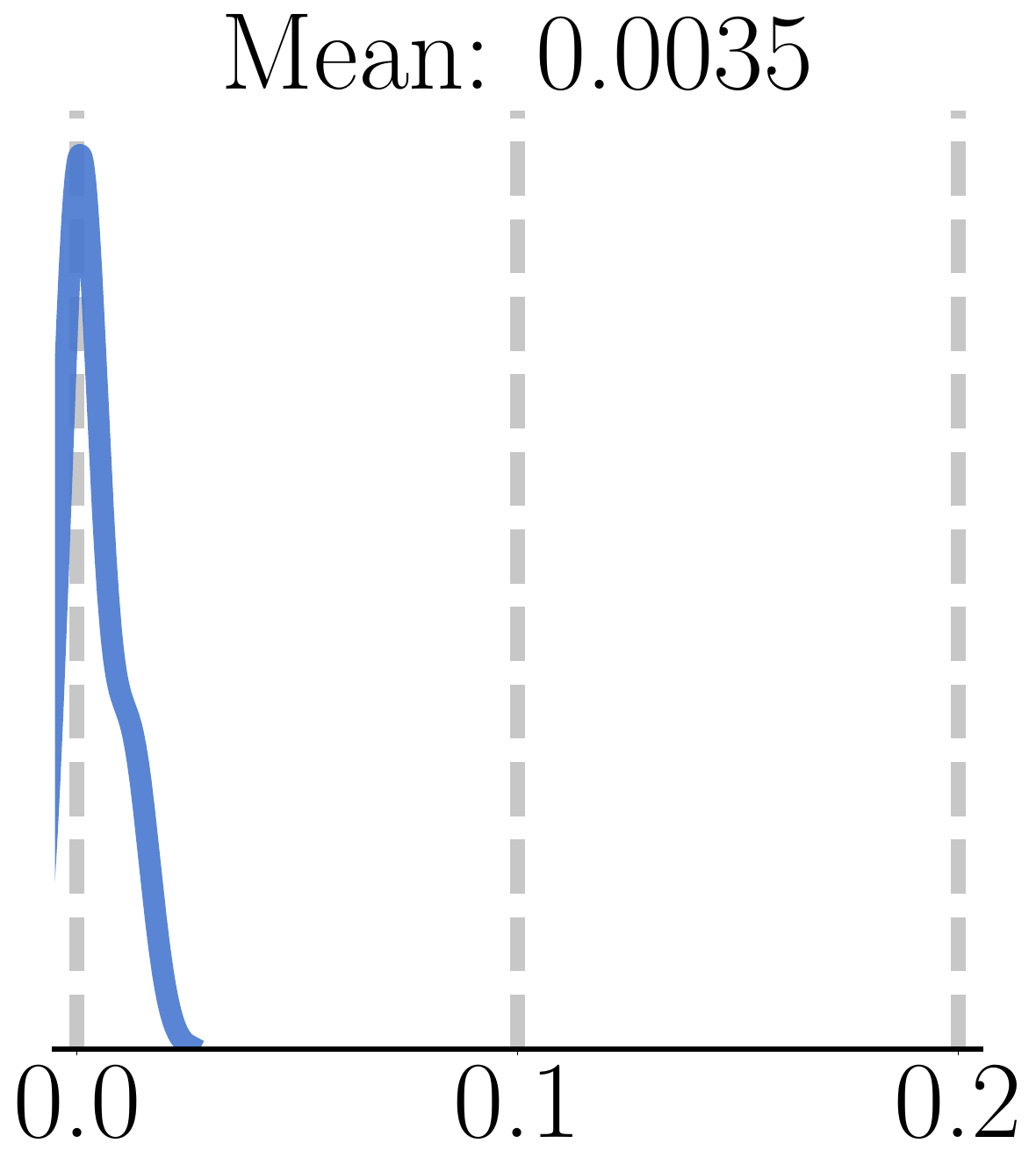}
            \end{subfigure} & &
            \begin{subfigure}
                \centering
                \includegraphics[height=0.5\linewidth]{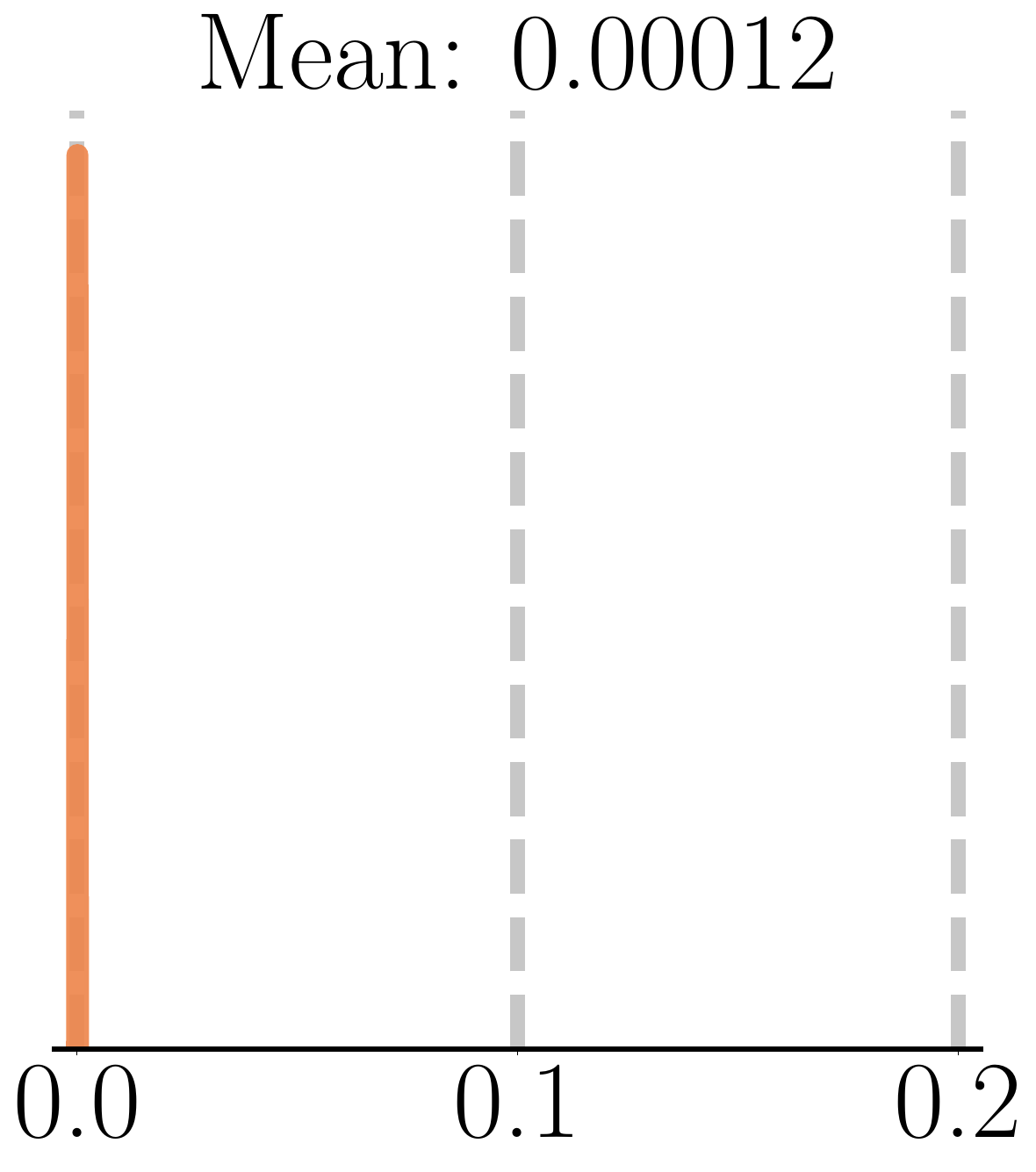}
            \end{subfigure} &
            \begin{subfigure}
                \centering
                \includegraphics[height=0.5\linewidth]{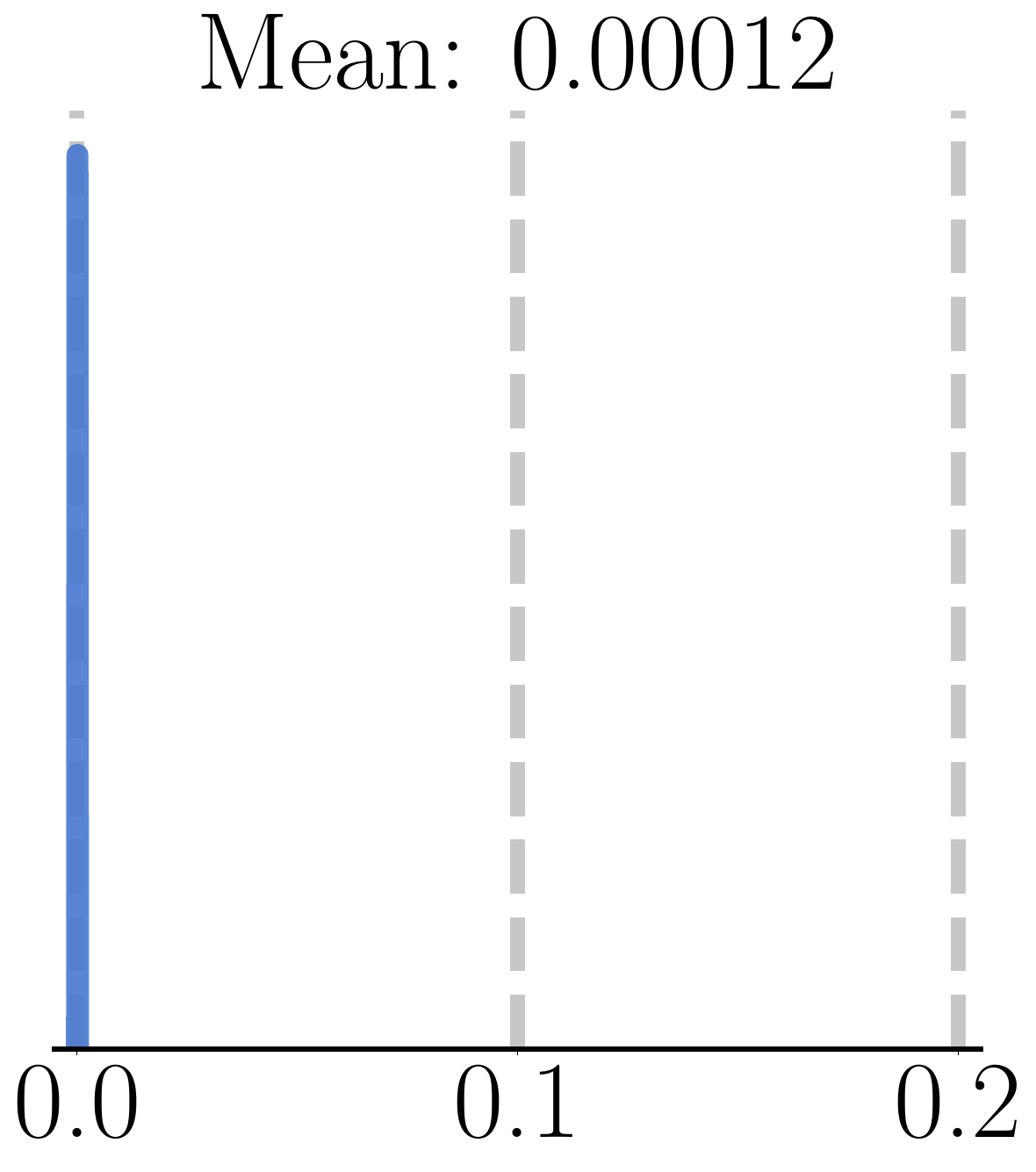}
            \end{subfigure} \\

            \multicolumn{2}{c}{\textbf{Book}} & & \multicolumn{2}{c}{\textbf{Camera}} & & \multicolumn{2}{c}{\textbf{Cellphone}} \\
            \addlinespace[5mm]
            
            \begin{subfigure}
                \centering
                \includegraphics[height=0.5\linewidth]{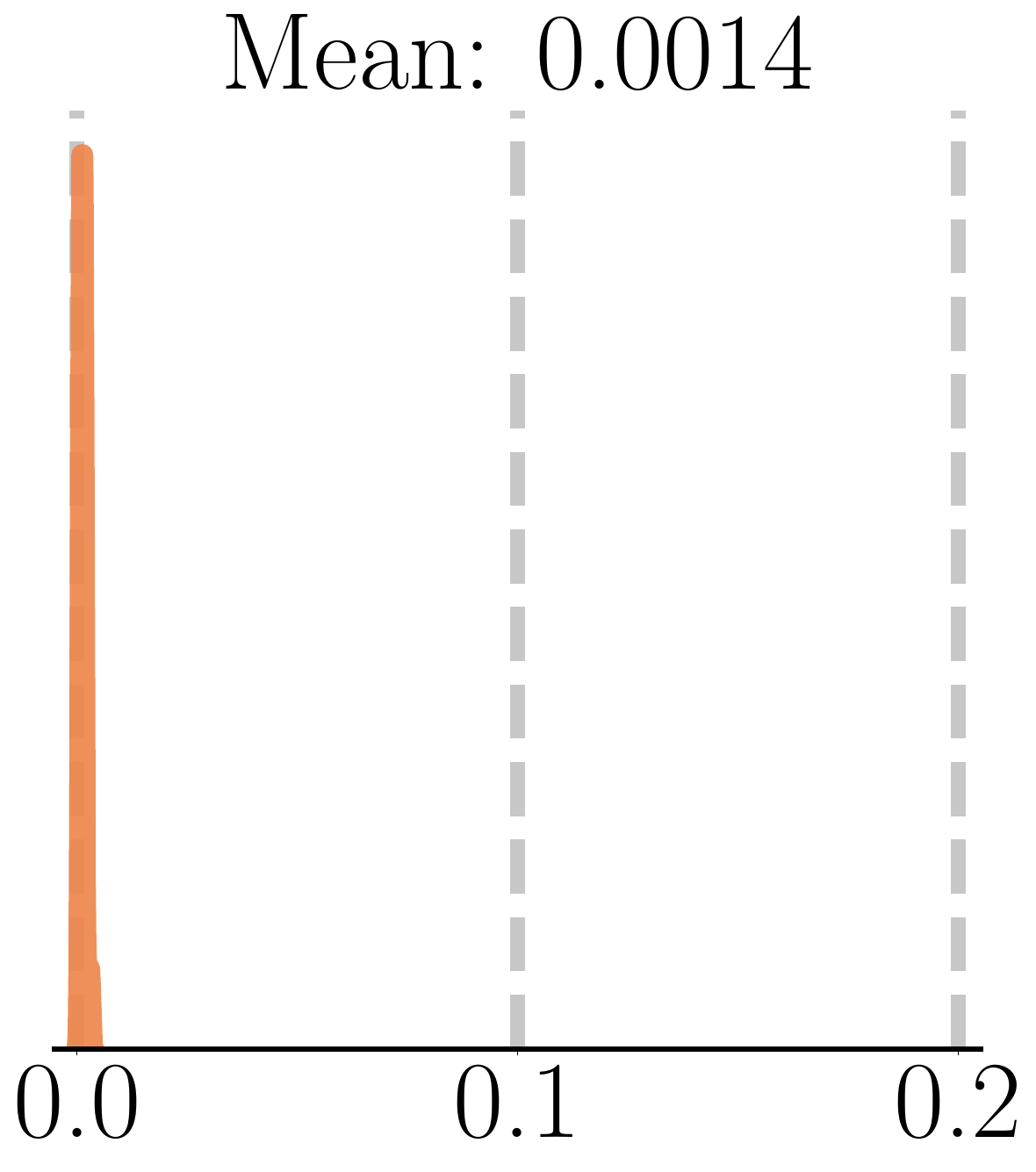}
            \end{subfigure} &
            \begin{subfigure}
                \centering
                \includegraphics[height=0.5\linewidth]{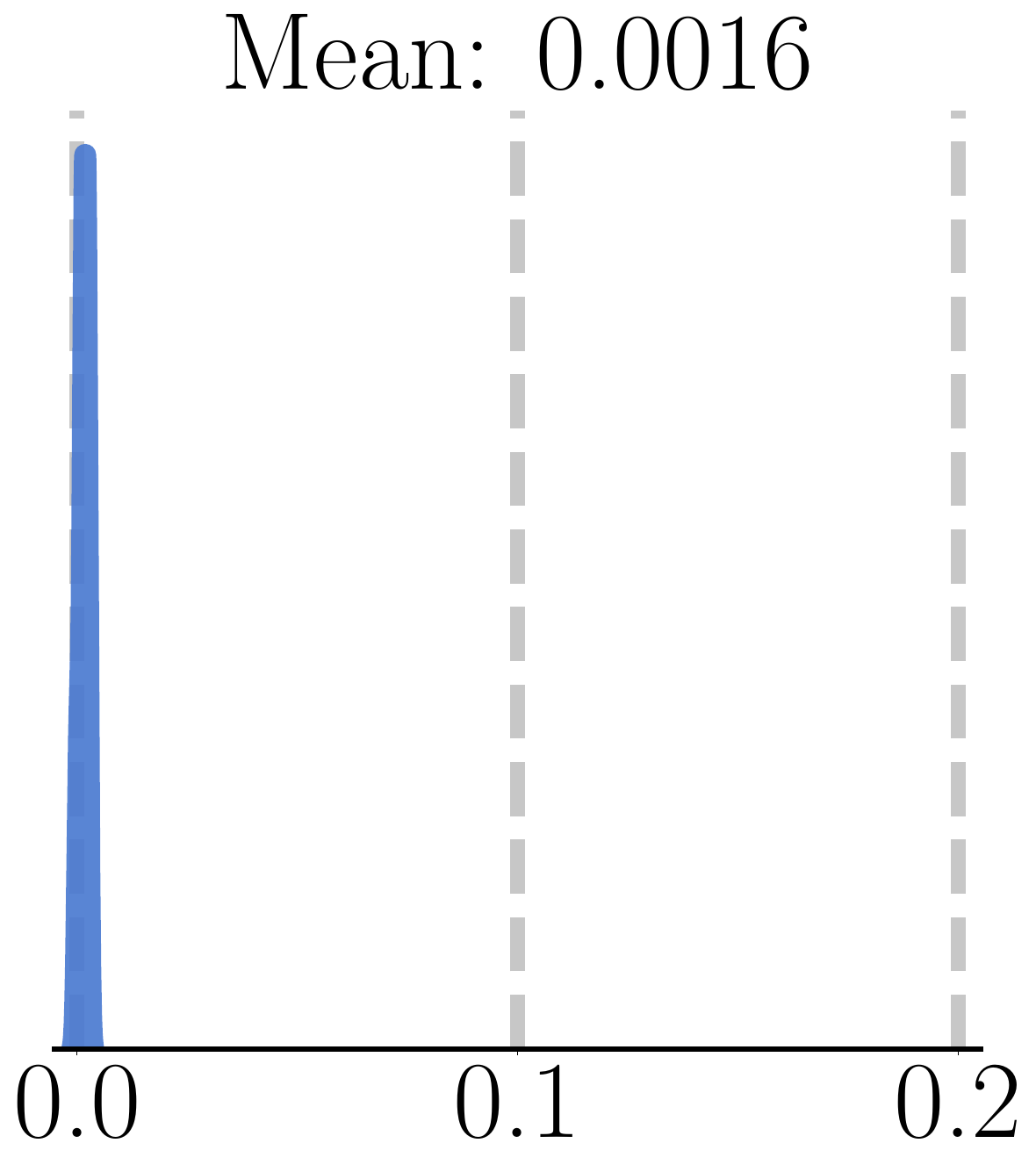}
            \end{subfigure} & &
            \begin{subfigure}
                \centering
                \includegraphics[height=0.5\linewidth]{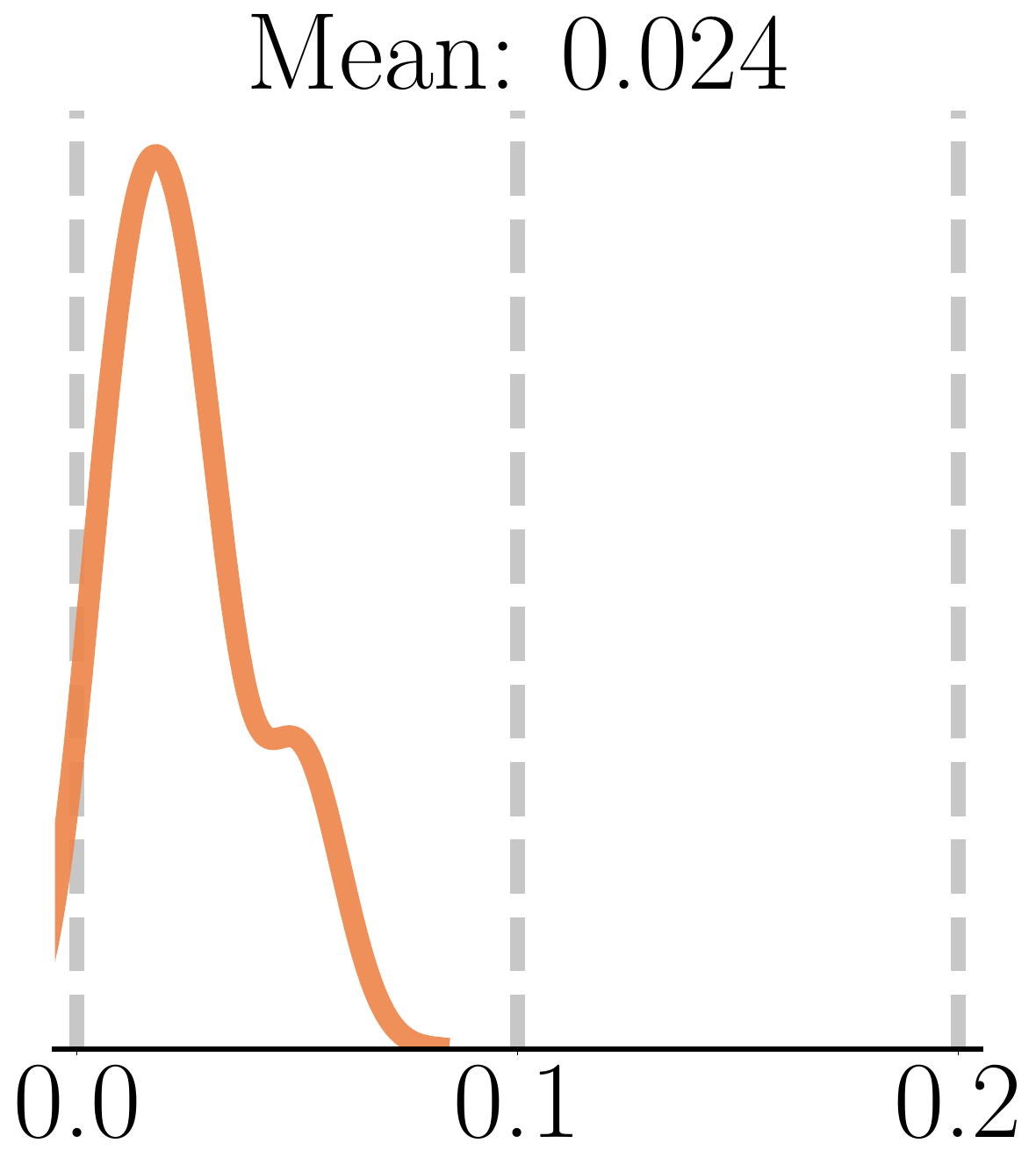}
            \end{subfigure} &
            \begin{subfigure}
                \centering
                \includegraphics[height=0.5\linewidth]{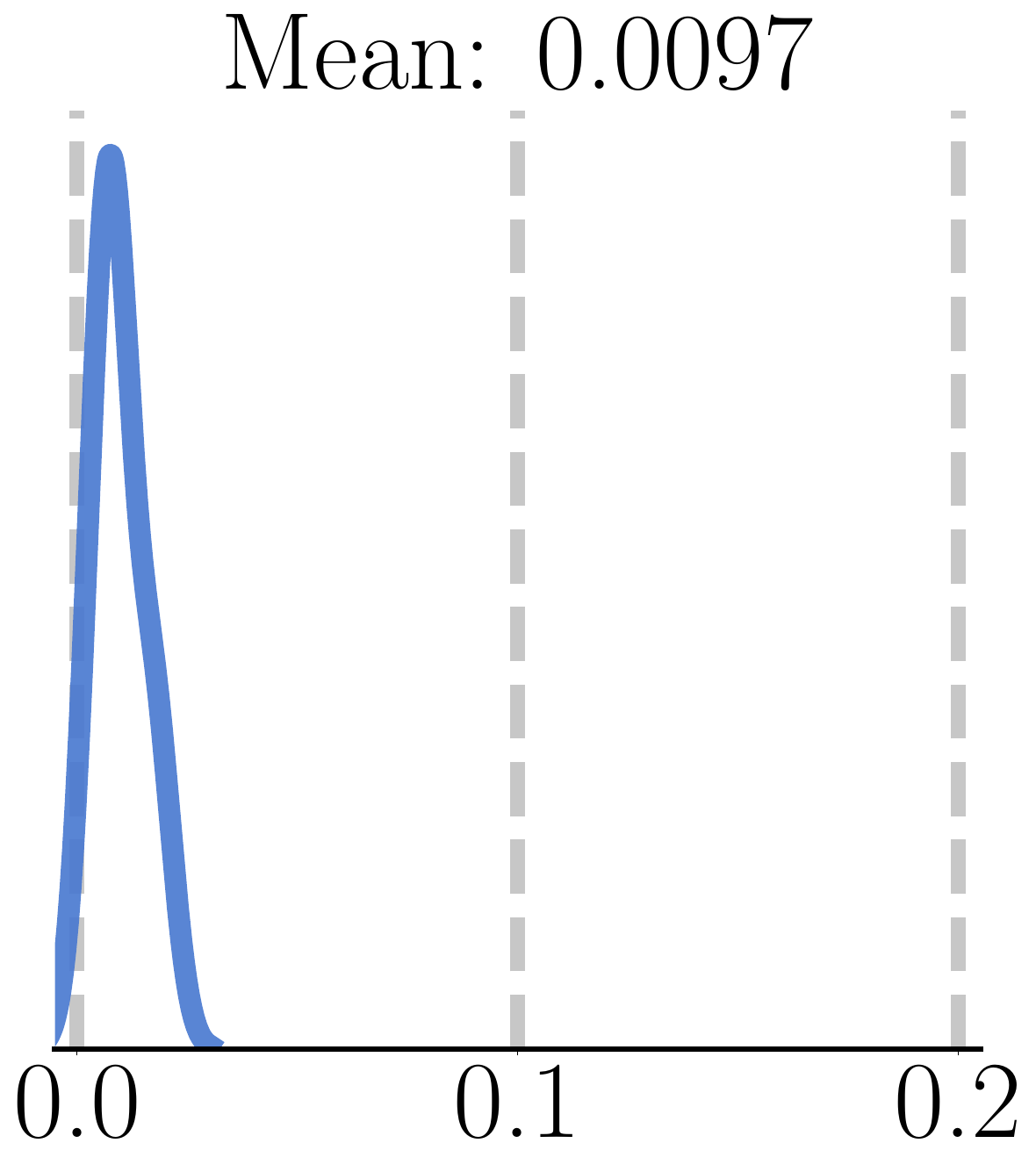}
            \end{subfigure} & &
            \begin{subfigure}
                \centering
                \includegraphics[height=0.5\linewidth]{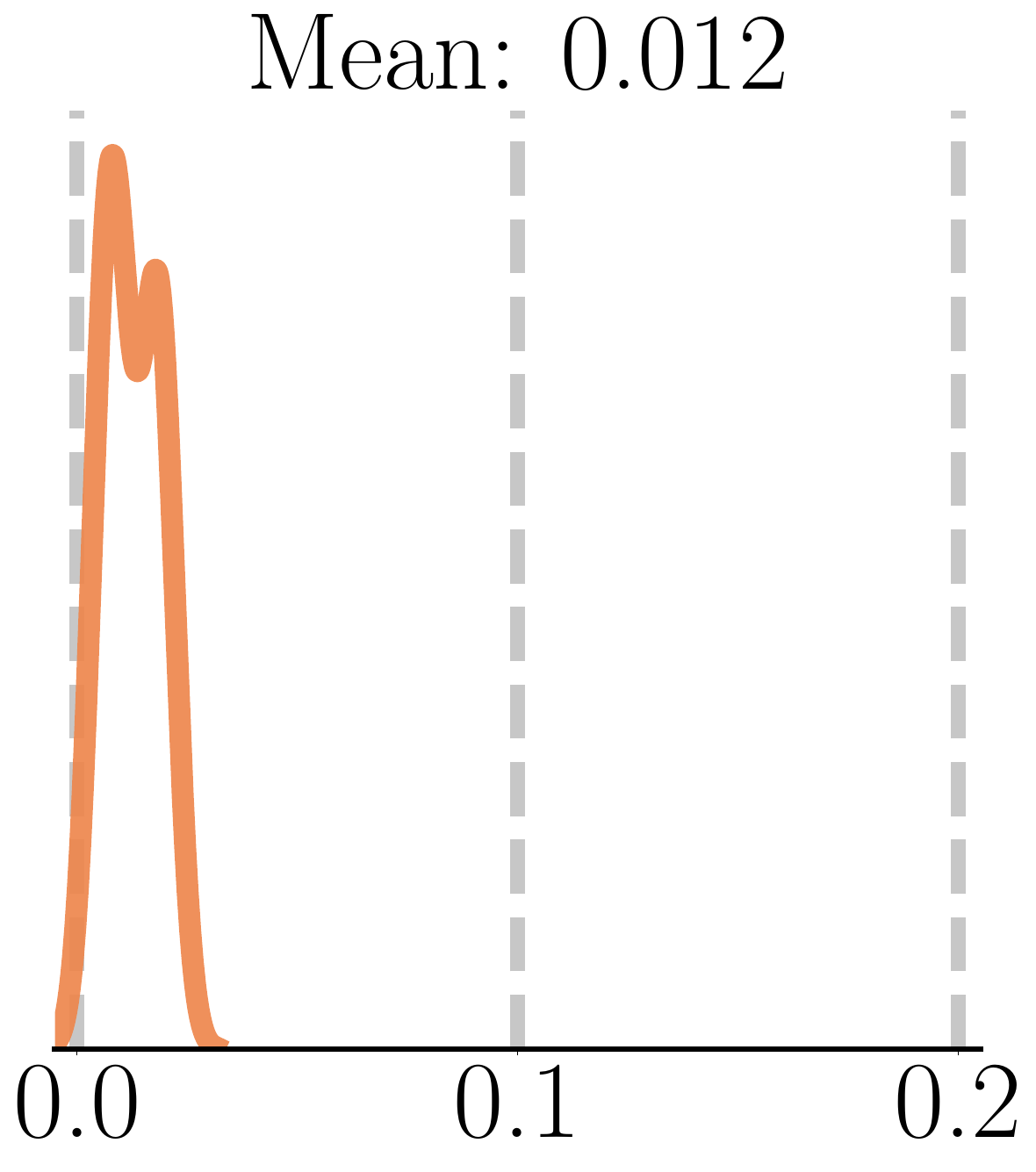}
            \end{subfigure} &
            \begin{subfigure}
                \centering
                \includegraphics[height=0.5\linewidth]{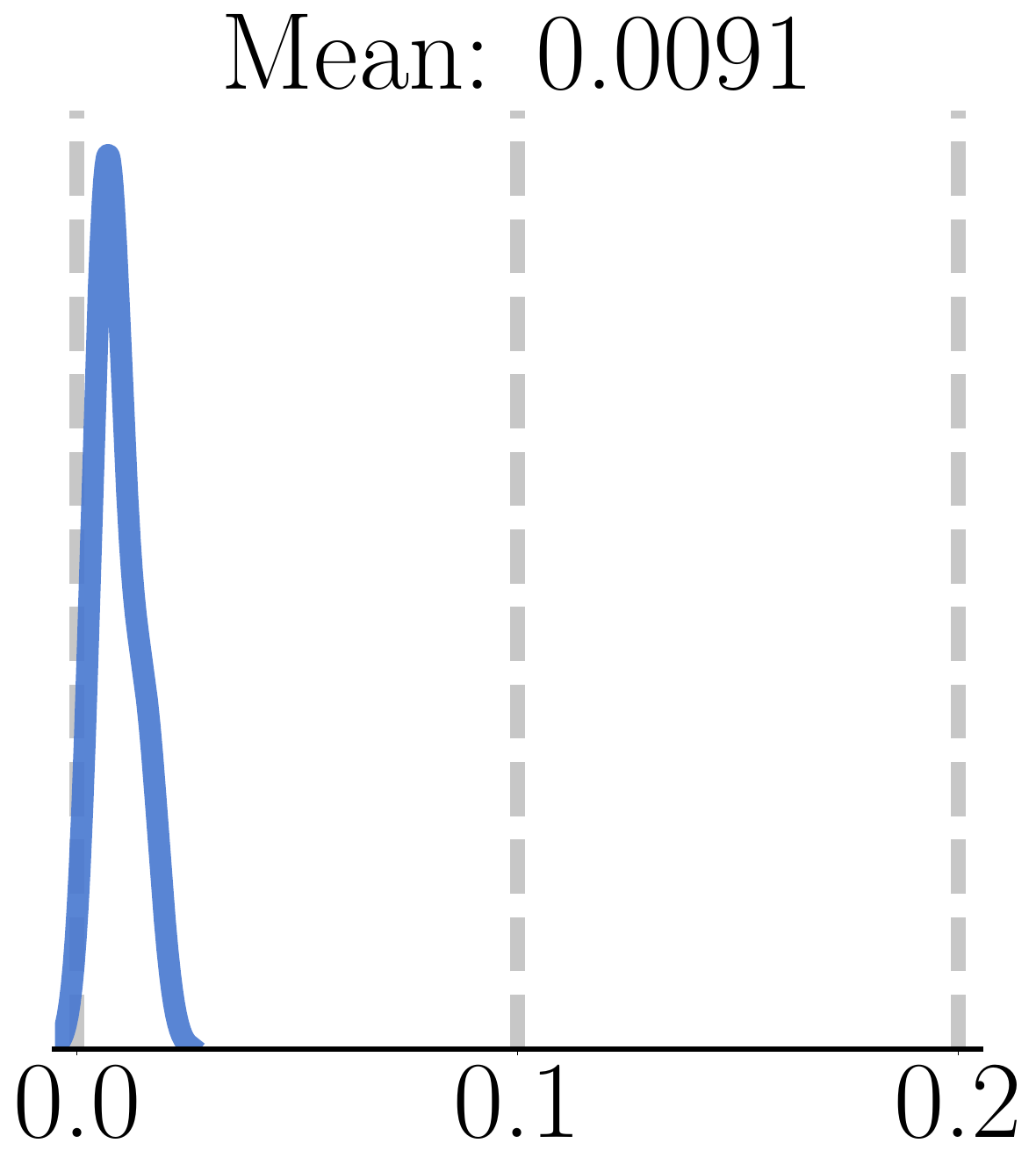}
            \end{subfigure} \\
            
            \multicolumn{2}{c}{\textbf{Eyeglasses}} & & \multicolumn{2}{c}{\textbf{Hat}} & & \multicolumn{2}{c}{\textbf{Headphones}} \\
            \addlinespace[5mm]
            
            \begin{subfigure}
                \centering
                \includegraphics[height=0.5\linewidth]{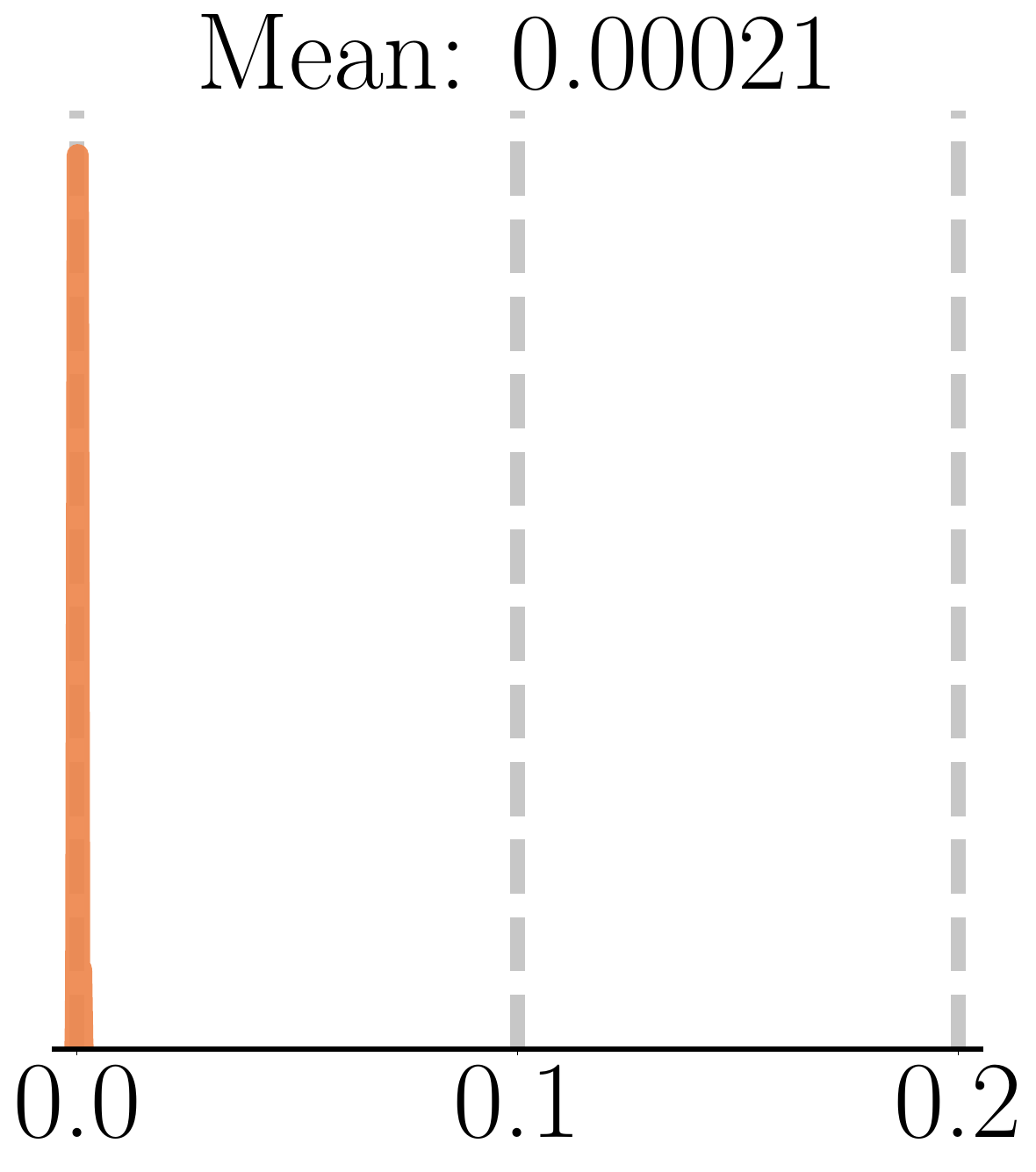}
            \end{subfigure} &
            \begin{subfigure}
                \centering
                \includegraphics[height=0.5\linewidth]{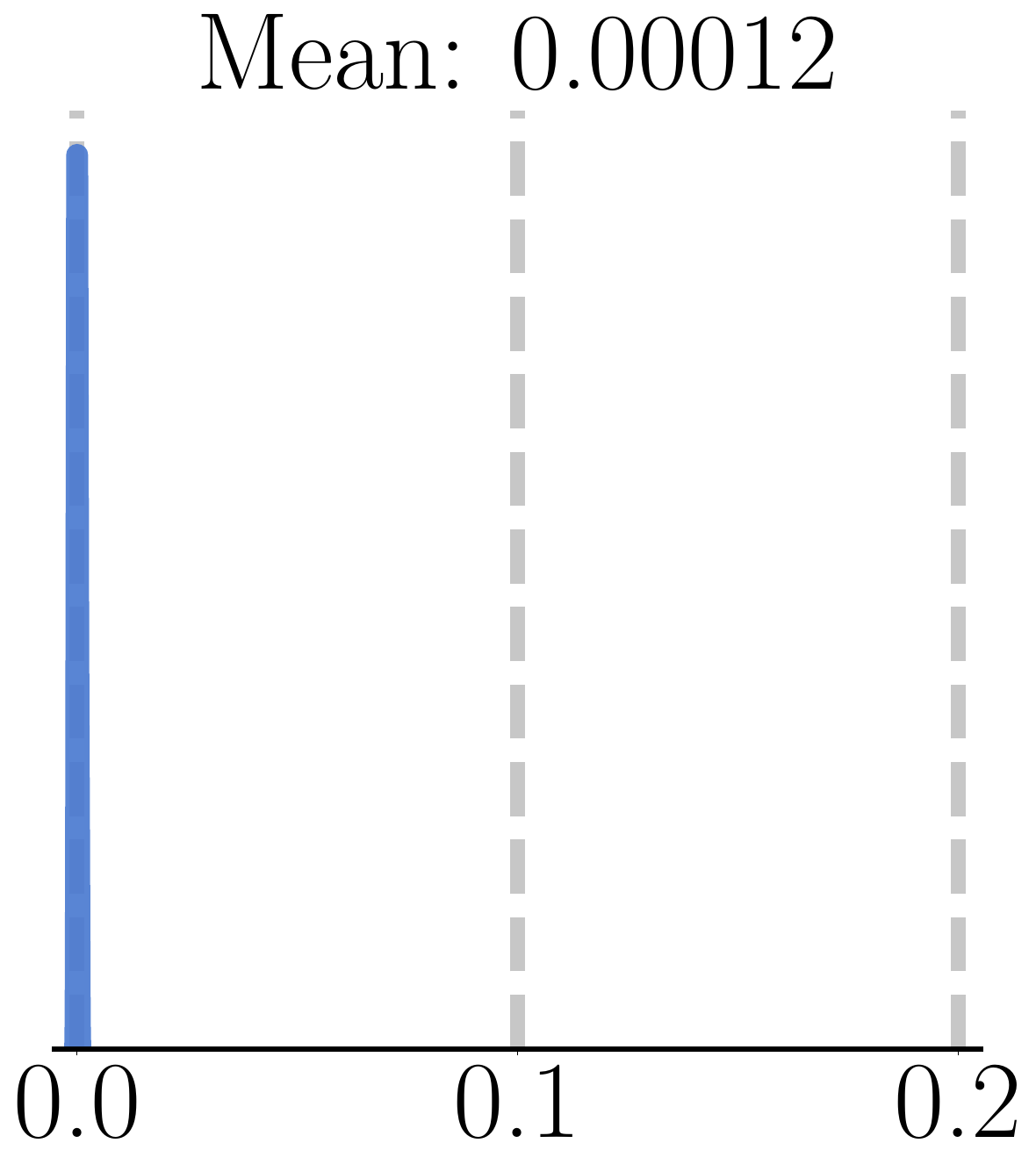}
            \end{subfigure} & &
            \begin{subfigure}
                \centering
                \includegraphics[height=0.5\linewidth]{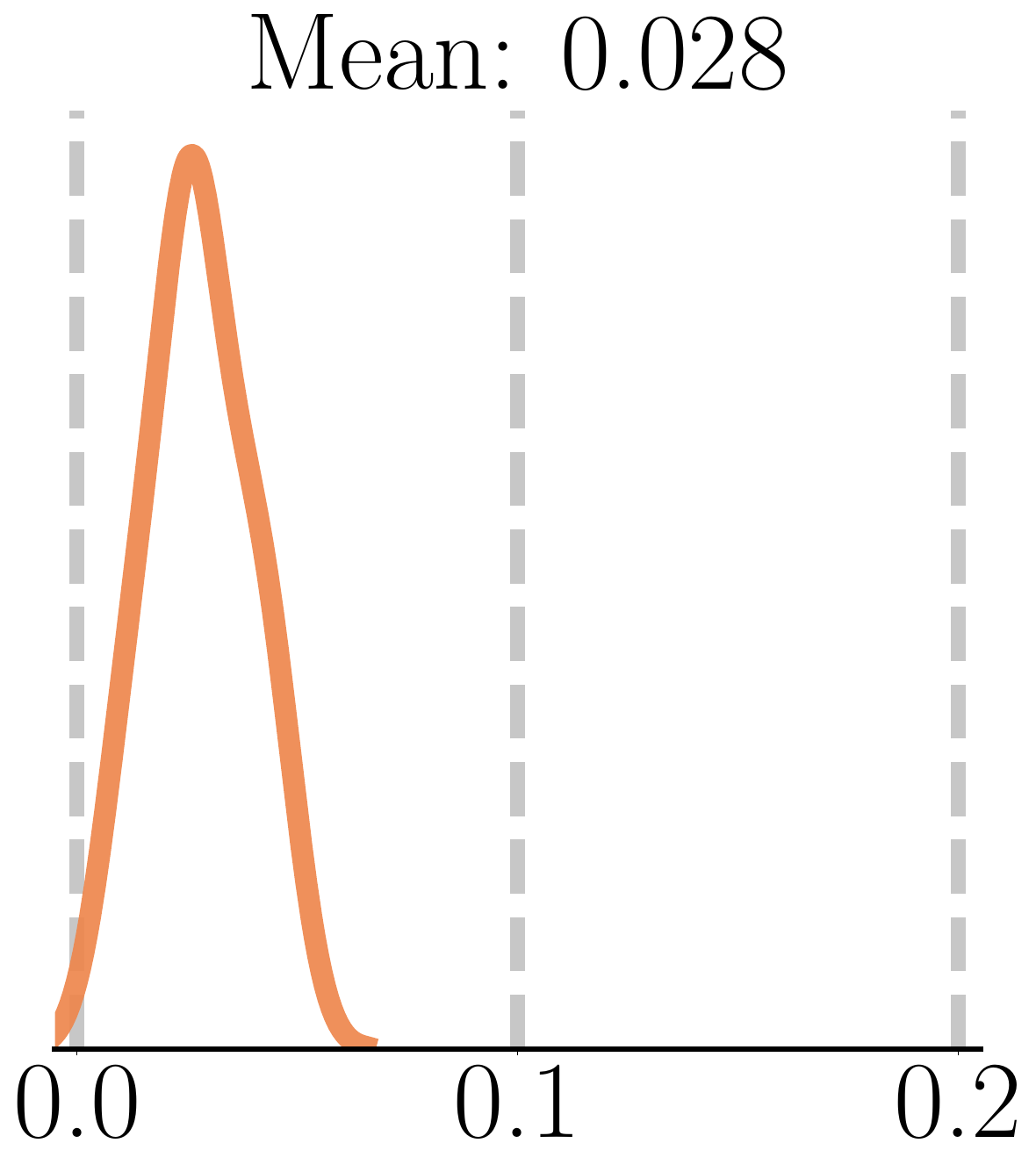}
            \end{subfigure} &
            \begin{subfigure}
                \centering
                \includegraphics[height=0.5\linewidth]{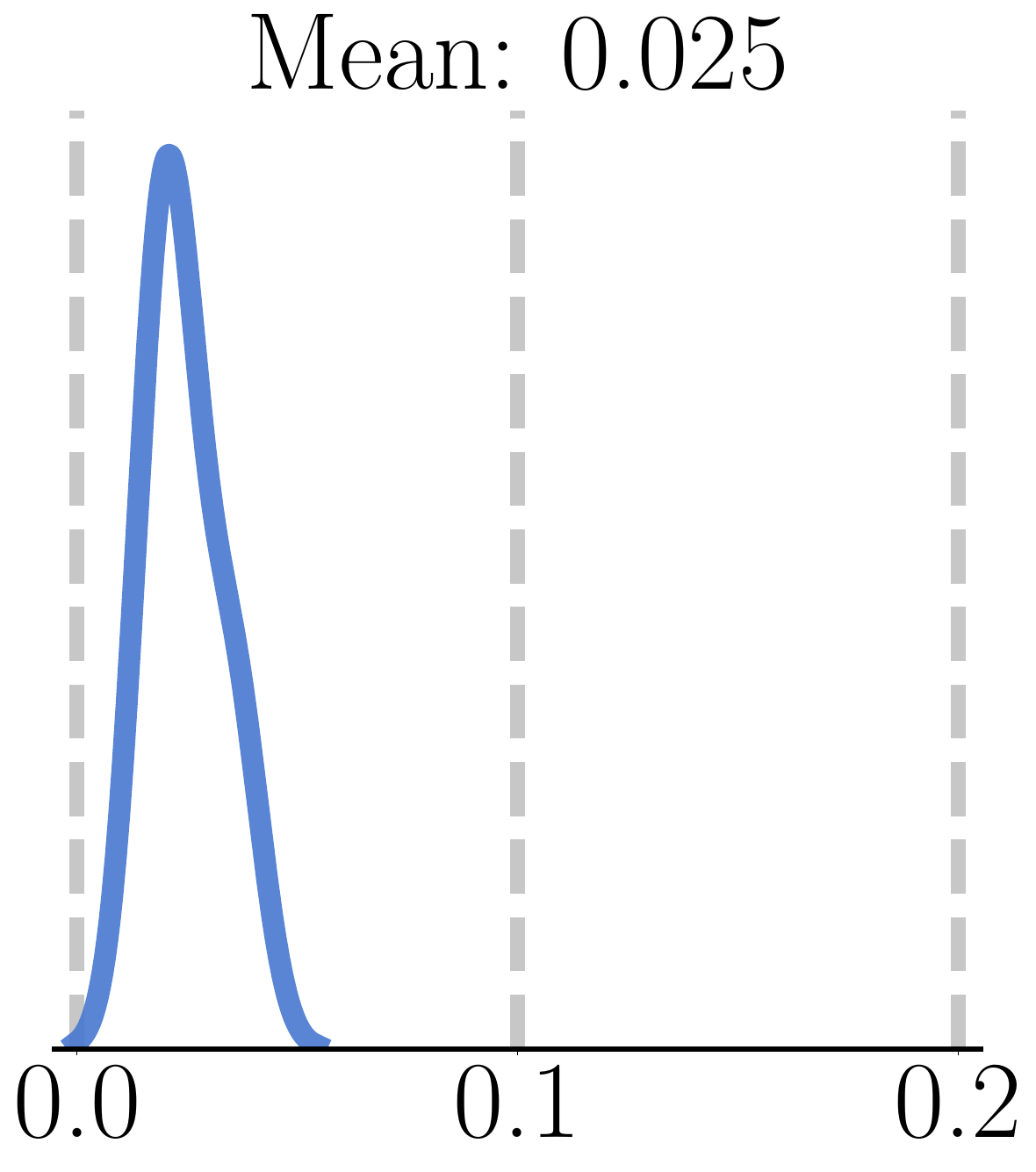}
            \end{subfigure} & &
            \begin{subfigure}
                \centering
                \includegraphics[height=0.5\linewidth]{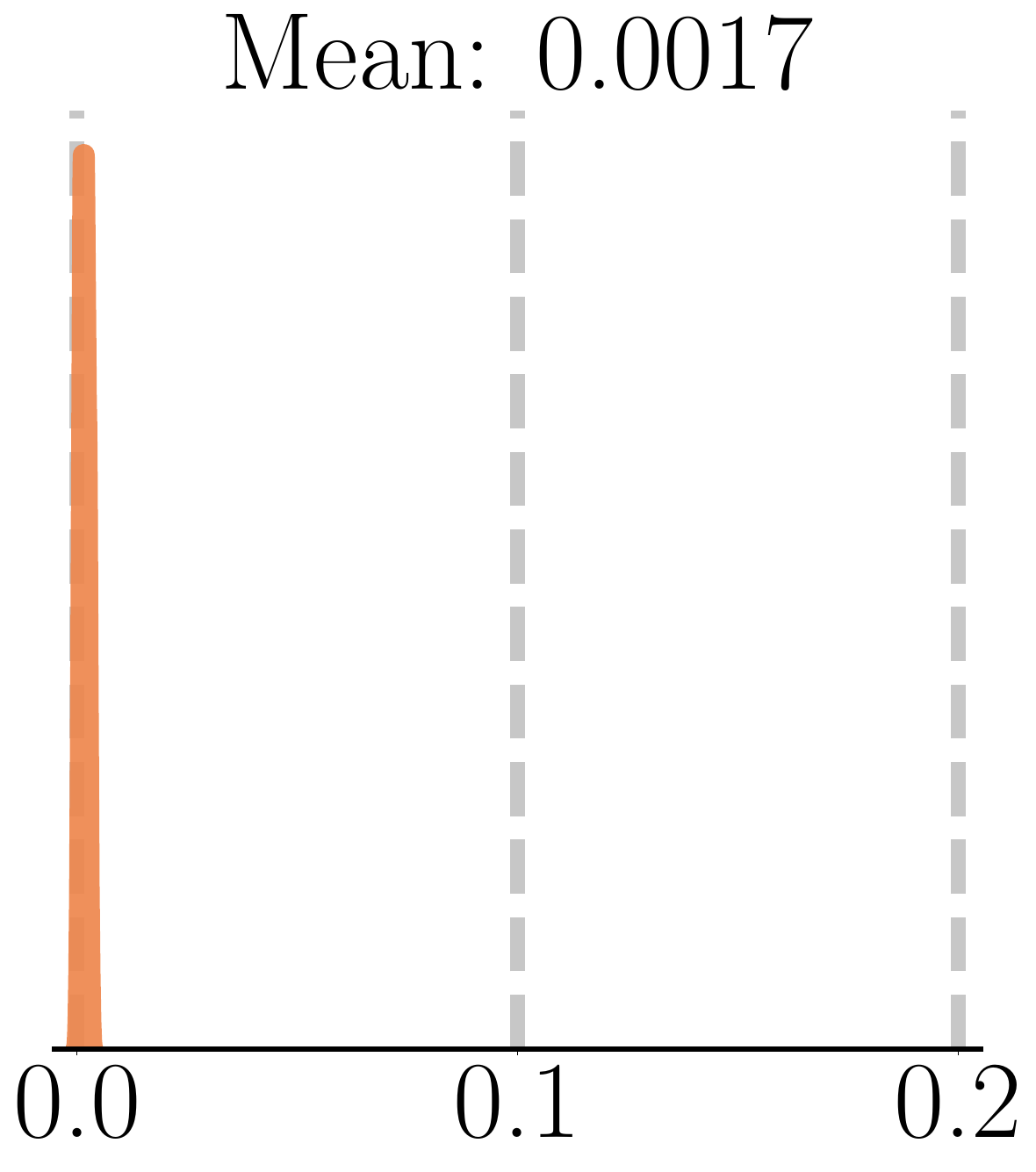}
            \end{subfigure} &
            \begin{subfigure}
                \centering
                \includegraphics[height=0.5\linewidth]{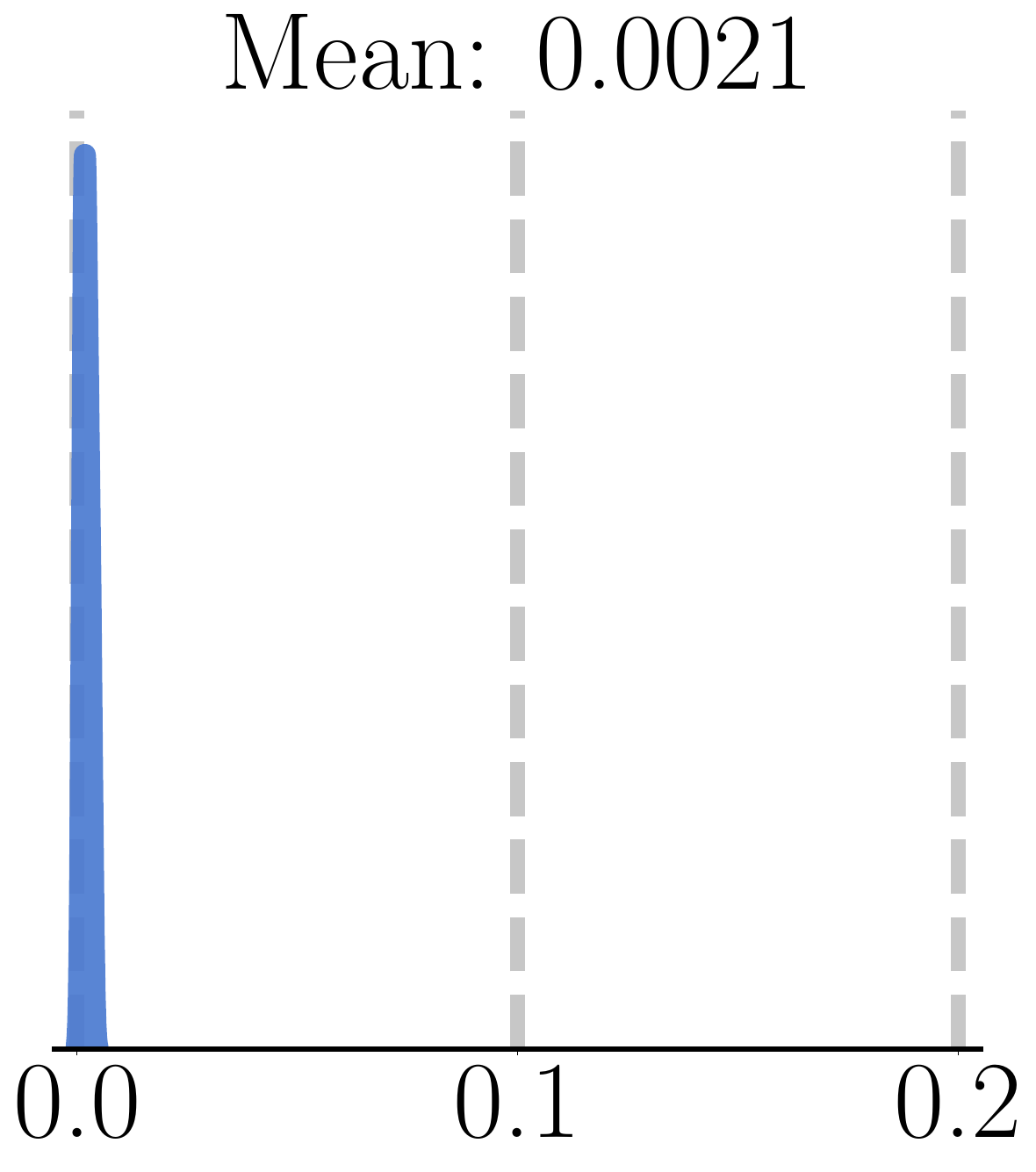}
            \end{subfigure} \\
            
            \multicolumn{2}{c}{\textbf{Keys}} & & \multicolumn{2}{c}{\textbf{Laptop}} & & \multicolumn{2}{c}{\textbf{Mug}} \\
            \addlinespace[5mm]
            
            \begin{subfigure}
                \centering
                \includegraphics[height=0.5\linewidth]{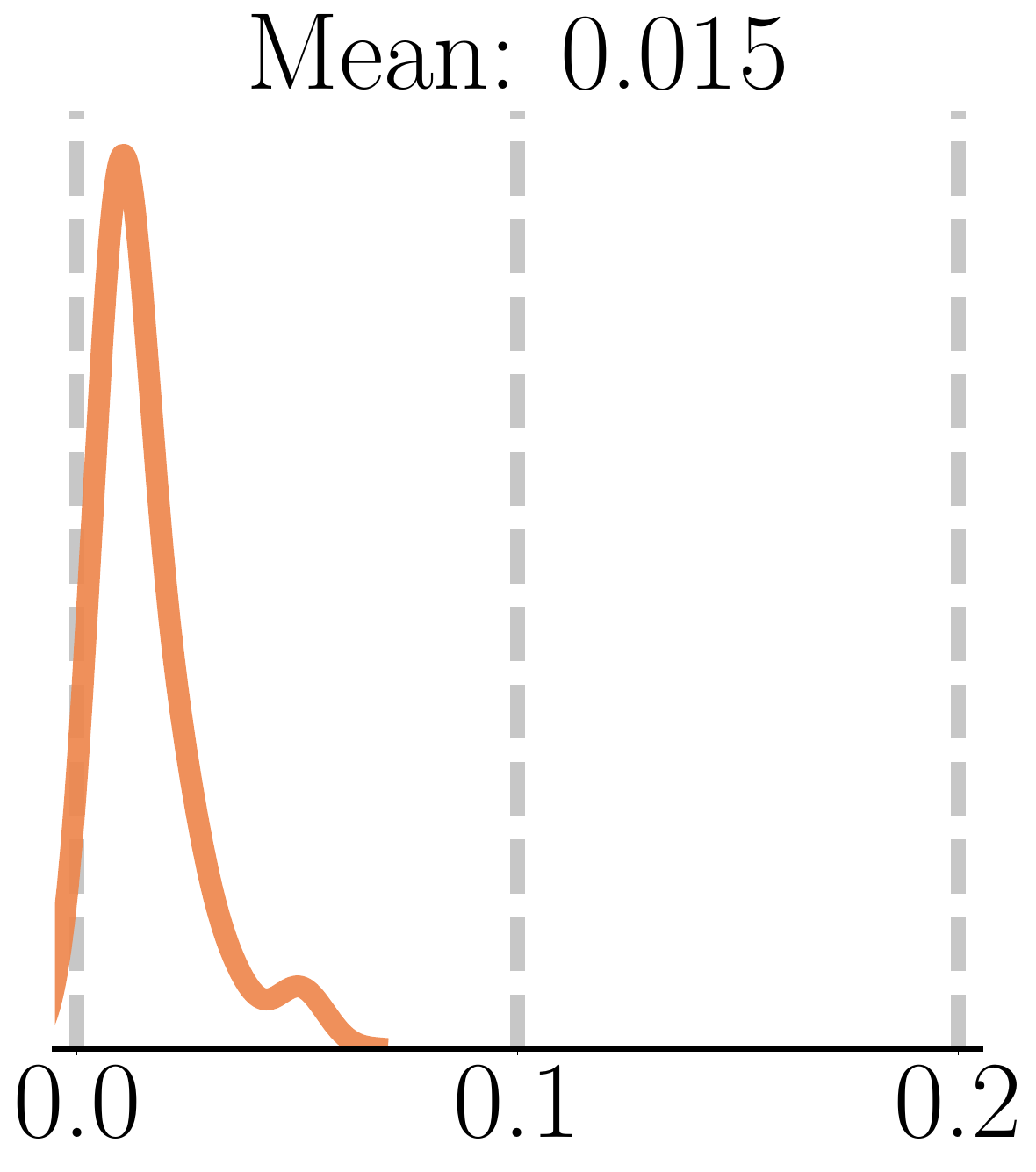}
            \end{subfigure} &
            \begin{subfigure}
                \centering
                \includegraphics[height=0.5\linewidth]{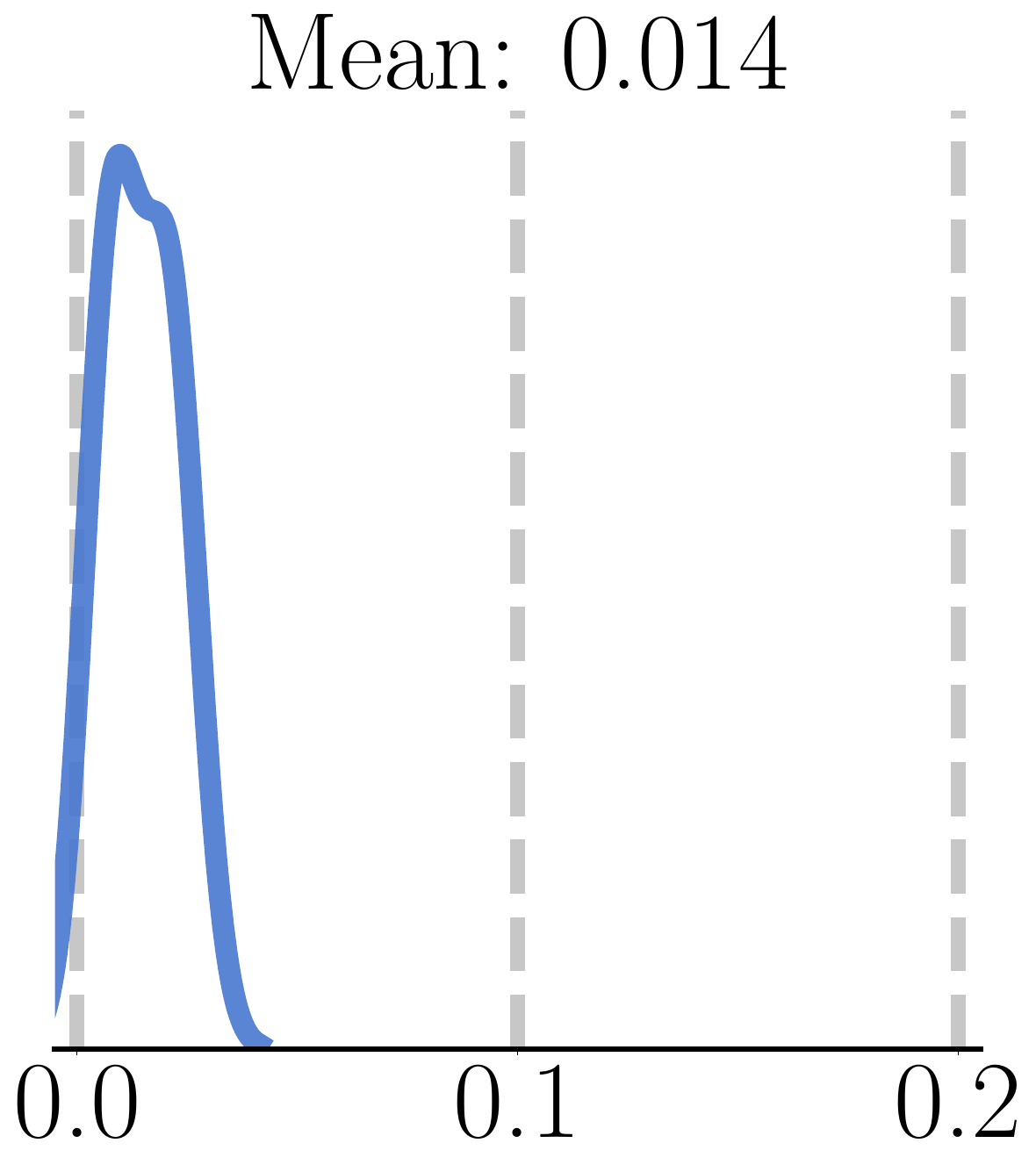}
            \end{subfigure} & &
            \begin{subfigure}
                \centering
                \includegraphics[height=0.5\linewidth]{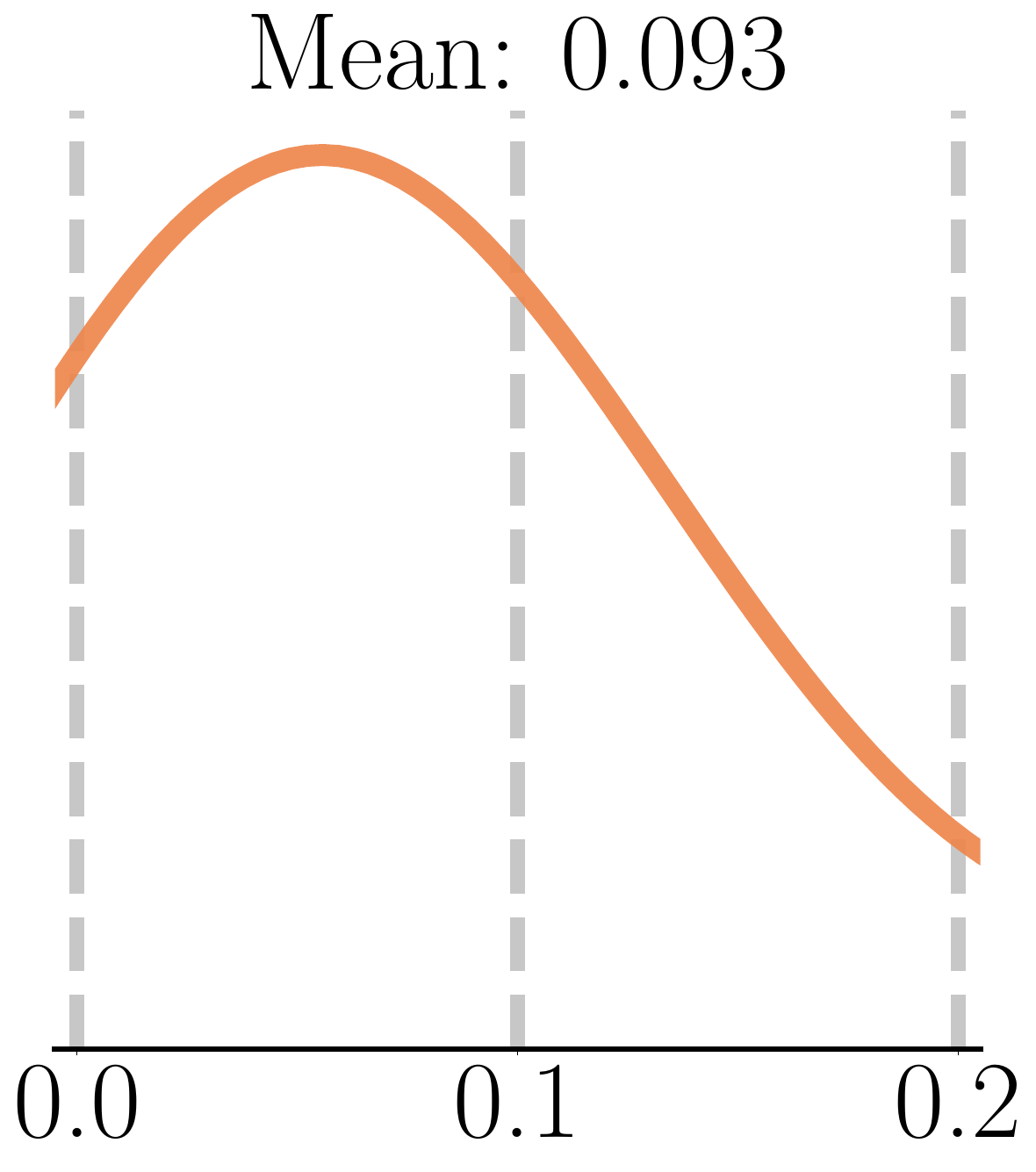}
            \end{subfigure} &
            \begin{subfigure}
                \centering
                \includegraphics[height=0.5\linewidth]{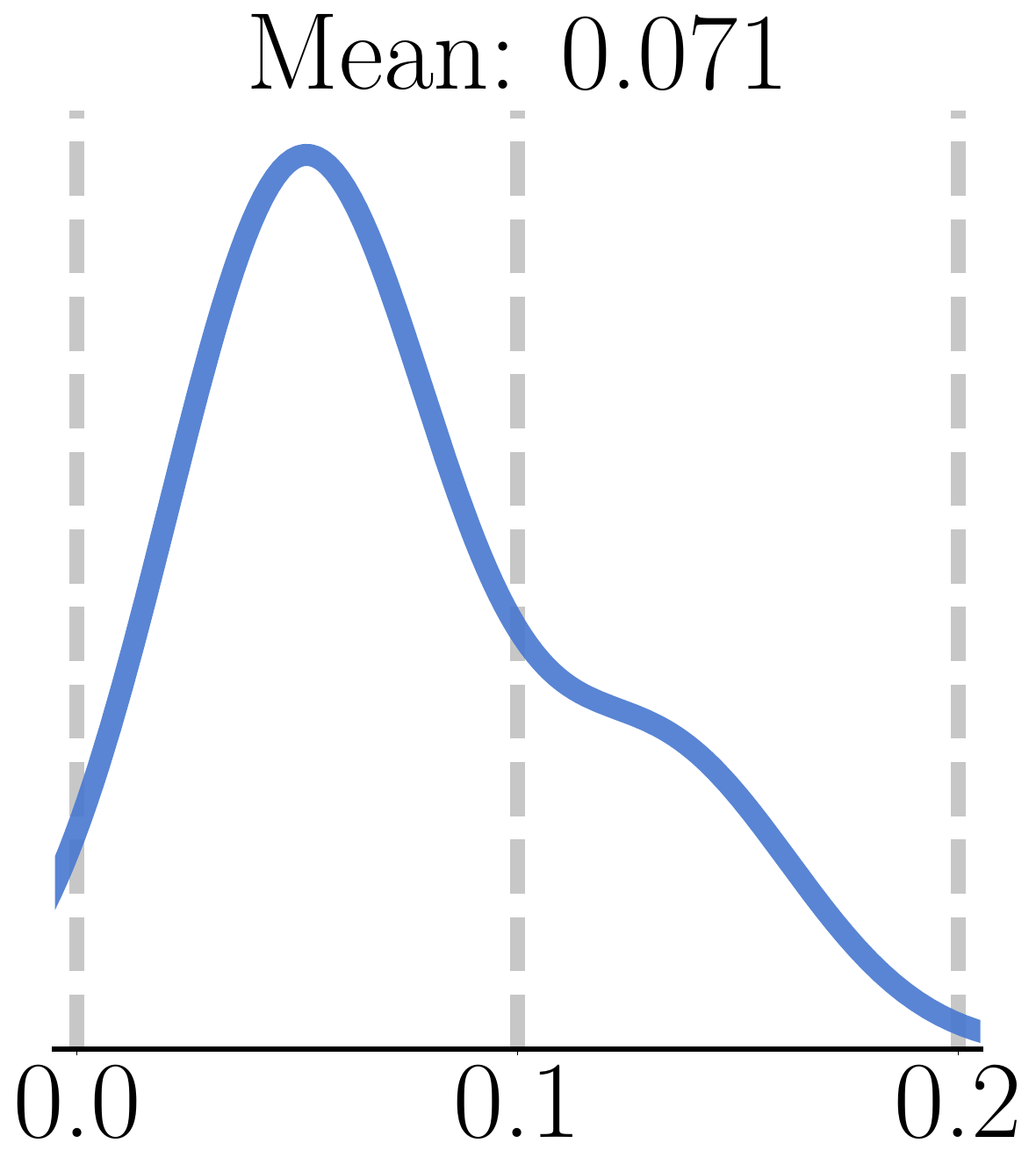}
            \end{subfigure} & &
            \begin{subfigure}
                \centering
                \includegraphics[height=0.5\linewidth]{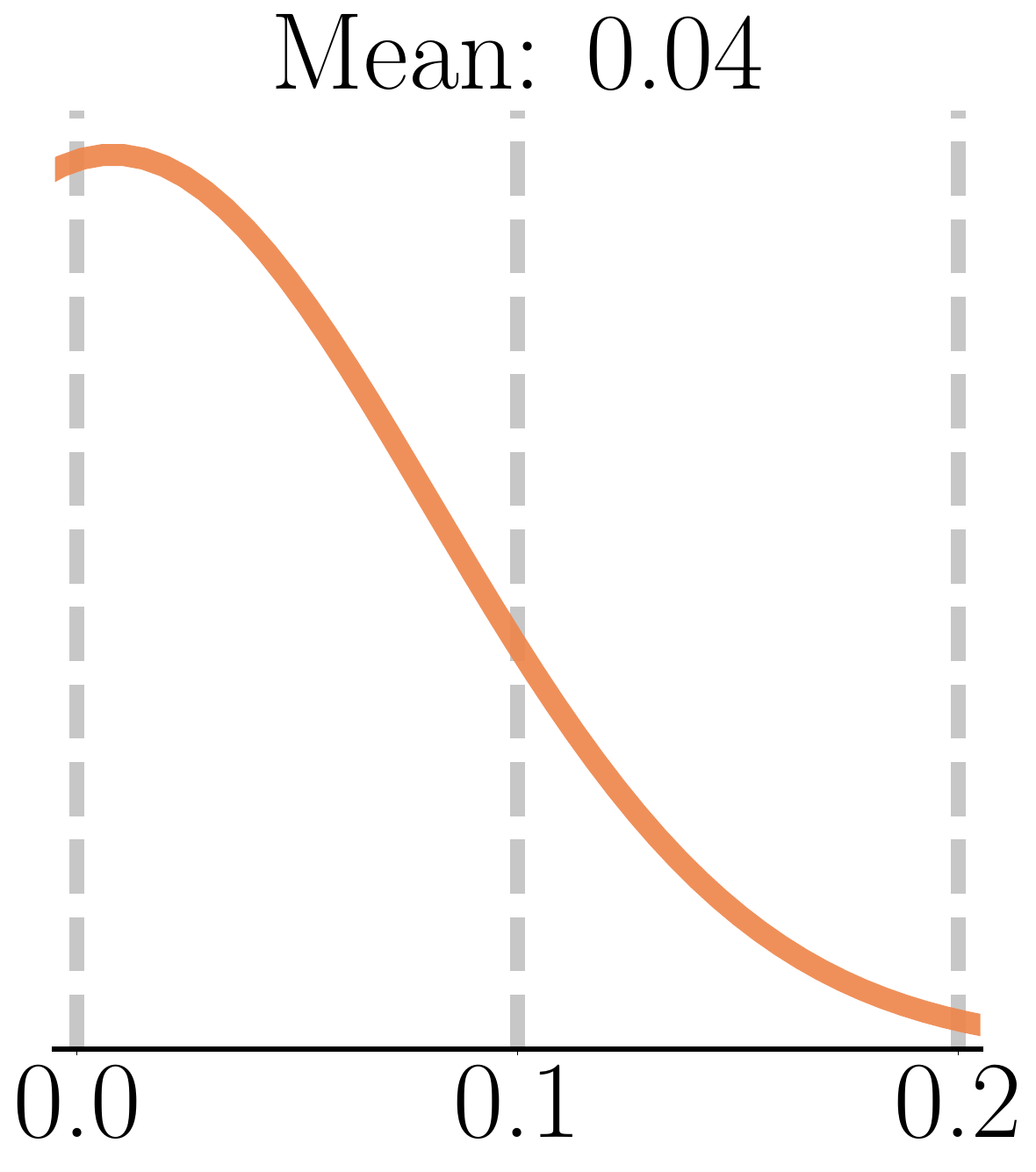}
            \end{subfigure} &
            \begin{subfigure}
                \centering
                \includegraphics[height=0.5\linewidth]{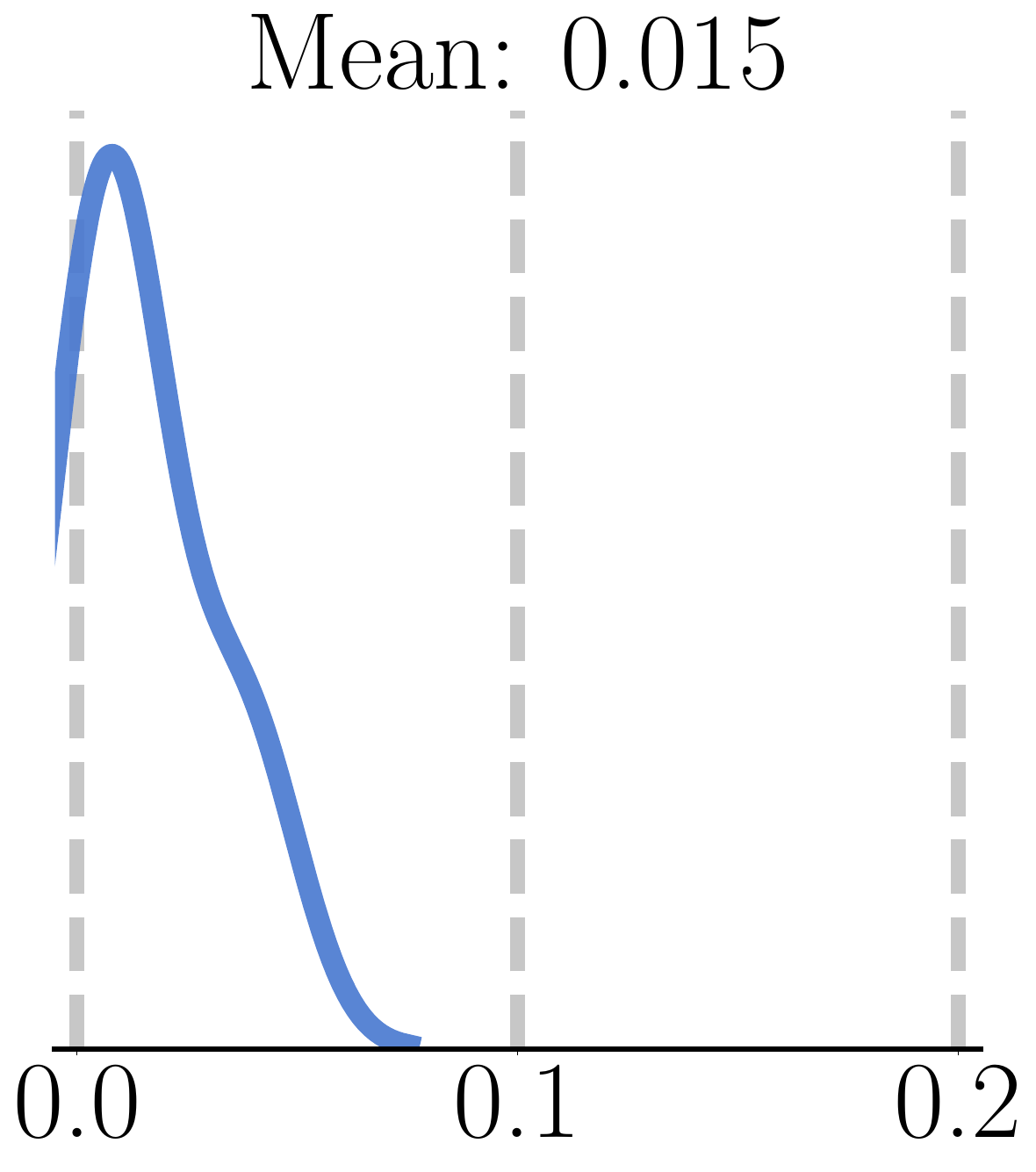}
            \end{subfigure} \\
            
            \multicolumn{2}{c}{\textbf{Shoes}} & & \multicolumn{2}{c}{\textbf{Teddy Bear}} & & \multicolumn{2}{c}{\textbf{Toy}} \\
            \addlinespace[5mm]
            
            \begin{subfigure}
                \centering
                \includegraphics[height=0.5\linewidth]{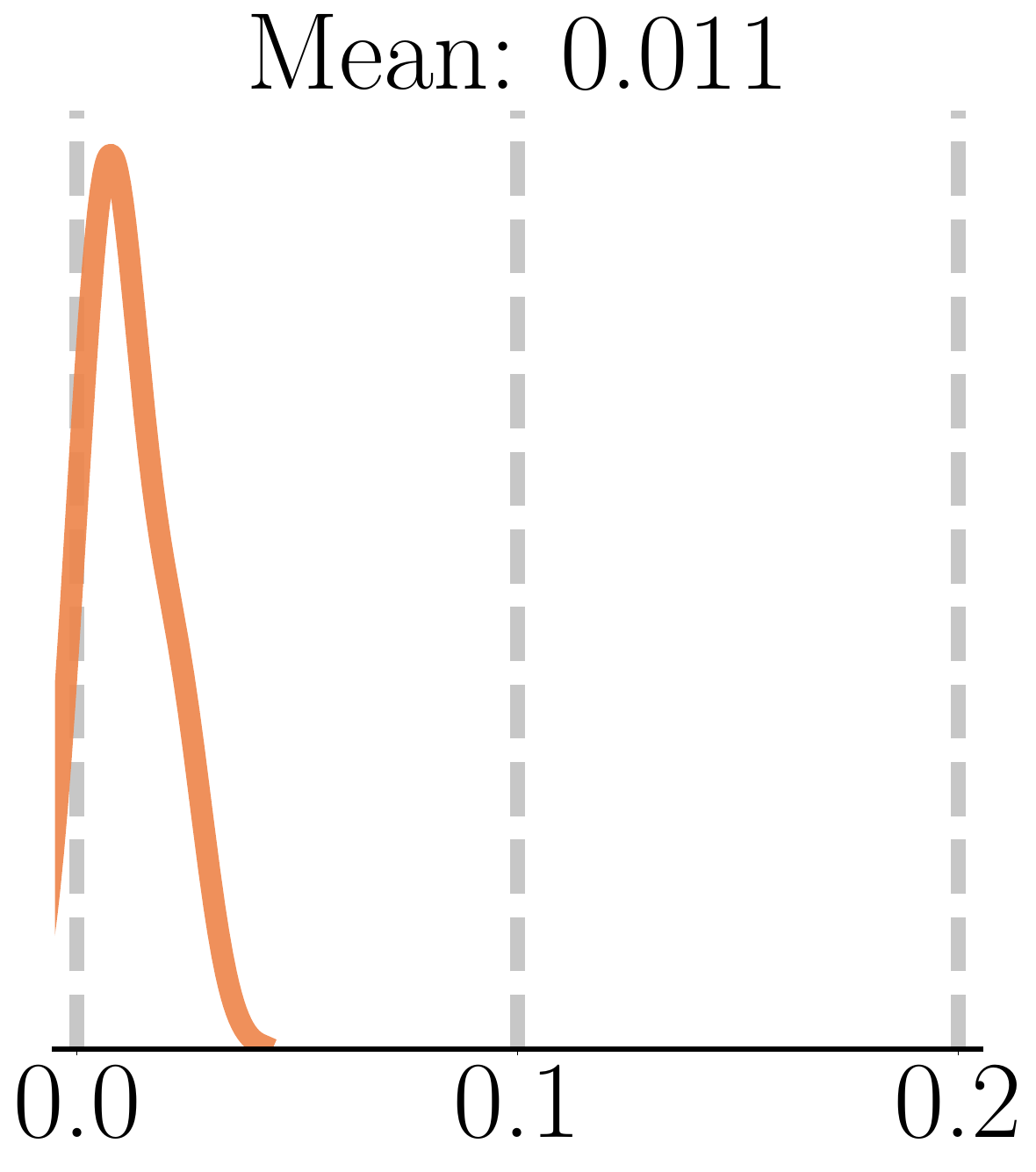}
            \end{subfigure} &
            \begin{subfigure}
                \centering
                \includegraphics[height=0.5\linewidth]{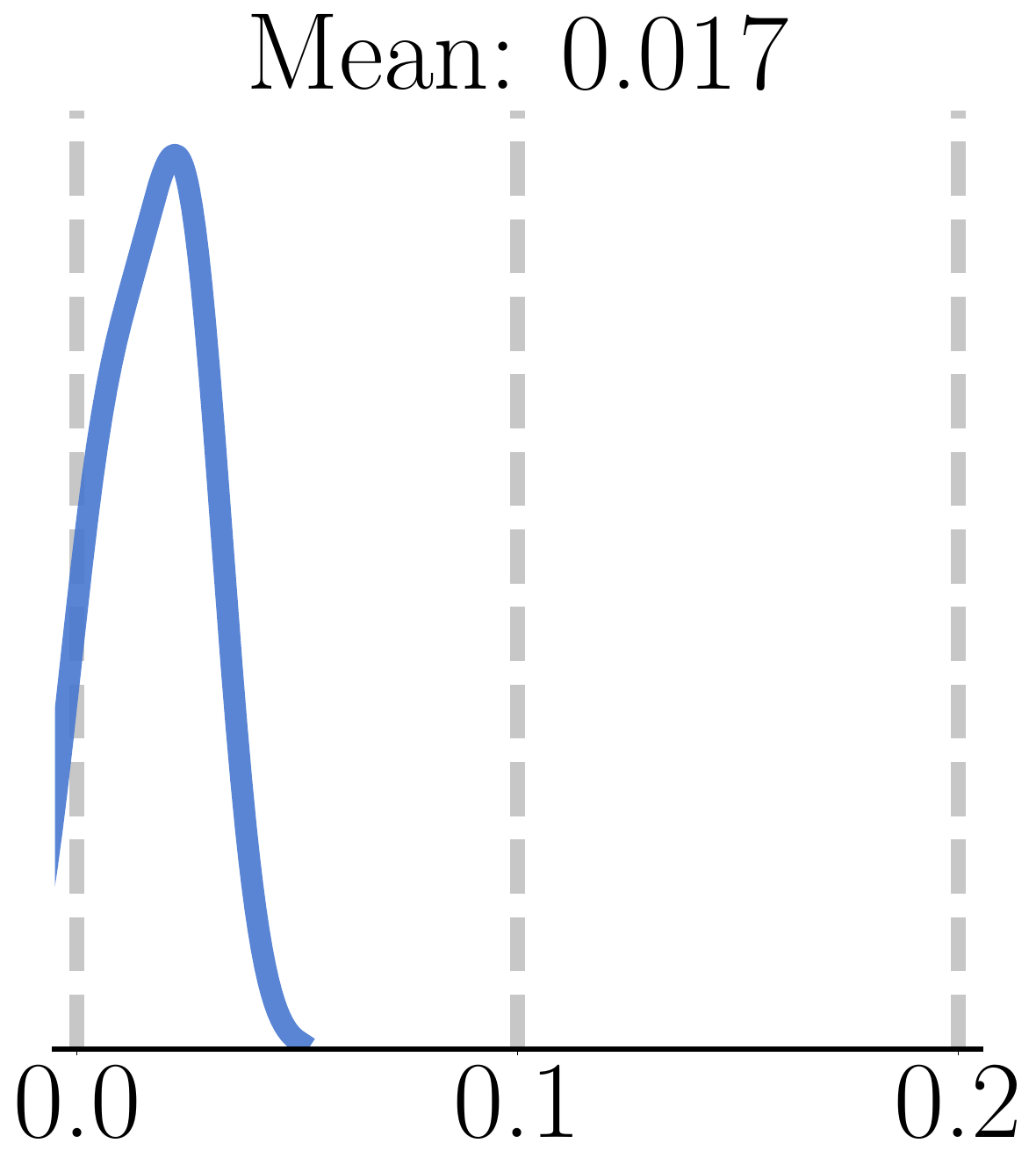}
            \end{subfigure} & &
            \begin{subfigure}
                \centering
                \includegraphics[height=0.5\linewidth]{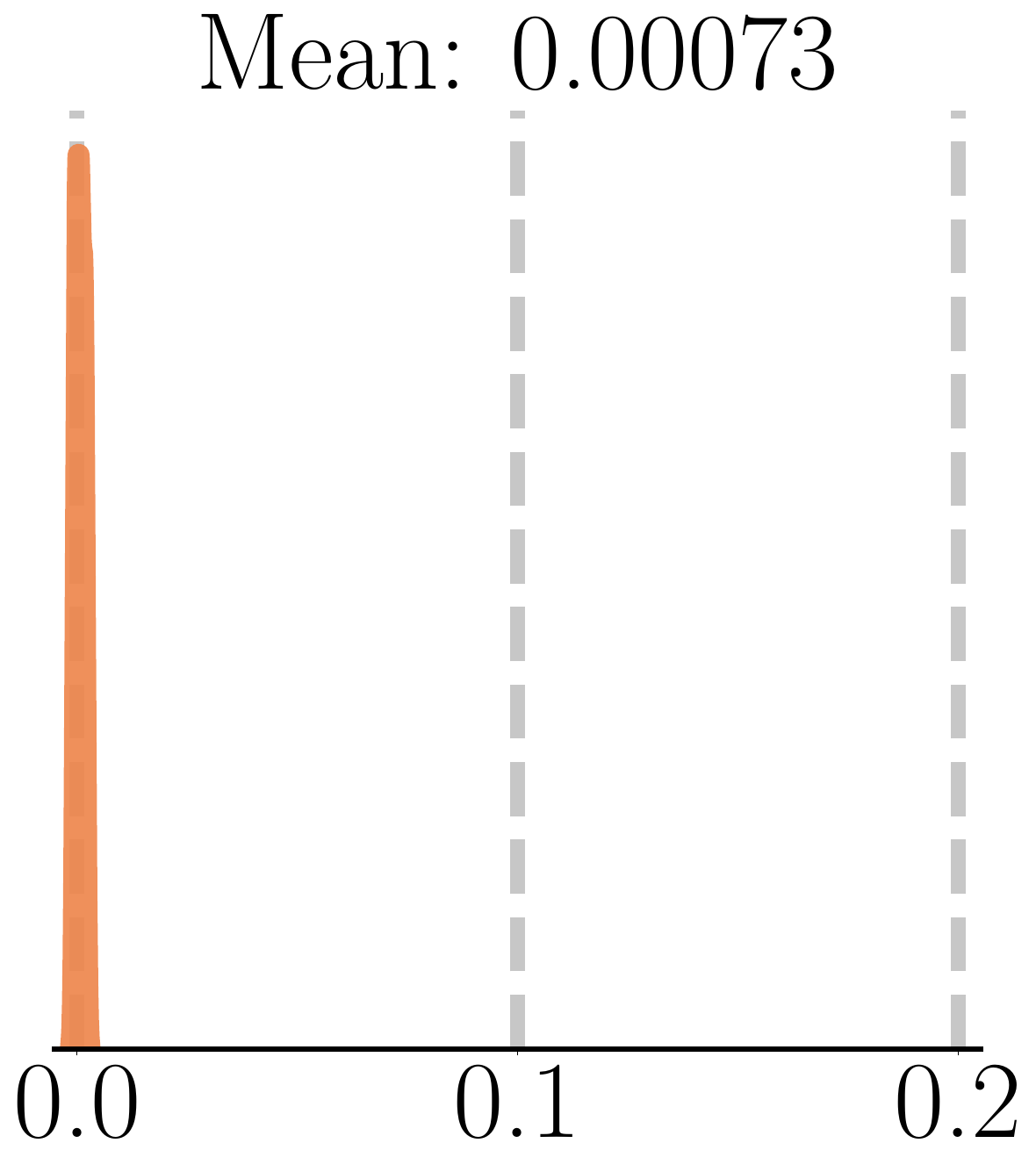}
            \end{subfigure} &
            \begin{subfigure}
                \centering
                \includegraphics[height=0.5\linewidth]{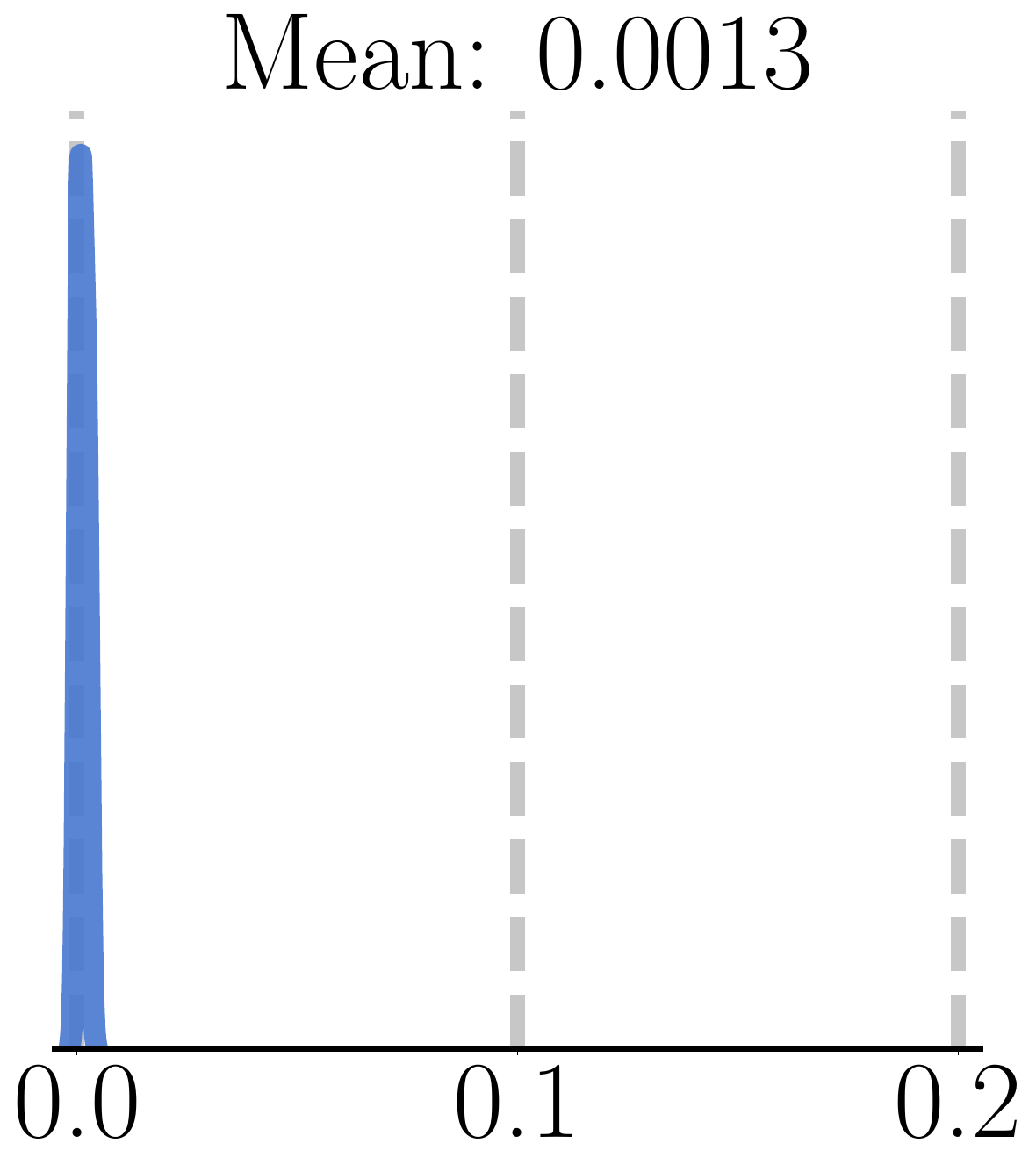}
            \end{subfigure} & &
            \begin{subfigure}
                \centering
                \includegraphics[height=0.5\linewidth]{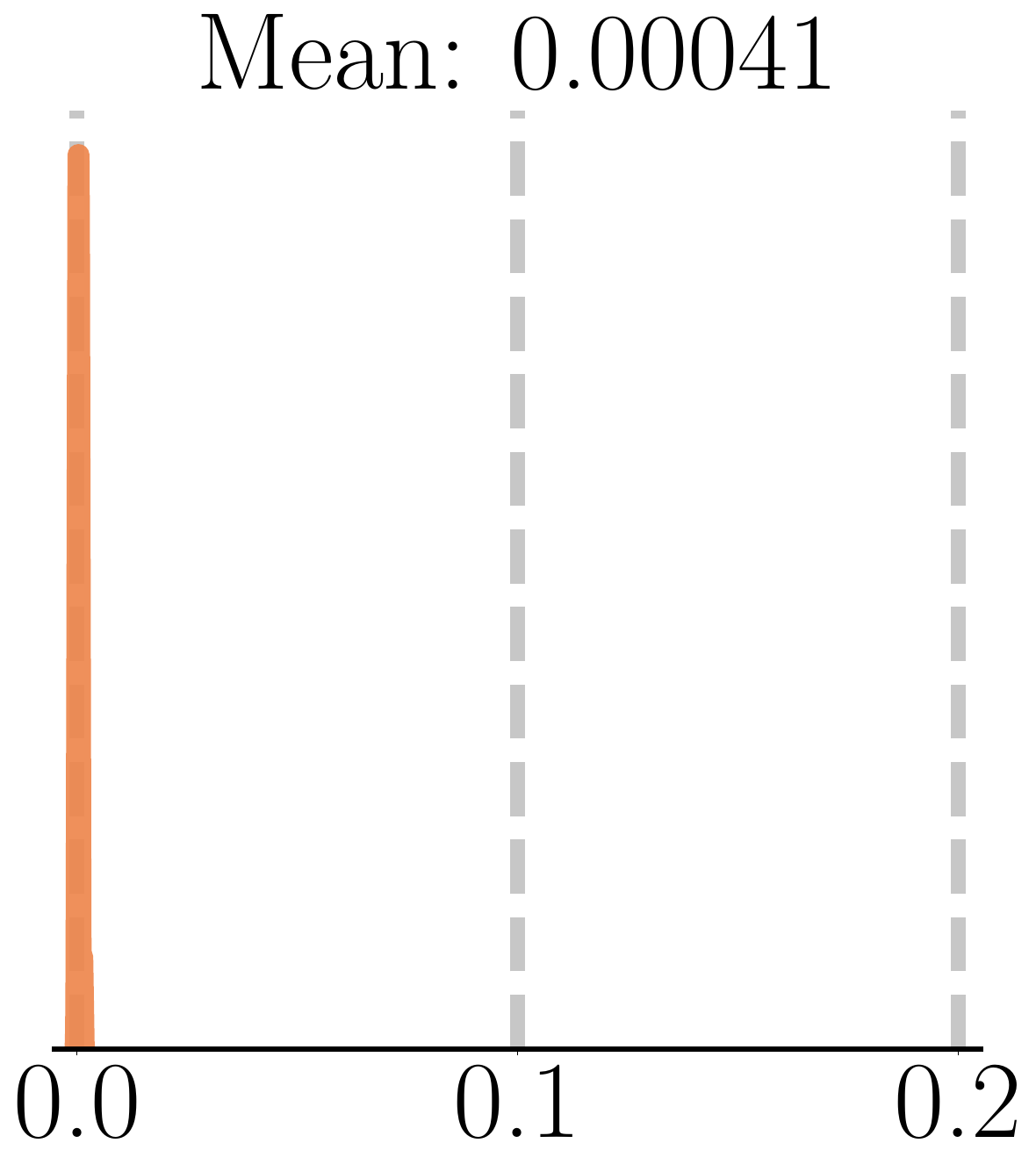}
            \end{subfigure} &
            \begin{subfigure}
                \centering
                \includegraphics[height=0.5\linewidth]{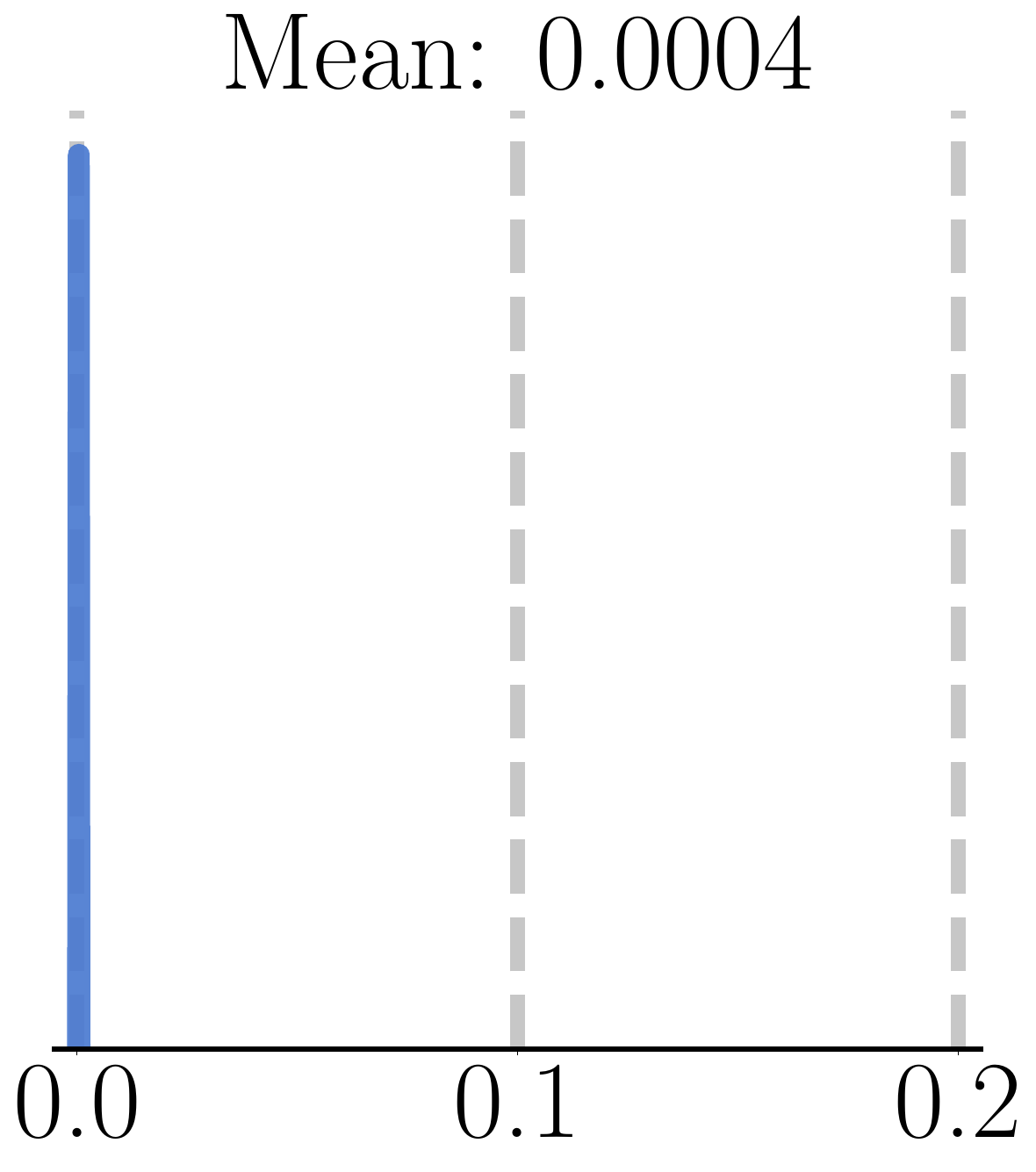}
            \end{subfigure} \\
            
            \multicolumn{2}{c}{\textbf{Visor}} & & \multicolumn{2}{c}{\textbf{Wallet}} & & \multicolumn{2}{c}{\textbf{Watch}} \\
            \addlinespace[1mm]
        \end{tabular}
    }
    \caption{Distribution of the volumes in meters of the bounding boxes of the objects in \ours dataset. Two plots are shown for each semantic category, reflecting respectively the objects of training (\textcolor{orange}{orange}) and validation (\textcolor{RoyalBlue}{blue}) splits.
    Each plot is accompanied by the corresponding mean of bounding box volumes of the objects in each split.}
    \label{fig:sizes_plots}
    \vspace{-.2cm}
\end{figure}

\tit{Object Size Analysis} 
Taking into account that personalized objects are defined as predefined instances with distinct characteristics, the primary challenge in the \shorttask task lies in effectively recognizing these specific details, especially when dealing with subtle features and limited interaction capabilities within the environment.
In this analysis, we present a category-wise size analysis of the objects in the dataset by computing and measuring the 3D bounding box of each object.
In Fig.~\ref{fig:sizes_plots}, we plot the distribution of the volumes of the bounding boxes associated with each object category showing that the distributions between training and validation splits remain consistent.

\tit{Category-wise Navigation Results}
In Table~\ref{tab:dinov2_per_category} we present the navigation results of the modular agent based on DINOv2 as the matching backbone in which we compute the metrics for each category. From the results on SR and SPL we can note that there are categories that are easier to locate and reach, such as `\textit{backpack}', `\textit{bag}', `\textit{ball}', `\textit{hat}', `\textit{laptop}', and `\textit{toy}', and there are instances from categories that are never correctly reached, such as `\textit{keys}', `\textit{wallet}', and `\textit{watch}'. This result returns the inability of the vanilla matching modules to distinguish these categories in the embodied setting. Moreover, we can observe that there is an overall positive correlation between SR and average category size, implying that small objects are particularly challenging to detect.

\begin{table}[!t]
    \centering
    \caption{Navigation results of the modular agent that employs DINOv2 as the matching module on the validation episodes of PInNED dataset, considering the performance of the agent for each category. Moreover, we report the average intra-category and inter-category cosine similarities computed on the frontal goal images.}
    \setlength{\tabcolsep}{.32em}
    \resizebox{\linewidth}{!}{
    \begin{tabular}{l c ccccc c cccc c cc}
        \toprule
        & & \multicolumn{5}{c}{\textbf{Navigation Metrics}} & & \multicolumn{4}{c}{\textbf{Detection Metrics}} && \multicolumn{2}{c}{\textbf{Similarity}} \\
        \cmidrule{3-7} \cmidrule{9-12} \cmidrule{14-15}
        Category & & SR$\uparrow$ & SPL$\uparrow$ & CE$\downarrow$ & D2G$\downarrow$ & Steps & & \%Match$\uparrow$ & TM$\uparrow$ & CM$\downarrow$ & NM$\downarrow$ & & Intra-Category & Inter-Category \\
        \midrule
        Backpack & & 26.47 & 14.04 & 36.77 & 5.79 & 408.7 & & 85.29 & 53.27 & 46.57 & 0.16 && 0.510 & 0.110 \\
        Bag & & 23.08 & 13.65 & 40.00 & 6.16 & 406.5 & & 93.85 & 44.62 & 55.04 & 0.34 && 0.348 & 0.121 \\
        Ball & & 20.90 & 10.29 & 23.88 & 6.48 & 613.1 & & 61.19 & 36.06 & 63.87 & 0.07 && 0.258 & 0.068 \\
        Book & & 19.40 & 10.83 & 35.82 & 5.71 & 484.3 & & 86.57 & 58.51 & 40.16 & 1.33 && 0.613 & 0.106 \\
        Camera & & 7.46 & 3.38 & 7.50 & 8.57 & 883.2 & & 20.90 & 69.23 & 23.08 & 7.69 && 0.152 & 0.050 \\
        Cellphone & & 8.96 & 3.11 & 14.92 & 8.63 & 844.8 & & 32.84 & 7.81 & 90.96 & 1.23 && 0.506 & 0.112 \\
        Eyeglasses & & 10.45 & 5.08 & 32.83 & 7.70 & 682.0 & & 62.69 & 79.80 & 19.95 & 0.25 && 0.846 & 0.104 \\
        Hat & & 26.87 & 11.95 & 23.88 & 6.45 & 652.8 & & 67.16 & 88.08 & 11.89 & 0.03 && 0.549 & 0.084 \\
        Headphones & & 16.92 & 9.71 & 40.00 & 7.35 & 492.8 & & 84.62 & 14.58 & 85.29 & 0.13 && 0.764 & 0.098 \\
        Keys & & 0.00 & 0.00 & 8.82 & 8.38 & 974.2 & & 2.94 & 0.00 & 0.00 & 100.00 && 0.558 & 0.102 \\
        Laptop & & 21.54 & 11.50 & 49.23 & 7.01 & 455.3 & & 93.85 & 16.86 & 82.60 & 0.54 && 0.348 & 0.084 \\
        Mug & & 10.61 & 4.47 & 10.61 & 8.10 & 911.8 & & 22.73 & 92.00 & 4.50 & 3.50 && 0.298 & 0.073 \\
        Shoes & & 16.92 & 12.44 & 44.62 & 6.75 & 318.8 & & 95.38 & 8.31 & 91.69 & 0.00 && 0.631 & 0.087 \\
        Teddy Bear & & 19.12 & 13.48 & 52.94 & 7.07 & 335.5 & & 91.18 & 68.92 & 16.62 & 14.46 && 0.548 & 0.066 \\
        Toy & & 26.56 & 13.18 & 3.12 & 6.16 & 754.6 & & 48.44 & 99.27 & 0.00 & 0.73 && 0.137 & 0.087 \\
        Visor & & 11.94 & 5.99 & 31.34 & 7.99 & 657.0 & & 52.24 & 52.47 & 45.33 & 2.20 && 0.316 & 0.148 \\
        Wallet & & 0.00 & 0.00 & 6.15 & 8.39 & 985.3 & & 1.54 & 0.00 & 0.00 & 100.00 && 0.282 & 0.105 \\
        Watch & & 0.00 & 0.00 & 7.69 & 8.39 & 999.0 & & 0.00 & 0.00 & 0.00 & 100.00 && 0.566 & 0.102 \\
        \bottomrule
    \end{tabular}
    }
    \vspace{-0.3cm}
\label{tab:dinov2_per_category}
\end{table}

\tit{Similarity Analysis}
The similarity of objects is a critical factor in the \shorttask task. The presence of distractors increases the challenge of the proposed task, as the agent must balance between being overly cautious and overly confident when identifying target instances. This trade-off is central to the effectiveness of the navigation approaches. In particular, concerning images as references of the target object, re-identification methods should be a robust solution against distractors due to considering the matching between keypoints instead of the semantic similarity between observation and reference. Indeed, in Table 2 of the main paper, the state-of-the-art re-identification method SuperGlue has a lower category error than DINOv2 and CLIP. However, it presents the worst results according to SR and SPL, showing difficulties in matching keypoints when observation and reference have discrepancies in appearance. For methods based on semantic features, the similarity threshold is the key element in balancing confidence and caution.

In Table~\ref{tab:dinov2_per_category}, we report the average cosine similarities in the DINOv2 embedding space per category. In particular, we extracted the CLS token from each frontal goal image of the validation set and computed the cosine similarities against the other goal images from the same category (i.e. intra-category) and against goal images from different categories (i.e. inter-categories). The results show that the intra-category similarity presents a strong relation with the category error (CE) and category matches (CM) metrics. Indeed, the agent tends to mistake instances from categories with large intra-category similarity values, such as `\textit{eyeglasses}', `\textit{headphones}', and `\textit{shoes}', while these mistakes are reduced in categories such as `\textit{camera}' and `\textit{toy}' that are characterized by a larger variability in their instances. When we adopt textual references as targets, the challenges concern how well multimodal spaces embed fine-grained details, and how similarity behaves accordingly. Previous work~\cite{bianchi2024devil,bravo2023open} has shown that this challenge is non-trivial and still open. Our dataset represents a further step in this direction, providing a benchmark to evaluate the capabilities of visual-language models in recognizing fine-grained details. Future works can exploit our training set to instruct the models to distinguish instances of the same category by focusing on adjectives and attributes.

\tit{Surfaces Details}
As described in Sec.~\ref{subsec:dataset}, the spawning position of each object in the \ours dataset is selected by sampling from the positions of a curated set of suitable surface macro-categories included in the semantic annotations of HM3D. 
The surface categories selected for the creation of the dataset are: \textit{armchair, bed, bench, cabinet, piano, rug, sofa, table}. These surfaces are valid for all the object categories and there are no subsets of surfaces dedicated to specific categories. There are categories, especially '\textit{shoes}', that are unlikely to be placed on certain surfaces. However, the scope of the task is to have objects that could be placed everywhere and teach a robotic agent to find them. A teddy bear is not necessarily located on the bed, but could be located anywhere, even on the kitchen table. If we assume a real-world scenario in which a child forgets the teddy bear on the kitchen table, the agent should not go directly to the bedroom, but look for the object in the whole environment. This is the reason for which we adopted a consistent spawning mechanism across all the categories. We identify this combination of objects that could be placed everywhere and the consistent spawning mechanisms as the correct approach for providing a dataset covering a large set of possible real-world scenarios that avoid the exploitation of prior knowledge on the object placement.

\begin{figure}[!t]
    \centering
    \small
    \resizebox{.78\linewidth}{!}{
        \begin{tabular}{rl}
            \begin{subfigure}
                \centering
                \includegraphics[height=\linewidth]{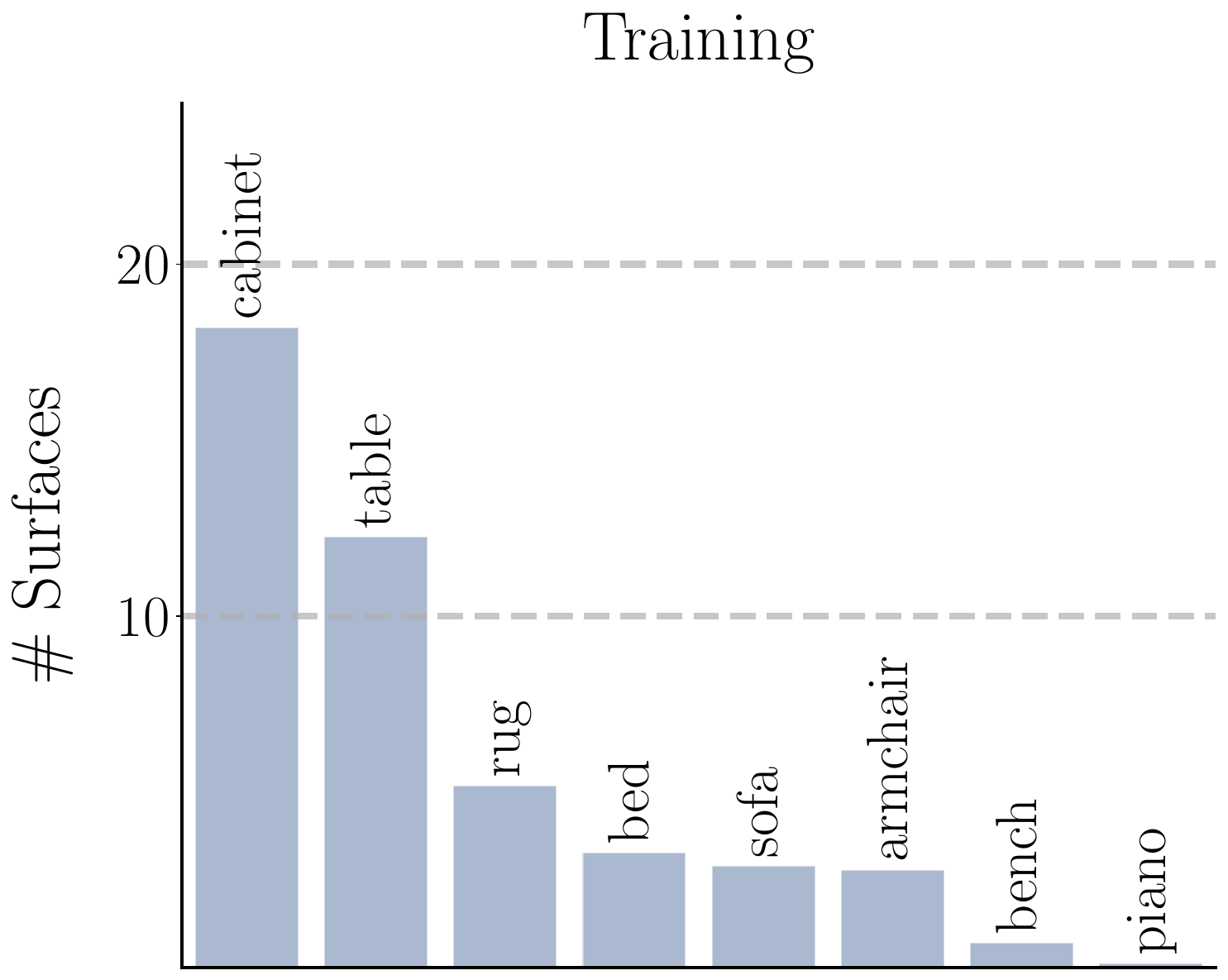}
            \end{subfigure} &
            \begin{subfigure}
                \centering
                \includegraphics[height=\linewidth]{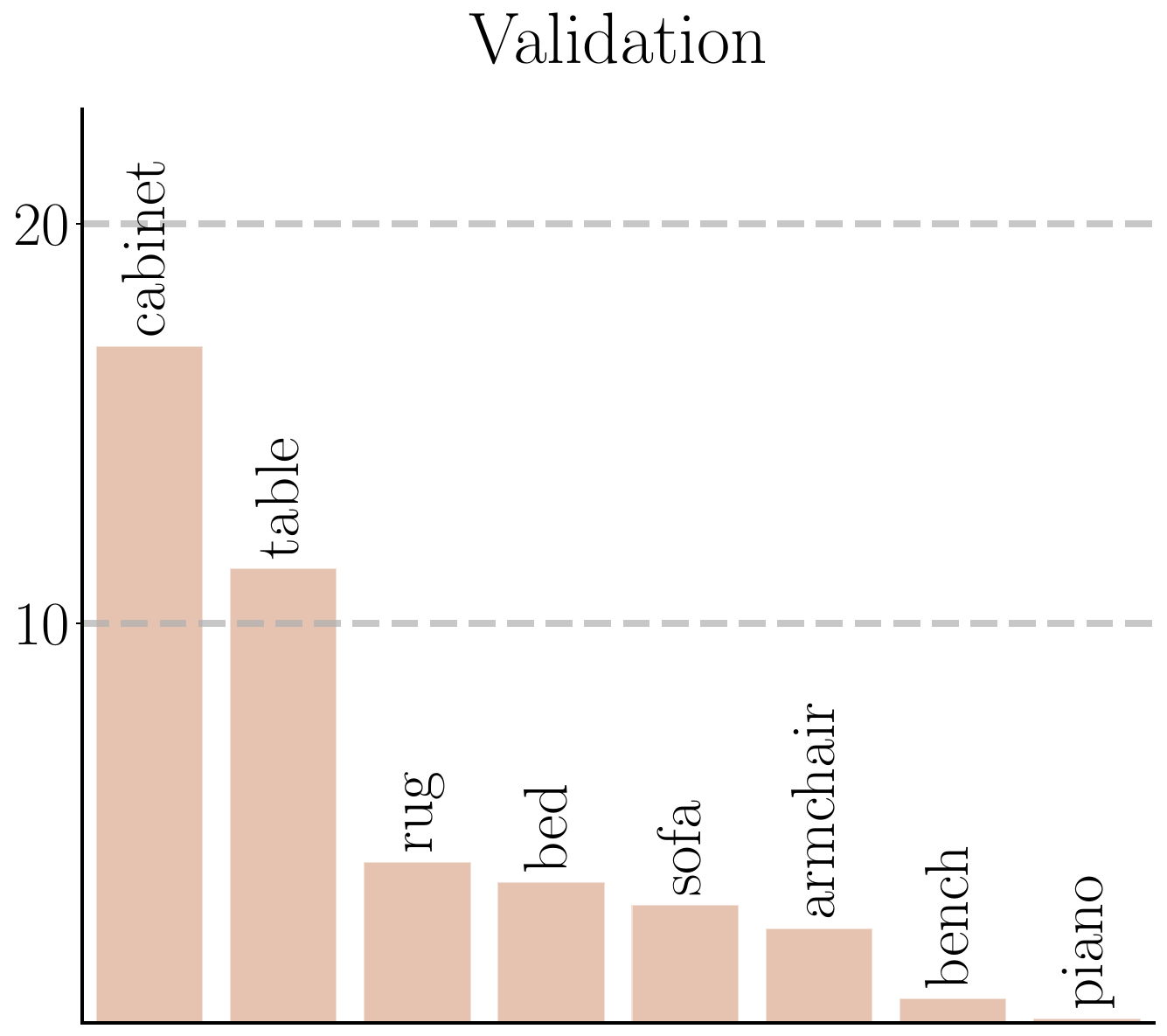}
            \end{subfigure} 
    
        \end{tabular}
    }
    \vspace{-.2cm}
    \caption{Plot of the mean number of surfaces in each environment that are suitable for object placement in the training (left) and validation (right) splits of the \ours dataset.}
    \label{fig:target_surfaces_frequencies}
    \vspace{-.2cm}
\end{figure}

In Fig.~\ref{fig:target_surfaces_frequencies}, we showcase the occurrences of the suitable surfaces in the environments of HM3D~\cite{ramakrishnan2021hm3d}. Notably, the distribution of spawnable surfaces remains consistent between the training and validation splits. This implies a recurring pattern in the furnishing of indoor spaces contained in the HM3D dataset and used for the \shorttask task. 

\begin{figure}[!t]
\vspace{0.3cm}
    \centering
    \Large
    \resizebox{\linewidth}{!}{
        \setlength{\tabcolsep}{.06em}
        \begin{tabular}{c c c c}
            % Row 1
            \includegraphics[height=0.15\textwidth]{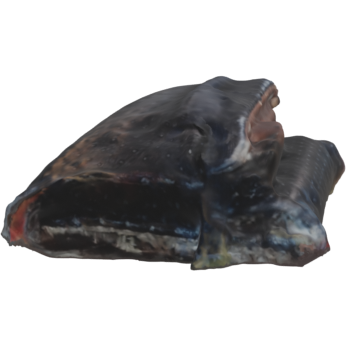} &
            \includegraphics[height=0.15\textwidth]{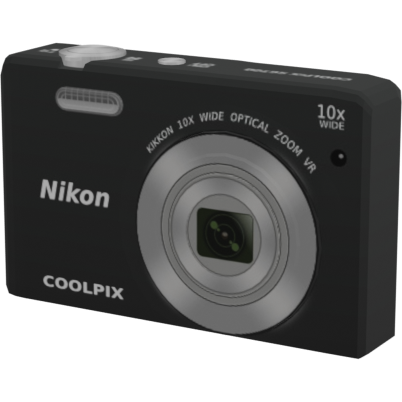} & \includegraphics[height=0.15\textwidth]{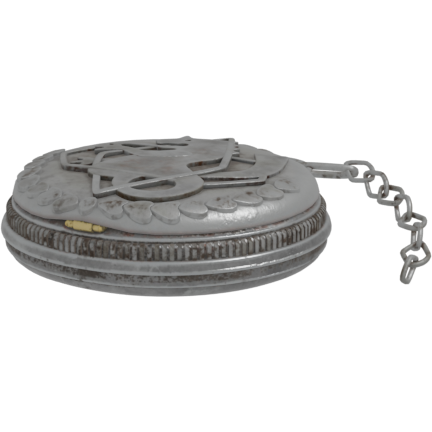} &
            \includegraphics[height=0.15\textwidth]{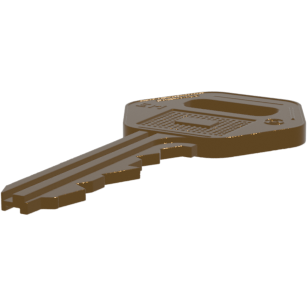} \\
            \addlinespace[6mm]
            \includegraphics[height=0.3\textwidth]{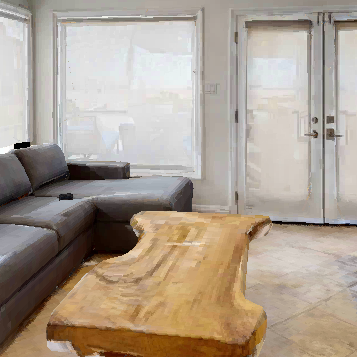} &
            \includegraphics[height=0.3\textwidth]{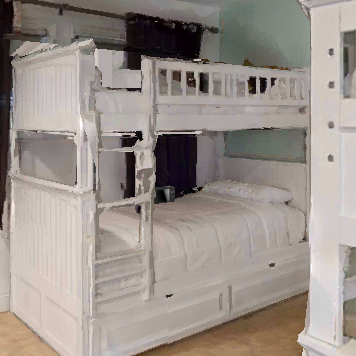} & \includegraphics[height=0.3\textwidth,width=0.3\textwidth]{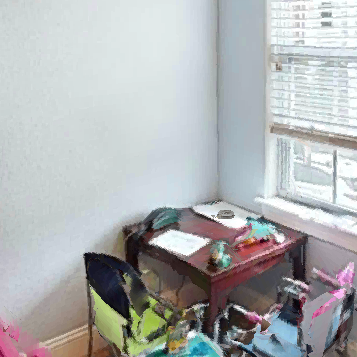} &
            \includegraphics[height=0.3\textwidth]{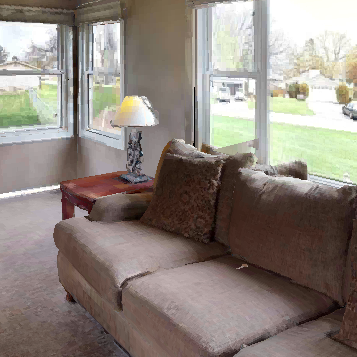} \\

        \end{tabular}
    }
    \caption{Examples of situations in which detecting the target in the embodied environment is particularly challenging. We depict the frontal visual references of the target in the first row and a portion of an agent's observation containing the target in the second row.}
    \label{fig:hard_detection}
\end{figure}

\vspace{0.3cm}
\tit{Hard Detection Cases}
\label{sec:hard_cases}
In Fig.~\ref{fig:hard_detection}, we show four episodes in which detecting the target is particularly challenging. These targets belong, respectively, to the `\textit{wallet}', `\textit{camera}', `\textit{watch}', and `\textit{keys}' categories. Table~\ref{tab:dinov2_per_category} shows that these categories are the most challenging ones for the modular agent with DINOv2, which is the best-performing agent according to Table~\ref{tab:pin_main}. Indeed, the categories `\textit{keys}', `\textit{wallet}', and `\textit{watch}' all yielded no successful episodes. These objects are hard to detect even for a human, confirming how challenging the \shorttask task is. Future work should investigate the possibility of moving the agent closer to areas in which there are small objects that cannot be identified as the target from longer distances.

\begin{table}[!t]
    \centering
    \caption{Navigation results of the modular agent that employs SuperGlue as the matching module on the validation episodes of \ours dataset, considering different resize values of the visual references of the target provided to the matching module.}
    \setlength{\tabcolsep}{.32em}
    \resizebox{.7\linewidth}{!}{
    \begin{tabular}{l c ccccc c cccc}
        \toprule
        & & \multicolumn{5}{c}{\textbf{Navigation Metrics}} & & \multicolumn{4}{c}{\textbf{Detection Metrics}} \\
        \cmidrule{3-7} \cmidrule{9-12}
        Resize & & SR$\uparrow$ & SPL$\uparrow$ & CE$\downarrow$ & D2G$\downarrow$ & Steps & & \%Match$\uparrow$ & TM$\uparrow$ & CM$\downarrow$ & NM$\downarrow$ \\
        \midrule
        360 & & 2.51 & 0.82 & \textbf{7.05} & 8.48 & 881.6 & & 17.77 & \textbf{43.76} & 5.17 & \textbf{51.07} \\
        180 & & 3.02 & 1.20 & 7.21 & 8.48 & 864.1 & & 20.70 & 21.72 & 3.58 & 76.35 \\
        180, 360 & & \textbf{3.27} & \textbf{1.28} & 7.58 & \textbf{8.36} & \textbf{804.0} & & \textbf{29.42} & 16.96 & \textbf{3.44} & 79.60 \\
        \bottomrule
    \end{tabular}
    }
    \vspace{-0.2cm}
\label{tab:superglue_resolutions}
\vspace{0.3cm}

\end{table}

\tit{Fine-Grained vs General Descriptions Comparison} In Table \ref{tab:pin_descr_reb} we present an ablation study in which we compare the performance of the baselines with both fine-grained and general object categories. Specifically, we conducted the following experiments. For the modular agents based on CLIP and OWL as the matching module, we leveraged the general object category (e.g. backpack) instead of the fine-grained textual descriptions as navigation targets, while maintaining the same similarity threshold. The results on both CLIP and OWL present similar behaviors: the number of successful episodes is slightly increased, but also the number of episodes in which the agent mistakes reaching distractors of the same category and the number of matches with them increased. Moreover, the reduction in the average number of steps indicates that similarities, on average, are higher. The increase in successful episodes is surprising but in line with the findings of previous works in the literature~\cite{bianchi2024devil,bravo2023open}, which demonstrate that current vision-language models struggle with fine-grained details. These results show that our work can help future works in the realization and evaluation of vision-language models with improved understanding capabilities of details. Concerning the end-to-end agent RIM, we trained the model on the CLIP embeddings extracted from the general object categories instead of the fine-grained textual descriptions. The results show a lower number of successful episodes and a higher number of episodes in which the agent reaches a distractor of the same category.

\begin{table}[!t]
    \centering
    \caption{Navigation results on \ours on the environments of HM3D dataset, comparing categorical and fine-grained textual modalities.}
    \setlength{\tabcolsep}{.32em}
    \resizebox{\linewidth}{!}{
    \begin{tabular}{lcc c ccccc c cccc}
        \toprule
        & & & & \multicolumn{5}{c}{\textbf{Navigation Metrics}} & & \multicolumn{4}{c}{\textbf{Detection Metrics}} \\
        \cmidrule{5-9} \cmidrule{11-14}
        & Backbone & Modality & & SR$\uparrow$ & SPL$\uparrow$ & CE$\downarrow$ & D2G$\downarrow$ & Steps & & \%Match$\uparrow$ & TM$\uparrow$ & CM$\downarrow$ & NM$\downarrow$ \\
        \midrule
        \rowcolor{light_gray}
        \textit{Modular Agents} & & & & & & & & & & & & & \\
        \addlinespace[1mm]
        \hspace{0.3cm}CLIP~\cite{radford2021learning} & ViT-B/16 & Categorical & & 3.52 & 2.75 & 10.23 & 7.98 & 148.1 & & 95.47 & 5.12 & 15.73 & 79.15 \\
        \hspace{0.3cm}CLIP~\cite{radford2021learning} & ViT-B/16 & Fine-Grained & & 3.10 & 1.82 & 9.31 & 7.94 & 503.1 & & 62.95 & 20.07 & 22.07 & 57.86 \\
        \addlinespace[1mm]
        \hspace{0.3cm}OWL~\cite{gadre2022cow,minderer2022simple} & ViT-B/32 & Categorical & & 7.79 & 3.73 & 19.96 & 7.96 & 780.6 & & 38.81 & 26.50 & 58.49 & 15.01 \\
        \hspace{0.3cm}OWL~\cite{gadre2022cow,minderer2022simple} & ViT-B/32 & Fine-Grained & & 7.29 & 3.36 & 12.66 & 7.90 & 871.7 & & 22.97 & 62.60 & 32.88 & 4.52\\
        \midrule
        \rowcolor{light_gray}
        \textit{End-to-end Agents} & & & & & & & & & & & & & \\
        \addlinespace[1mm]
        \hspace{0.3cm}RIM~\cite{chen2023object} & ResNet-50 & Categorical & & 4.61 & 3.78 & 14.25 & 9.23 & 336.0 & & - & - & - & - \\
        \hspace{0.3cm}RIM~\cite{chen2023object} & ResNet-50 & Fine-Grained & & 7.12 & 6.67 & 10.44 & 8.43 & 409.3 & & - & - & - & -\\
        \bottomrule
    \end{tabular}
    }
    \vspace{-0.2cm}
\label{tab:pin_descr_reb}
\end{table}

\section{Additional Implementation Details}

\tit{Modular Agents} In Sec.~\ref{subsec:modular_agents}, we introduce the modular agents tested on the \shorttask task. Their ability to distinguish a specific instance in a given observation depends on the score threshold that maximizes the detection results. We tune this threshold on a subset of the training episodes. For all the backbones except for SuperGlue~\cite{sarlin2020superglue}, we extract two squared crops with size $360\times360$ from the $360\times640$ observation and resize them to the image resolutions on which the backbones have been trained. Then, we consider all the matches resulting from the two crops. At least a match over the threshold is required to consider the goal detected in an observation. For the textual modalities, we employ the 80 prompt templates proposed by Radford~\etal~\cite{radford2021learning} for ImageNet~\cite{deng2009imagenet}. In this section, we report additional implementation details for each backbone.
\begin{itemize}[left=3mm]%,noitemsep,topsep=0pt]
    \item[\pin] \textbf{SuperGlue~\cite{sarlin2020superglue}}: We observe that SuperGlue struggles to match the visual references with the observations of the agent and that the resolution of the references influences the matching capabilities. In particular, we provide the visual references to SuperGlue as squared images $360\times360$, corresponding to the shortest side of the observation of the agent. For each visual reference, namely for each of the three views of the object, we provide two resizes of the object such that the longest side is, respectively, 360 and 180. This procedure results in two reference images for each view of the object, an image entirely occupied by the object and an image where the object occupies a quarter of it. In Table~\ref{tab:superglue_resolutions} we show that this approach results in a higher success rate than having a single image per object view. Moreover, we employ the \textit{indoor} weights of SuperGlue with a threshold of 0.2 on the confidence of each matched keypoints pair and a matching threshold $\sigma$ of 8.0 on the confidence sum of all the matched keypoints pairs.
    \item[\pin] \textbf{CLIP~\cite{radford2021learning}}: We employ CLIP ViT-B/16 with the pre-trained weights from OpenAI for both the experiments with visual and textual references. We resize the two observation crops to $224\times224$, resulting in a grid of $14\times14$ patches. The best matching threshold $\sigma$ for the visual and textual modalities are, respectively, 0.575 and 0.28.
    \item[\pin] \textbf{CLIP-Grad}: We follow the implementation of the network interpretability method proposed in CoW~\cite{gadre2022cow} on top of CLIP with textual references. We employ CLIP ViT-B/32 with the pre-trained weights from OpenAI and matching threshold 0.85.
    \item[\pin] \textbf{OWL~\cite{minderer2022simple}}: OWL is an open-vocabulary detector that is trained in two steps: (i) a large contrastive image-text pre-training following LiT~\cite{zhai2022lit} and (ii) an object-level training on publicly available detection datasets (Open Images V4~\cite{kuznetsova2020open}, Objects 365~\cite{shao2019objects365}, and Visual Genome~\cite{krishna2017visual}). We employ a matching threshold of 0.25 applied to the predicted bounding box scores.
    \item[\pin] \textbf{DINO~\cite{caron2021emerging}/DINOv2~\cite{oquab2023dinov2}}: DINO is a self-supervised backbone pre-trained according to a self-distillation training paradigm. DINOv2 is an improved version of DINO with the aim of producing general-purpose visual features. We employ DINO ViT-B/16 and DINOv2 ViT-B/14 trained, respectively, on ImageNet-1k~\cite{deng2009imagenet} and LVD-142M~\cite{oquab2023dinov2}. We use the same input image resolutions on which they are trained, namely $224\times224$ and $518\times518$, producing $14\times14$ and $37\times37$ grids of patches. The best matching scores are, respectively, 0.575 and 0.5.
    \item[\pin] \textbf{PerSAM/PerSAM-F~\cite{zhang2023personalize}}: We leverage the implementation of PerSAM on SAM ViT-B/16, trained on SA-1B, with input image resolution at $1,024$. PerSAM-F is a variant of PerSAM that fine-tunes the model on the reference image, We follow the training configuration of the original implementation. We consider the maximum patch-level similarity between the reference images and the observation crop as the matching score on which we apply the thresholds 0.925 and 0.61 for, respectively, PerSAM and PerSAM-F.
\end{itemize}

\tit{End-to-End Agents}
As mentioned in Sec.~\ref{subsec:endtoend_agents} end-to-end approaches use a neural network policy which is trained end-to-end to directly process sensor observations and predict the atomic actions needed to fulfill the required task. In our case, we adapted two recent end-to-end approaches for ObjectNav finetuning them to perform \shorttask task: RIM~\cite{chen2023object} and ZSON~\cite{majumdar2022zson}.
\begin{itemize}[left=3mm]%,noitemsep,topsep=0pt]
    \item[\pin] \textbf{RIM~\cite{chen2023object}}: The model is finetuned using behavior cloning following Chen~\etal~\cite{chen2023object} approach and starting from the pre-trained weights for ObjectNav~\cite{batra2020objectnav}. 
    We evaluate two variants of the fine-tuned model, conditioned on visual features and conditioned on textual features. 
    In RIM approach, besides the episodic implicit map that is updated recursively, the input of the policy at each timestep is composed of the concatenation of the features extracted from RGB and depth observation, the pose of the agent, previous action, and the target object category. To adapt RIM for the \shorttask task, we modify the features extracted from the object category label. 
    Originally each label is associated with a row in a lookup table containing learnable embeddings of length $32$. 
    In our adaptation, we replace such embeddings with CLIP (ViT-B/16) features extracted using the visual or textual references. Since each input reference modality is described by $3$ images or descriptions, we compute the mean of the features extracted from each reference. Following, a learnable linear layer is trained to project CLIP features to a vector of length $32$. 
    The resulting embedding is used to condition the navigation of the RIM agent. 
    The fine-tuning process is performed on a single GPU for a total of $\approx2$M fine-tuning steps over $\approx24$ hours.
    
    \item[\pin] \textbf{ZSON~\cite{majumdar2022zson}}: For the adaptation of the ZSON method, we fine-tuned the model pre-trained on the ImageNav task, following the same approach as Majumdar~\etal~\cite{majumdar2022zson}. The agent is fine-tuned with reinforcement learning using an adaptation of ZSON reward but ignoring the angle to the goal since it is not a component considered in the \shorttask task. 
    The resulting reward is $r_t = r_{success} - \Delta_{dtg} + r_{slack}$. We refer to Majumdar~\etal~\cite{majumdar2022zson} for a description of the components of the reward. 
    Moreover, while the original approach uses ImageNav goals that are represented as photos captured at the position that the agent is required to reach, we used image references of the target instance to perform the fine-tuning. 
    The model is fine-tuned on a single GPU for $\approx24$ hours for a total of $\approx5$M fine-tuning steps.
\end{itemize}

\tit{Compute Information} We performed our experiments on a computing platform composed of NVIDIA RTX5000 GPUs and 8 GB of CPU memory for each job. A job can be computed on a single GPU. Each episode step for the modular agents requires an average of $\approx200$ms to be executed. Hence, the entire DINOv2 experiment on the $1,193$ validation episodes, with an average number of steps equal to $658.7$, requires $\approx44$ computation hours. The entire evaluation on the validation split for the end-to-end agents requires $\approx5$ computation hours.

\begin{lstlisting}[caption={Python dictionary containing a sample of the episodes contained in \ours dataset. The list of distractors is skimmed for better visualization.},captionpos=b,language=json,firstnumber=1,label={lst:episodejson}]
{
    "episode_id": "0",
    "scene_id": "hm3d/val/00800-TEEsavR23oF/TEEsavR23oF.basis.glb",
    "start_position": [-0.28, 0.013, -6.54],
    "start_rotation": [0, 0.98, 0, 0.20],
    "info": {"geodesic_distance": 8.24},
    "goals": [
        {
            "object_category": "backpack",
            "object_id": "3f5948f7f47343acb868072a7fe92ada",
            "position": [-5.13, 1.08, -0.81]
        }
    ],
    "distractors": [
        {
            "object_category": "backpack",
            "object_id": "3c47af8b6a3e413f94c74f86d4c396ed",
            "position": [-3.46, 2.20, -4.30]
        },
        {
            "object_category": "backpack",
            "object_id": "0b795895343b44b69191ef9b55b35840",
            "position": [-11.17, 0.88, -0.36]
        },
        {
            "object_category": "backpack",
            "object_id": "d86ee61984544b45a9f11f49e5e02c43",
            "position": [-9.13, 1.22, -3.52]
        },
        {
            "object_category": "mug",
            "object_id": "d26e9bfce2644bb7af6710c6511ea718",
            "position": [-7.84, 0.62, -0.14],
        },
        {
            "object_category": "laptop",
            "object_id": "6495988c6c044c76a2fc9f9278543c16",
            "position": [-1.64, 0.87, -6.15],
        },
        {
            "object_category": "headphones",
            "object_id": "ccf60b0502784fb38e483a6b07cfad53",
            "position": [3.41, 0.84, -8.21],
        },
    ],
    "scene_dataset_config": "data/scene_datasets/hm3d/hm3d_annotated_basis.scene_dataset_config.json",
    "object_category": "backpack",
    "object_id": "3f5948f7f47343acb868072a7fe92ada"
}
\end{lstlisting}

\begin{lstlisting}[caption={Python dictionary containing the information used by Habitat simulator to instantiate a specific object instance in the environment.},captionpos=b,language=json,firstnumber=1,label={lst:objectjson}]
{
    "scale": [0.116, 0.116, 0.116],
    "render_asset": "0a96f1f19afc432bb22c3d74da546338.glb",
    "requires_lighting": true,
    "up": [0.0, 1.0, 0.0],
    "front": [0.0, 1.0, 0.0],
    "COM": [0.0, 0.0, 0.0],
    "gravity": [0, 0, 0],
    "force_flat_shading": true,
    "is_collidable": true,
    "use_mesh_for_collision": true,
    "semantic_id": 2,
    "semantic_category": "ball"
}
\end{lstlisting}

\section{Licenses and Terms of Use}
The episodes of the \ours dataset are built using the scenes from the HM3D dataset~\cite{ramakrishnan2021hm3d}. The scenes of the HM3D dataset are released under the Matterport End User License Agreement, which permits non-commercial academic use.

For the augmentation of HM3D scenes with additional objects, \ours dataset utilizes 3D object assets from Objaverse-XL dataset~\cite{deitke2023objaverse}. Objaverse-XL is distributed under the ODC-By 1.0 license, with individual objects retrieved from various sources, including GitHub, Thingiverse, Sketchfab, Polycam, and the Smithsonian Institution. Each object is subject to the licensing terms of its respective source, necessitating users to evaluate license compliance based on their specific downstream applications.

Nevertheless, the specific objects included in our dataset are restricted to assets sourced from Sketchfab which are released under various Creative Commons licenses. Specifically, the dataset includes assets under the following licenses: CC BY (311 objects), CC BY-NC (14 objects), CC BY-SA (8 objects), CC BY-NC-SA (3 objects), and CC0 (2 objects).

The episodes of the \ours dataset, along with the manually annotated object descriptions are released under the CC BY license, while the codebase for the \shorttask task is released under the MIT license.

The authors accept full responsibility for any rights violations arising from the use or publication of the data and content in this paper. All licenses related to external content included in this paper ensure no infringement on third-party rights.

\section{Assets} 
\label{sec:assets}

The episodes of \ours dataset are defined as Python dictionaries containing relevant information for the execution of the \shorttask task with the Habitat simulator. An example of episode annotation is presented in Listing~\ref{lst:episodejson}. Each episode specifies the environment where it is taking place, the starting position and rotation of the agent, along with the position and object identifier of the target instance and the distractors.

The information used by the Habitat simulator to resize and instantiate each 3D object at the position specified by the episodes of \ours dataset is also contained in a Python dictionary, where a specific file represents each object. 
A sample of object annotation is showcased in Listing~\ref{lst:objectjson}.

\section{Datasheet}
In this section, we present a comprehensive datasheet~\cite{gebru2021datasheets} for the proposed dataset, providing a unified reference for relevant information on the \ours episodes and the objects used to build the dataset.

\subsection{Motivation}
\titsheet{For what purpose was the dataset created?} The \ours dataset has been built with the motivation of fostering future research on smart navigation agents. Such agents need to acquire the capability of distinguishing between different instances of the same object category and leverage different modalities of inputs to reach a specific object asked by the user. The dataset introduces a novel task in Embodied AI research and, in order to run the episode of the \ours dataset, the Habitat simulator needs to be used. Instructions on how to run and instantiate the episodes of \ours dataset are included in the public repository described in Sec.~\ref{sec:code_release}.

\titsheet{Who created the dataset and on
behalf of which entity?} The dataset was created by researchers at the University of Modena and Reggio Emilia.

\titsheet{Who funded the creation of the dataset?} Refer to the Acknowledgments and Disclosure of Funding section in the main paper.

\subsection{Composition}
\titsheet{What do the instances that comprise the dataset represent?} The \ours dataset consists of generated navigation episodes designed to address the \shorttask task, accompanied by a list of object identifiers used in each episode within the Habitat simulator. As the dataset is composed of navigation episodes, containing all necessary information for the simulator to execute the task, no additional metadata is provided. However, an example of episode annotations is included in Listing~\ref{lst:episodejson}.

\titsheet{How many instances are there in total?}
The dataset of episodes for the \shorttask task is composed of a total of $865, 519$ training episodes and $1,193$ validation episodes. Moving on to the objects contained in the \ours dataset, the total number of unique object instances that are injected in the navigation environments is $338$.

\titsheet{Does the dataset contain all possible instances or is it a sample
(not necessarily random) of instances from a larger set?}
While episodes of the \ours dataset are generated procedurally by the authors of the paper, the objects used as additional objects are part of the objects released from Objaverse-XL dataset~\cite{deitke2023objaverse}, which is composed of 3D models from different online sources such as GitHub, Thingiverse, Sketchfab, Polycam, and the Smithsonian Institution. The objects of \ours are however restricted to 3D models included in Sketchfab.

\titsheet{What data does each instance consist of?}
The dataset content is defined by the information of the episodes for the \shorttask task. Each episode is represented as a dictionary containing the information needed by the Habitat simulator~\cite{savva2019habitat} to execute the task. A \textit{.json} file including a list of the navigation episodes is produced for each scene included in HM3D dataset. 
We refer to Listing~\ref{lst:episodejson} for a sample of episode annotation. 
Each episode in the dataset specifies additional objects that are placed at a specific location loading \textit{.glb} files containing the meshes of the objects. The \textit{.glb} files used to instantiate the episodes of the \ours dataset are downloadable from Objaverse-XL API using the Python script provided in the codebase.
Each 3D object is associated with a \textit{.json} file containing a dictionary with the information needed by the Habitat simulator to correctly instantiate the object in the environment in terms of size and appearance.

\titsheet{Is there a label or target associated with each instance?} Each object used for the \ours dataset is manually associated with an object category label to correctly perform the placement procedure of distractors belonging to the same category of the target instance, as well as computing metrics related to the \shorttask task. 
However, the object category label should not be used by the agent to tackle the \shorttask task.
For each episode, only one instance is defined as the correct target to complete the task successfully.

\titsheet{Are there recommended data splits?}
The episodes of the \ours dataset are divided into training and validation splits depending on the environment where the episodes are taking place. The environments are divided into training and validation splits following the environmental-level division performed by Ramakrishnan~\etal~\cite{ramakrishnan2021hm3d}.
Regarding the additional objects included in \ours dataset, the object instances are divided into $266$ training instances and $72$ validation instances. It is worth noting that the sets of instances used for the training and validation splits do not overlap.

\titsheet{Are there any errors, sources of noise, or redundancies in the dataset?}
The additional objects on the surfaces of HM3D environments could be 
misplaced due to noise in the original annotations of the scene, or due to the presence of clutter at the acquisition time of the environment.
Other sources of noise could be related to possible typos in the process of annotation of the textual descriptions of the additional objects.

\titsheet{Is the dataset self-contained, or does it link to or otherwise rely on external resources?}
The \ours dataset relies on the scenes included in the HM3D dataset of 3D spaces and on the 3D object assets included in the Objaverse-XL dataset.

\titsheet{Does the dataset contain data that might be considered confidential? Does the dataset contain data that, if viewed directly, might be offensive, insulting, threatening, or might otherwise cause anxiety?}
No confidential or disturbing data is contained in the content of \ours dataset.

\subsection{Collection Process}
\titsheet{How was the data associated with each instance acquired? What mechanisms or procedures were used to collect the data?} The additional objects used for the \ours dataset are manually selected using the Python API from Objaverse-XL dataset. 

The generation of the visual references of the target objects has been performed using Blender, where the 3D mesh of the object is rendered and captured in an isolated setting. 
The camera performs a 30-degree yaw rotation around the object to capture a favorable view of the objects. Then, each instance is rotated by 180 degrees in yaw to view its reverse side, while a 90-degree pitch rotation is used to observe the upper side of the object. This procedure produces three visual references for each target object.

The process of annotation of the textual descriptions of each object is performed by the authors of the paper. 
Two objects of the same object category are shown to each annotator that is required to describe one of the two instances in a way that is distinguishable from the other. The final procedure used three annotators, for a total of three textual descriptions for each object. Samples of the input references related to the objects of \ours dataset are shown in Fig.~\ref{fig:dataset_descriptions1} and Fig.~\ref{fig:dataset_descriptions2}.

The episodes of \ours are generated by spawning the selected additional objects in the scene after extracting all suitable surfaces from the semantic annotation of the environment. We refer to Sec.~\ref{subsec:dataset} for more details on object placement.

\titsheet{If the dataset is a sample from a larger set, what was the sampling strategy?} The sampling strategy for selecting the objects from Objaverse-XL is based on a manual assessment of the photo-realistic properties of the selected objects and the corresponding visual appearance of the object when rendered using the Habitat simulator.
The sampling strategy of the objects contained in the episodes of \ours dataset is a random sampling. For each episode, a goal object category is selected, and a specific target instance is sampled from the set of suitable objects. Instances belonging to the same object category of the target object are sampled and positioned in the environment as distractors. If other spawnable surfaces are available, more distractors belonging to other object categories are placed in the environment. Details on the number of additional objects placed in the episodes of \ours dataset are included in Sec.~\ref{subsec:additional_navigation_details}.
For the final generation of the episodes of \ours dataset, $400$ episodes are generated for each possible object category on the environments of the training split, while $2$ episodes for each object category are generated in every environment of the validation split.

\titsheet{Who was involved in the data collection process?}
The actors performing the data collection and annotation process of the dataset are the authors of the paper.

\titsheet{Over what timeframe was the data collected?}
The dataset assets were collected and the episodes of the \ours dataset were generated between November 2023 and May 2024.

\titsheet{Were any ethical review processes conducted?}
No ethical review process was necessary for the collection of the dataset.

\subsection{Preprocessing / Cleaning / Labeling}

\titsheet{Was any preprocessing/cleaning/labeling of the data done?}
The objects used in the \ours dataset are manually resized when rendered with the Habitat simulator adjusting their dimension compared with the surrounding environment to be similar to their real-world counterpart. 
Each 3D object is associated with a corresponding object category label to allow the usage of different instances of the same object category when tackling the \shorttask task.
The episodes of \ours dataset are, instead, validated using the Habitat simulator to remove any episode containing objects that are not reachable from the starting position of the agent. 

\subsection{Uses}

\titsheet{Has the dataset been used for any tasks already?}
The \ours dataset can be used to train and evaluate agents for the \fulltask (\shorttask) task. See Sec.~\ref{sec:task} and Sec.~\ref{sec:experiments} for more details on the task definition and the experimental evaluation.

\titsheet{What (other) tasks could the dataset be used for?}
The dataset could be used for other tasks involving recognition or manipulation on specific instances using visual or textual references as input.

\titsheet{Is there anything about the composition of the dataset or the way it was collected and preprocessed/cleaned/labeled that might impact future uses?}
Users need to follow and respect the licenses associated with the additional 3D objects and the episodes contained in this dataset.

\subsection{Distribution}

\titsheet{How will the dataset will be distributed?} 
The dataset is made public through the release of a public GitHub repository. The repository containing dataset and codebase is released at this url: \href{https://github.com/aimagelab/pin}{https://github.com/aimagelab/pin}.

\titsheet{When will the dataset be distributed?} The dataset has been publicly released on October 2024.

\titsheet{Will the dataset be distributed under a copyright or other intellectual property (IP) license, and/or under applicable terms of use (ToU)?} The dataset and the object annotations are released under the CC BY license. The codebase is released under the MIT license. 
The additional objects contained in the episodes of \ours dataset are subject to the licenses that they are released under.

\titsheet{Have any third parties imposed IP-based or other restrictions on the data associated with the instances?} Any restrictions are related to additional objects and to the licenses which they are released under. Users need to assess license questions based on their use.

\subsection{Maintenance}

\titsheet{Who will be supporting/hosting/maintaining the dataset?}
The dataset will be maintained by the authors of the paper who commit to maintaining the dataset long-term.

\titsheet{How can the owner/curator/manager of the dataset be contacted?}
The authors can be contacted at \texttt{\{firstname.lastname\}@unimore.it}.

\titsheet{Will the dataset be updated?}
A potential future update could involve extending the dataset to include a test split, upon receiving permission from the HM3D dataset owners to access the environments in the test split.

\titsheet{If others want to extend/augment/build on/contribute to the dataset, is there a mechanism for them to do so?}
Users are free to extend the dataset at the condition of following and respecting the licenses associated with the dataset and associated additional objects by contacting the authors on the public repository.

\end{document}